\documentclass[lettersize,journal]{IEEEtran}
\usepackage{amsmath,amsfonts}
\usepackage{algorithmic}
\usepackage{algorithm}
\usepackage{array}
\usepackage[caption=false,font=normalsize,labelfont=sf,textfont=sf]{subfig}
\usepackage{textcomp}
\usepackage{stfloats}
\usepackage{url}
\usepackage{verbatim}
\usepackage{graphicx}
\usepackage{cite}
\usepackage{orcidlink}
\usepackage{multirow}
\usepackage{booktabs}
\usepackage{amsmath}

\begin{document}

\title{AnySR: Realizing Image Super-Resolution as Any-Scale, Any-Resource}

\author{
Wengyi Zhan,~\IEEEmembership{}
Mingbao Lin,~\IEEEmembership{}
Chia-Wen Lin,~\IEEEmembership{Fellow,~IEEE,}
Rongrong Ji,~\IEEEmembership{Senior Member,~IEEE}
\thanks{Manuscript received May XX, 2024.  (Corresponding author: Rongrong Ji)}
\thanks{W. Zhan and R. Ji are with the Key Laboratory of Multimedia Trusted Perception and Efficient Computing, Ministry of Education of China, Xiamen University, China. E-mail: zhanwy@stu.xmu.edu.cn, rrji@xmu.edu.cn}
  \thanks{M. Lin is with the Skywork AI, Singapore. E-mail: linmb001@outlook.com}
  \thanks{C.-W. Lin is with the Department of Electrical Engineering and Institute of Communications Engineering, National Tsing Hua University, Taiwan. E-mail: cwlin@ee.nthu.edu.tw}
  }

\markboth{Journal of \LaTeX\ Class Files,~Vol.~14, No.~8, August~2021}%
{Shell \MakeLowercase{\textit{et al.}}: A Sample Article Using IEEEtran.cls for IEEE Journals}


\maketitle

\begin{abstract}
In an effort to improve the efficiency and scalability of single-image super-resolution (SISR) applications, we introduce AnySR, to rebuild existing arbitrary-scale SR methods into any-scale, any-resource implementation.
As a contrast to off-the-shelf methods that solve SR tasks across various scales with the same computing costs, our AnySR innovates in:
1) building arbitrary-scale tasks as any-resource implementation, reducing resource requirements for smaller scales without additional parameters;
2) enhancing any-scale performance in a feature-interweaving fashion, inserting scale pairs into features at regular intervals and ensuring correct feature/scale processing.
The efficacy of our AnySR is fully demonstrated by rebuilding most existing arbitrary-scale SISR methods and validating on five popular SISR test datasets.
The results show that our AnySR implements SISR tasks in a computing-more-efficient fashion, and performs on par with existing arbitrary-scale SISR methods.
For the first time, we realize SISR tasks as not only any-scale in literature, but also as any-resource.
Our code is available at \url{https://github.com/CrispyFeSo4/AnySR}.

\end{abstract}

\begin{IEEEkeywords}
Super-Resolution \and Any-Scale \and Any-Resource
\end{IEEEkeywords}

\section{Introduction}
\label{sec:intro}
\IEEEPARstart{S}{ingle} image super-resolution (SISR) is the process of reconstructing a low-resolution (LR) image into a high-resolution (HR) one rich in detail. 
The ill-posed nature has made SISR one of the most challenging tasks in low-level computer vision.
The past decades have witnessed the advent of many classic studies that continuously address such a challenge. 
Typical works such as EDSR~\cite{Lim_2017_CVPR_Workshops}, RCAN~\cite{Zhang_2018_ECCV}, and RDN~\cite{RDNZhang_2018_CVPR}, serving as network backbones, effectively extract both texture and semantic information from LR images and lay the foundation for HR reconstruction. 
The pixel-shuffle upsampling method~\cite{Shi_2016_CVPR} takes a step further to enhance the image reconstruction quality.
Standing upon the shoulders of these pioneering works, the majority of SISR methods~\cite{Lim_2017_CVPR_Workshops, Zhang_2018_ECCV, RDNZhang_2018_CVPR} adhere to a straightforward and efficient paradigm in which features are extracted by feature extractors, followed by upscaling modules for reconstruction.

\begin{figure*}[!t] %
    \centering
    \includegraphics[width=0.9\textwidth]{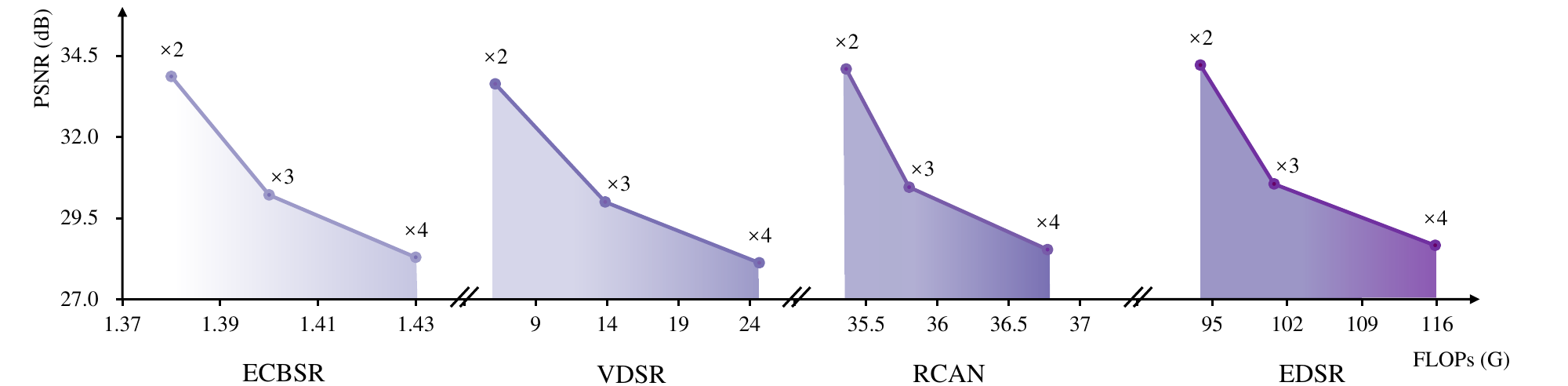}
    \vspace{-0.5em}
    \caption{PSNR performance \emph{v.s.} computing cost measured as GFLOPs for existing methods of ECBSR~\cite{zhang2021edge}, VDSR~\cite{kim2016accurate}, RCAN~\cite{Zhang_2018_ECCV} and EDSR~\cite{Lim_2017_CVPR_Workshops}. For fairness, all reported data is borrowed from \cite{wang2022adaptive} and the test set is DIV2K~\cite{agustsson2017ntire}.}
    \vspace{-1.0em}
    \label{fig:psnr&flops}
\end{figure*}

Albeit the encouraging achievement, most of the above methods are confined to addressing the image reconstruction of a fixed upsampling scale, and are therefore stuck in repeatedly rebuilding an SR network, when the fully trained model cannot accommodate the resolutions in demand.
Actually, there is a significant demand for multi-scale SR tasks. Training individual models for respective resolutions seems to be a great waste of computing resources and fails to adapt to real-time online deployment.
Reflecting on this, many recent researchers have shifted their focus to integrating any-scale tasks into only one integrated SR model.
For example, Meta-SR~\cite{hu2019meta} learns to predict coordinates and scale-oriented convolutional weights to generate HR images. 
Taking into consideration the neighborhoods in the LR image and a scale-related cell size, the subsequent LIIF~\cite{chen2021learning} utilizes an MLP to predict the pixel values of the queried coordinates in an HR one.
To enhance the representational capabilities, LTE~\cite{lte-jaewon-lee} further introduces a local texture estimator that transforms coordinate information into the Fourier domain.
The very recent SRNO~\cite{srnowei2023super} and OPE-SR~\cite{opesrsong2023ope} enable HR tasks of continuous scales by respectively incorporating Neural Operators~\cite{kovachki2021neural} and introducing an orthogonal position encoding upscale module.

With the fast development of the computer vision community, more stringent requirements on efficient and scalable vision applications have been put forward~\cite{liang2021pruning,han2021dynamic,gholami2022survey,menghani2023efficient}.
%
Delving into a reflection on most of the existing arbitrary-scale SR methods, we realize that they are not fully excavated to realize the above advocacy.
Though succeeding in tackling SR tasks across various scales in one single network, they also cause additional resource overhead, given our empirical observation that lower-resolution scales can be dealt with more easily.
To verify this, we have explicitly conducted an extensive literature review and made a summary in Fig.\,\ref{fig:psnr&flops}, outlining the performance trend \emph{w.r.t.} the SR scale and network complexity.
We can find that a network with relatively lower complexity is sufficient to ensure satisfactory performance for SR tasks with smaller reconstruction scales.
For example, when performing a $\times2$ SISR task, ECBSR~\cite{zhang2021edge} obtains 33.86 dB PSNR with only 1.38 GFLOPs consumption while EDSR~\cite{Lim_2017_CVPR_Workshops} achieves a slightly improved performance of 34.21 dB PSNR by eating up totally 93.89 GFLOPs.
However, as the scale increases a lot, a more complex network is required for the sake of a better performance. For instance, only 28.29 dB PSNR is gained by the method of ECBSR under 1.43 GFLOPs while the result increases to 28.67 dB PSNR by EDSR when more GFLOPs of 115.83 are taken to construct a $\times4$ SISR task.

As analyzed in the existing research~\cite{chen2022arm}, the platforms to perform SISR tasks are often featured with limited and varying computing resources over time, due to the potential resource occupation of other applications.
On account of the efficiency and scalability of SR applications, more efforts are actually needed to take the edge off computing expenses. 
As a consequence, we take a step back in this paper and launch the first attempt to realize image super-resolution not only at any-scale in the literature but also with any-resource, which to our best knowledge by far has never been noticed and resolved.

To this end, we propose AnySR in this paper, a general method with ``SR'' in the method's name indicating not only Super-Resolution task but also any-Scale and any-Resource implementation.
To be more concrete, we reorganize scale information into multiple sub-groups, each of which consists of similar scale sizes.
In this context, we design multiple networks with varying levels of complexity. For cases where computational resources are limited, networks with lower complexity are assigned to handle smaller-scale SISR tasks to ensure inference efficiency, while more complex networks are used for larger-scale SISR tasks to achieve better performance.
Further, to circumvent the introduction of cumbersome parameters, we train and infer all networks in a parameter-sharing manner~\cite{zhang2023realtime,yang2021mutualnet} in which the weights of smaller networks become parts of larger ones. Such a fashion also conveniences the case of abundant computational resources, where opting for the most complex network produces superior reconstruction results for all scales. Importantly, unlike traditional methods~\cite{zhang2023realtime,yang2021mutualnet} that forward a batch of samples multiple times during training, we forward each batch a time and thus do not enlarge the training costs.

Conventional practices have often employed the same feature extraction approach across images of various scales~\cite{hu2019meta,chen2021learning}. 
However, tasks at different scales demand specific adjustments to their features tailored to their respective scales, which can also, to some extent, compensate for the performance loss brought about by reduced computational resources and mutual weight influence from parameter-sharing.
For better scale information injection, we enhance any-scale feature in a feature-interweaving fashion, by repeating and inserting scale pairs into features at regular intervals. These operations decouple mutual weight influence and bring ``any-scale'' implementation closer to the performance of the original arbitrary-scale network.
Our AnySR method can be incorporated into most existing arbitrary-scale SR methods, and extensive experiments on typical test datasets have demonstrated its effectiveness.

Overall, the major contributions of this paper are three-fold: 
\begin{itemize}
    \item We present AnySR, a versatile approach designed to adapt existing SISR methods to function seamlessly across ``any-scale'' and with ``any-resources''.

    \item Any-scale enhancement through feature-interweaving fashion ensures sufficient scale information and correct feature/scale processing.
    \item Extensive experiments show that when integrated with existing arbitrary-scale SR methods, the proposed AnySR approach consistently achieves comparable performance in a more efficient manner.

\end{itemize}

\section{Related Work}
\subsection{Arbitrary-Scale Super-Resolution}
%
%
%
Over the past few years, many classic convolutional neural networks (CNNs) based methods, such as SRCNN~\cite{srcnndong2015image}, SRResNet~\cite{srresnetledig2017photo}, EDSR~\cite{Lim_2017_CVPR_Workshops}, and RDN~\cite{RDNZhang_2018_CVPR}, have been proposed and shown commendable promise in SR tasks. 
To further improve SR performance, more methods incorporate residual blocks~\cite{cavigelli2017cas,kim2016accurate,zhang2021plug}, dense blocks~\cite{wang2018esrgan,RDNZhang_2018_CVPR,zhang2020residual}, and other techniques~\cite{chen2016trainable,cheng2021mfagan,deng2021deep, 10314443, 9776607, 10140179}. 
Additionally, certain SR approaches leverage attention mechanisms, such as self-attention~\cite{liang2021swinir, 10387229}, channel attention~\cite{dai2019second,niu2020single,hatchen2023activating}, and non-local attention~\cite{liu2018non,mei2021image, 10382425}. 
Yet, most of these methods are tailored to particular scales, constraining their versatility and adaptability.

Therefore, recent researchers have turned their attention to arbitrary-scale SR tasks, aiming to tackle SR problems with arbitrary scales within a unified model. 
For example, Meta-SR~\cite{hu2019meta} innovatively introduces the first arbitrary-scale meta-upscale module, predicting convolutional filters based on coordinates and scales to generate HR images.
Subsequently, employing implicit neural representation, LIIF~\cite{chen2021learning} predicts the RGB value at an arbitrary query coordinate by incorporating image coordinates and features from the backbone around that point. 
Further enhancements have been achieved by techniques like LTE~\cite{lte-jaewon-lee}, which introduces a local texture estimator characterizing image textures in Fourier space.
Additionally, recent advancements in the SR field have introduced novel structures. 
For instance, SRNO~\cite{srnowei2023super} incorporates Neural Operators, and OPE-SR~\cite{opesrsong2023ope} introduces orthogonal position encoding (OPE), an extension of position encoding. 
These innovations have demonstrated substantial performance improvements.

\subsection{Efficient Image Super-Resolution}
In recent years, numerous approaches~\cite{fsrcnndong2016accelerating,carnahn2018fast,rfdnliu2020residual,fmendu2022fast, 10472422} have emerged to develop efficient super-resolution networks. 
Building upon the pioneering work of SRCNN~\cite{srcnndong2015image}, FSRCNN~\cite{fsrcnndong2016accelerating} substantially accelerates the SISR networks. 
This acceleration is achieved by taking the original LR image as input without bicubic interpolation, using smaller sizes of convolution kernels, and incorporating a deconvolutional layer at the final stage of the network for upsampling.
LapSRN~\cite{lapsrnlai2017deep} incrementally reconstructs the sub-band residuals of HR images through the utilization of the Laplacian pyramid.
CARN~\cite{carnahn2018fast} achieves improved efficiency by ingeniously designing cascading residual networks with group convolution.
IMDN~\cite{imdnhui2019lightweight} presents information multi-distillation blocks featuring a contrast-aware channel attention layer.
In parallel, RFDN~\cite{rfdnliu2020residual} optimizes the architecture using a feature distillation mechanism through the proposed residual feature distillation block.
In the wake of RFDN, RLFN~\cite{rlfnkong2022residual} adds more channels to compensate for discarded feature distillation branches and introduces the residual local feature block, resulting in enhanced inference speed and superior performance with fewer parameters. 
On the other hand, FMEN~\cite{fmendu2022fast} introduces the re-parameterization technique, expanding the optimization space during training through re-parameterizable building blocks~\cite{ding2021repvgg}, without incurring additional inference time.
HAT~\cite{hatchen2023activating} proposes same-task pre-training, which directly performs pre-training on a larger-scale dataset such as ImageNet~\cite{deng2009imagenet} based on the same task. Although it requires careful control of the pre-training iteration count and the use of an appropriately small learning rate, this approach is straightforward and leads to significant performance improvements.

\begin{figure*}[!t]
    \centering
    \includegraphics[width=0.9\textwidth, height = 0.37\textwidth]{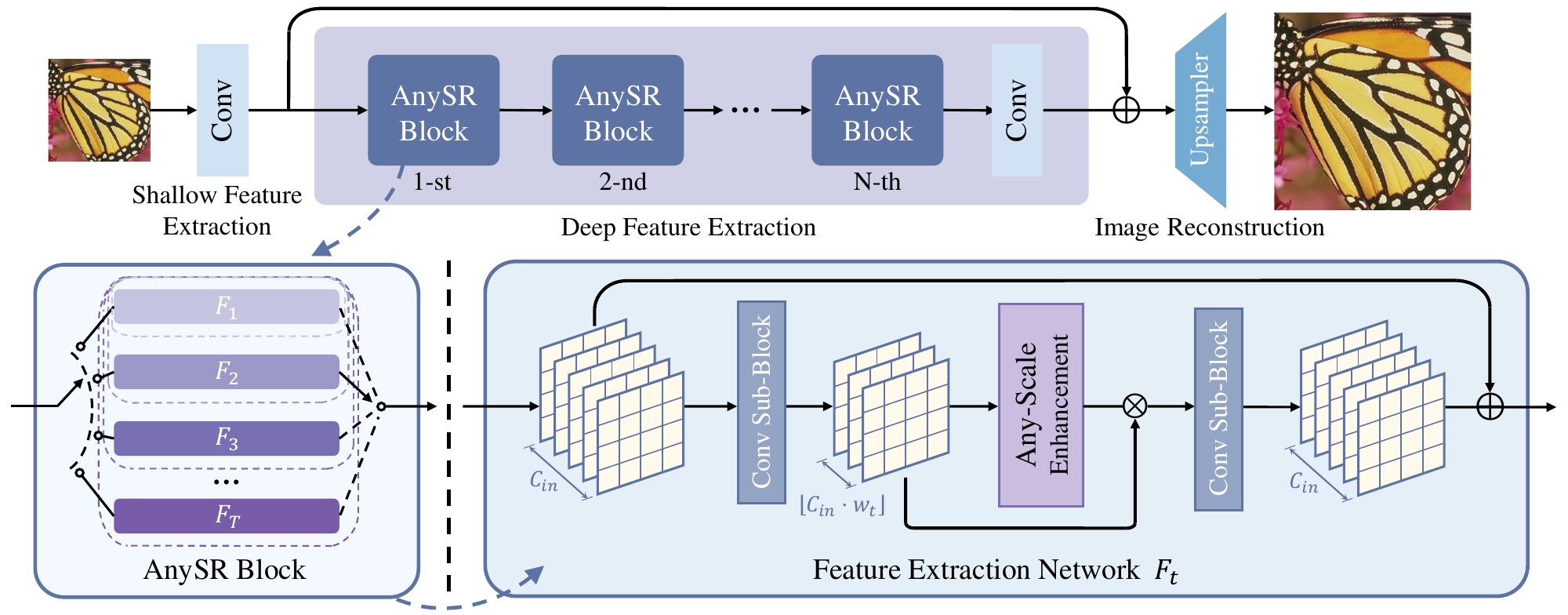}
    \vspace{-0.5em}
    \caption{
    Overview of the proposed AnySR: Includes Shallow Feature Extraction, Deep Feature Extraction with AnySR Blocks, and Image Reconstruction. 
    AnySR Block automatically selects sub-net $F_t$ based on task complexity for feature extraction and enhancement.
    }
    \vspace{-1.0em}
    \label{fig:overview}
\end{figure*}

\section{Methodology}
\label{sec:methodology}

\subsection{Preliminaries}

The general pipeline of our AnySR method acts in accordance with most existing methods~\cite{Lim_2017_CVPR_Workshops,wang2021learning,hatchen2023activating} as in the upper half of Fig.\,\ref{fig:overview}, typically highlighted with
1) A 3$\times$3 convolutional layer to convert a given LR image $I_\mathrm{LR} \in \mathbb{R}^{H \times W \times 3}$ into a shallow feature $f_s \in \mathbb{R}^{H \times W \times C_\mathrm{in}}$ where $C_\mathrm{in}$ represents the channel number.
2) $N$ consecutive blocks for deep feature extraction, standing out as the ``AnySR'' block in this paper, along with a common convolutional layer and a global residual connection to the shadow feature $f_s$.
3) An upsampler $U(\cdot)$ to reconstruct an HR version of the LR image, denoted as $I'_\mathrm{HR} \in \mathbb{R}^{H' \times W' \times 3}$ where $H' > H$ and $W' > W$. 
For ease of the following representation, we use $F(\cdot)$ to denote feature extraction operations in 1) and 2).

Given an LR image $I_\mathrm{LR}$, the process of constructing its HR version $I'_\mathrm{HR}$ under the arbitrary-scale setting~\cite{hu2019meta,chen2021learning,wang2021learning} can be formulated as
\begin{equation}\label{lrtohr}
    I'_\mathrm{HR} = U\big(F(I_\mathrm{LR}, S;\Theta_F);\Theta_U\big), 
\end{equation}
where $\Theta_F$ denotes the learnable parameters of feature extractor $F(\cdot)$, taking as inputs LR image $I_\mathrm{LR}$ and upsampling scale set $S = \{(s^h_i, s^w_i)\}_{i=1}^n$, $\Theta_U$ denotes the learnable parameters of upsampler $U(\cdot)$ taking the outputs of $F(\cdot)$ as inputs.
%

As analyzed in Sec.\,\ref{sec:intro}, such an arbitrary-scale pipeline constrains the efficiency and scalability of applying SR models. A less complex network can sometimes adequately handle a smaller reconstruction scale. Consequently, allocating a uniform inference overhead for all scales manifests insufficient scalability to accommodate running platforms of time-varying computational resources, and presents poor efficiency when dealing with smaller reconstruction scales.
Contemplating this, in Sec.\,\ref{sec:anyresource}, we rebuild current methods as an implementation of ``any-resource'', to enable the selection of a smaller network in the SR model based on the abundance of computing resources.
Then, we continue proposing ``feature-interweaving'' in Sec.\,\ref{sec:anyscale} to enhance the performance under our ``any-resource'' implementation.

\begin{figure*}[!t]
    \centering
    \includegraphics[width=\textwidth]{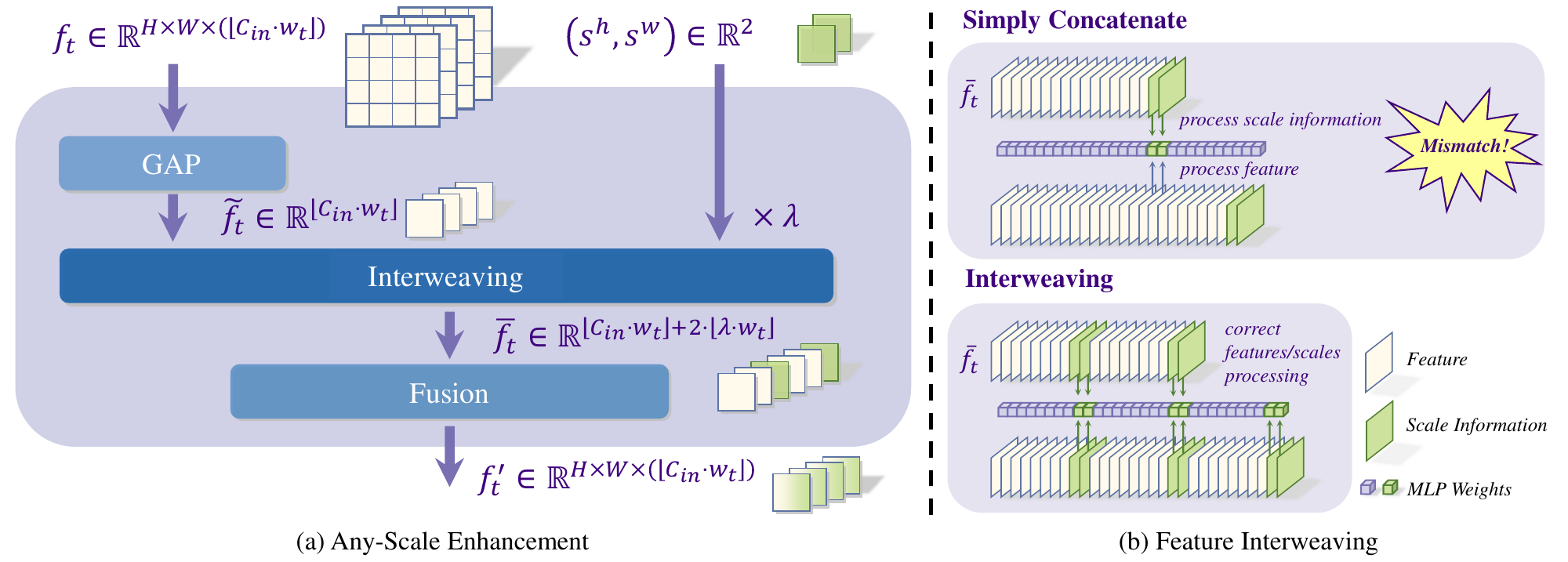}
    \caption{Framework of our any-scale enhancement including (a) any-scale enhancement pipeline and (b) feature-interweaving illustration.} 
    \label{fig:asm}
\end{figure*}

\begin{algorithm}[!t]
	\renewcommand{\algorithmicrequire}{\textbf{Input:}}
	\renewcommand{\algorithmicensure}{\textbf{Output:}}
	\caption{AnySR Training.}
	\label{alg:training}
	\begin{algorithmic}[1]
		\REQUIRE LR image $I_\mathrm{LR}$, HR image $I_\mathrm{HR}$, sorted and split scale set $S = S_1 \cup S_2 \cup ... \cup S_T$, a pre-trained arbitrary-scale SR model $F$ with weights $\Theta_F$, training step $K$.
     \FOR{$k = 1 \rightarrow K$}
        \STATE $t \sim [1, 2, ..., T]$;
        \STATE $(s^h, s^w) \sim S_t$;
        \STATE $t \leftarrow T$ with probability $p$;
        \STATE Forward $I_\mathrm{LR}$ and $(s^h, s^w)$ to $F$ with weights $\Theta_F[1:|\Theta_{F_t}|]$;
        \STATE Obtain reconstructed HR image $I'_\mathrm{HR}$ via Eq.\,(\ref{eq:final});
        \STATE $F\gets F - \beta\nabla \parallel  I_\mathrm{HR} - I'_\mathrm{HR}  \parallel_1$.
        \hspace{0.01cm} {\scriptsize \%  $\Theta_F[1:|\Theta_{F_t}|]$ are updated}.
      \ENDFOR
    \RETURN $F$
\end{algorithmic}  
\end{algorithm}

\subsection{Any-Resource Implementation}
\label{sec:anyresource}

The upper part of Fig.\,\ref{fig:overview} manifests our any-resource implementation. 
Considering the close relationship between scaling factor $(s_i^h, s_i^w)$ and reconstruction difficulty, we rearrange scale set $S$ in an ascending order and partition it into $T$ groups: 
\begin{equation}
S_1 \cup S_2 \cup ... \cup S_T = S \quad and \quad S_1 \cap S_2 \cap ... \cap S_T = \emptyset.
\end{equation}

Correspondingly, $T$ distinct feature extraction networks $\{F_t\}_{t=1}^T$ with various complexities are then constructed. Each network $F_t$ is affiliated with parameters of a particular size as $\Theta_{F_t}$, indicating the complexity. We have $|\Theta_{F_{t+1}}| > |\Theta_{F_t}|$.
A natural approach is to utilize network $F_t$ to process the SR tasks within group $S_t$ such that
1) Smaller-scale tasks can be efficiently fulfilled through a structurally simple network; 2) Larger-scale ones benefit from complex networks. 
Thus, the process of reconstructing group $S_t$ tasks becomes
\begin{equation}
    I'_\mathrm{HR}=U\big(F_t(I_\mathrm{LR}, S_t;\Theta_{F_t});\Theta_U\big).
\end{equation}

\textbf{All-in-One Training.} Although the above process greatly conserves computational resources, it defects in that
1) $T$ networks have to be repeatedly trained like traditional SR methods, and that
2) additional parameters are introduced compared with arbitrary-scale SR methods.
To solve this issue, we follow~\cite{zhang2023realtime,yang2021mutualnet} to train and infer all networks in a parameter-sharing manner by the following set of constraints:
\begin{equation}
\begin{aligned}
    \Theta_{F_1} &= \Theta_F[1:|\Theta_{F_1}|], \\
    \Theta_{F_2} &= \Theta_F[1:|\Theta_{F_2}|], \\
    & \vdots \\
    \Theta_{F_T} &= \Theta_F[1:|\Theta_{F_T}|] = \Theta_F,
\end{aligned}
\end{equation}
where $\Theta_F[1:|\Theta_{F_t}|]$ represents the first $|\Theta_{F_t}|$ filters of $\Theta_F$, making $F_t$ a subnet of $F$ as $\Theta_{F_1} \subset \Theta_{F_2} \subset ... \subset \Theta_{F_T} \subseteq \Theta_{F}$. 

Consequently, the process of our AnySR to reconstruct multiple scales can be unified as follows: 
\begin{equation}\label{eq:final}
    I'_\mathrm{HR}=U\big(F(I_\mathrm{LR}, S_t; \Theta_F[1:|\Theta_{F_t}|])  ;\Theta _U\big).
\end{equation}

Algorithm\,\ref{alg:training} shows how to train AnySR. 
At each training iteration, we randomly forward a subnet of the original model $F$ and update only the corresponding weights $\Theta_F[1:|\Theta_{F_t}|]$. 
It realizes multiple networks of different resources in one training, and introduces no additional parameters.
In Line 4 of Algorithm\,\ref{alg:training}, the subnet is probably reset to $F$ for the consideration of retaining the original ability of $F$, which performs multiple scale tasks through the entire weights $\Theta_F$.
Note that, we forward each batch once, a distinction from existing weight-sharing methods~\cite{zhang2023realtime,yang2021mutualnet} that perform multiple forwards, for the consideration of 1) better training efficiency, and 2) already fully-pretrained weights $\Theta_{F}$.

\subsection{Any-Scale Enhancement}
\label{sec:anyscale}

Though any-resource implementation boosts the scalability and efficiency of SR tasks, its parameter-sharing puts the performance at risk, due to fewer parameters for smaller scales and mutual weight influence among different scales. It is of great necessity to enhance the reconstruction results of smaller networks and bring them closer to the performance of the original arbitrary-scale network.
Previous research~\cite{wang2021learning} has proven beneficial from linking extracted features with scales, ultimately optimizing the performance of the upsampling module.
We realize scale information can be well excavated and ameliorated under our any-resource implementation for the following reasons: 1) The weights $\Theta_F[1:|\Theta_{F_t}|]$ are particularly trained to deal with the scale set $S_t$, leading to scale-aware features; 2) Partial weights are shared among different subnets, injecting information of other scales.
Therefore, we achieve any-scale enhancement by emphasizing the significance of customized handling for features across different scales. 
Specifically, as shown in Fig.\,\ref{fig:overview}, inside each ``AnySR'' block lies two sub-blocks with the first/second reducing/increasing the channels. 
We choose to inject better scale information, in a feature-interweaving fashion as illustrated in Fig.\,\ref{fig:asm}(a), into the outputs of the first sub-block, denoted as $f_t \in \mathbb{R}^{H \times W \times (\lfloor C_\mathrm{in} \cdot w_t \rfloor)}$, where $w_t = \frac{|\Theta_{F_t}|}{|\Theta_F|} $, and $\lfloor \cdot \rfloor$ denotes the floor function.

\textbf{Feature-Interweaving.}
Features $f_t \in \mathbb{R}^{H \times W \times ( \lfloor C_{in} \cdot w_t \rfloor )}$ from $F_t$ would first go through a global average pooling (GAP) to get a variant $\tilde{f}_t \in \mathbb{R}^{ \lfloor C_\mathrm{in} \cdot w_t \rfloor }$ where the scale information $(s^h, s^w) \in S_t$ is formally injected.

To this end, one naive fashion, as illustrated in the upper half of Fig.\,\ref{fig:asm}(b), is to follow existing methods~\cite{chen2021learning,cao2023ciaosr} which simply concatenate $(s^h, s^w)$ at the rear of $\tilde{f}_t$ to form $\bar{f}_t = [\tilde{f}_t, s^h, s^w] \in \mathbb{R}^{ \lfloor C_\mathrm{in}\cdot w_t \rfloor + 2}$. Then, a two-layer MLP, with weights $W_1 \in \mathbb{R}^{2\cdot C_\mathrm{in} \times (C_\mathrm{in} + 2)}$, $W_2 \in \mathbb{R}^{  C_\mathrm{in}  \times 2\cdot C_\mathrm{in}}$ and a ReLU layer inserted between, is created, followed by a Sigmoid function to weight the original features $f_t$ as
\begin{equation}\label{eq:weighted}
\begin{aligned}
    f_t' &= f_t \odot \mathrm{Sigmoid}\big(W_2[1: \lfloor C_\mathrm{in}\cdot w_t \rfloor, :] \cdot \\
    &\quad \mathrm{ReLU}(W_1[:,1: \lfloor C_\mathrm{in}\cdot w_t  \rfloor + 2] \cdot \bar{f}_t)\big),
\end{aligned}
\end{equation}
where $W_1[:,1: \lfloor C_\mathrm{in}\cdot w_t \rfloor + 2] \in \mathbb{R}^{2\cdot C_\mathrm{in} \times (\lfloor C_\mathrm{in} \cdot w_t \rfloor) + 2}$ and $W_2[1:  \lfloor  C_\mathrm{in} \cdot w_t  \rfloor, :] \in \mathbb{R}^{ \lfloor C_\mathrm{in} \cdot w_t  \rfloor  \times 2\cdot C_\mathrm{in}}$ are the shared MLP weights for the network $F_t$.
 $f_t'$ then serves as the input of the second sub-block in the ``AnySR'' block.

Though this naive approach facilitates the interaction between features and scale information, two notable limitations, as we analyze, arise:
1) Insufficient scale information. Compared to a total of $\lfloor C_\mathrm{in} \cdot w_t \rfloor$ channels in image features, where $\lfloor C_\mathrm{in} \cdot w_t \rfloor \gg 2$ signifies a limited influence of scale information on the weighted features, the scale information takes up only 2 dimensions.
2) Inappropriate scale processing. For the MLP weights $W_1 \in \mathbb{R}^{2 \cdot C_\mathrm{in} \times (C_\mathrm{in} + 2)}$, the sub-weights $W_1[: , \lfloor C_\mathrm{in} \cdot w_t \rfloor +1:  \lfloor C_\mathrm{in} \cdot w_t \rfloor + 2] \in \mathbb{R}^{2 \cdot C_\mathrm{in} \times 2}$ are used to process the scale information $(s^h, s^w) \in S_t$. 
For $(s^h, s^w) \in S_{t+1}$, the weights to process scales are $W_1[: , \lfloor C_\mathrm{in} \cdot w_{t+1}  \rfloor +1:  \lfloor C_\mathrm{in} \cdot w_{t+1} \rfloor + 2] \in \mathbb{R}^{2 \cdot C_\mathrm{in} \times 2}$ and $W_1[: ,  \lfloor C_\mathrm{in} \cdot w_t  \rfloor  +1:  \lfloor C_\mathrm{in} \cdot w_t  \rfloor + 2] \in \mathbb{R}^{2 \cdot C_\mathrm{in} \times 2}$ are now responsible for processing image features.
If off-the-shelf methods~\cite{chen2021learning,cao2023ciaosr} are still applied here, as a result, they may weaken the impact of scale information and thus fail to well excavate customized features at different scales.
Experimental demonstration has been provided in Table\,\ref{tab:ablation1-psnr}.

To better align features with scale information, for network $F_t$, our feature-interweaving inserts scale pair $(s^h, s^w)$ for $\lambda$ times into positions $\big(  \lfloor C_\mathrm{in}\cdot i/{\lambda} \rfloor + 2i -1,   \lfloor C_\mathrm{in} \cdot i/{\lambda} \rfloor  + 2i      \big)$ where $i=1, 2, ..., \lfloor \lambda \cdot w_t  \rfloor$, of the pooled features $\tilde{f}_t \in \mathbb{R}^{ \lfloor C_\mathrm{in} \cdot w_t  \rfloor}$, which leads to the following concatenated features $\bar{f}_t \in \mathbb{R}^{ \lfloor C_\mathrm{in} \cdot w_t \rfloor  + 2\cdot \lfloor \lambda \cdot w_t  \rfloor}$:
\begin{equation}
\begin{split}
    \bar{f}_t = \Big[ &\tilde{f}_t\big[1: \lfloor C_\mathrm{in} \cdot 1/{\lambda} \rfloor \big], s^h, s^w, \\
    &\tilde{f}_t\big[ \lfloor C_\mathrm{in} \cdot 1/{\lambda}  \rfloor +1 :  \lfloor C_\mathrm{in} \cdot 2/{\lambda} \rfloor \big], s^h, s^w, \\
    &\dots, \\
    &\tilde{f}_t\big[  \lfloor C_\mathrm{in}\cdot (i-1)/{\lambda}  \rfloor +1:  \lfloor C_\mathrm{in}\cdot {i}/{\lambda}  \rfloor \big], s^h, s^w, \dots, \\
    &\tilde{f}_t\big[  \lfloor C_\mathrm{in}\cdot  \lfloor \lambda \cdot w_t \rfloor/{\lambda}  \rfloor +1 : \big], s^h, s^w\Big].
\end{split}
\end{equation}
Notably, the scale pair $(s^h, s^w)$ is not always appended at the end of $\bar{f}_t$. 
For Convenience, we define the fixed interval between these insertions as $C_{fix} = C_{in} / \lambda $, and refer to the last portion of the feature $\bar{f}_t$ as $C_{last}$. 
When $C_{last} < C_{fix}$, the scale pair is not appended, ensuring correct feature and scale processing. Conversely, when $C_{last} = C_{fix}$, the scale pair is appended to the end of the feature.

Then, weights $W_1$ in the two-layer MLP become $W_1 \in \mathbb{R}^{2 \cdot C_\mathrm{in} \times (C_\mathrm{in} + 2 \cdot \lambda)}$, and the weighted features in Eq.\,(\ref{eq:weighted}) is:
\begin{equation}\label{eq:weighted2}
\begin{split}
    f_t' = &f_t \odot \mathrm{Sigmoid}\big( W_2\big[1: \lfloor C_\mathrm{in}\cdot w_t  \rfloor, :\big] \cdot \\
    &\mathrm{ReLU}(W_1\big[:,1: \lfloor C_\mathrm{in}\cdot w_t  \rfloor + 2\cdot \lfloor \lambda \cdot w_t  \rfloor  \big] \cdot \bar{f}_t)\big).
\end{split}
\end{equation}

As a consequence, the weights $W_1\big[:, \lfloor C_\mathrm{in}\cdot{i}/{\lambda}  \rfloor + 2i -1: \lfloor C_\mathrm{in} \cdot {i}/{\lambda} \rfloor + 2i\big] \in \mathbb{R}^{2\cdot C_\mathrm{in} \times 2}$, in which $i = 1, 2, ...,  \lfloor \lambda \cdot w_t \rfloor$ consistently deal with the scale information whatever the specific value of $t$.
In conclusion, our feature-interweaving overcomes the issues of traditional methods~\cite{chen2021learning,cao2023ciaosr} by taking into account two key operations:
1) repeating the scale pair $(s^h, s^w)$ $\lambda$ times to ensure sufficient scale information for network $F_t$;
2) inserting scale pairs into features in regular intervals to ensure the correct feature/scale processing.
The efficacy of our feature-interweaving will be shown in Sec.\,\ref{sec:experimentation}.

\begin{table*}[!t]
\caption{PSNR (dB) and LPIPS comparison of existing arbitrary-scale SR methods and their AnySR variants, highlighted by $^\dag$ (through different subnets) and $^\ddag$ (AnySR-retrained version via the largest network). $^\ast$ denotes our re-implementation. Following~\cite{opesrsong2023ope}, AnySR performance is highlighted in {\color{blue}blue} if within 0.15dB (PSNR) / 0.01 (LPIPS) of the vanilla method, and {\color{red}red} if improved.}
\vspace{-2mm}
\label{tab:psnr}
\centering
\setlength\tabcolsep{7.2pt}
\resizebox{\textwidth}{!}{
\renewcommand{\arraystretch}{1.5}
\begin{tabular}{c|c|ccc|ccc|ccc|ccc|ccc}
\toprule

\multirow{2}*{\begin{tabular}[c]{@{}c@{}}Backbone\\ Networks\end{tabular}} & \multirow{2}*{Methods} 
& \multicolumn{3}{c|}{Set5~\cite{bevilacqua2012low}}  
& \multicolumn{3}{c|}{Set14~\cite{zeyde2012single}}                          & \multicolumn{3}{c|}{B100~\cite{martin2001database}}                        & \multicolumn{3}{c|}{Urban100~\cite{huang2015single}}                       & \multicolumn{3}{c}{Manga109~\cite{aizawa2020building}}                      \\ \cline{3-17} 
&       
& \multicolumn{1}{c|}{$\times$2} 
& \multicolumn{1}{c|}{$\times$3} 
& $\times$4 
& \multicolumn{1}{c|}{$\times$2} 
& \multicolumn{1}{c|}{$\times$3} 
& $\times$4 
& \multicolumn{1}{c|}{$\times$2} 
& \multicolumn{1}{c|}{$\times$3} 
& $\times$4 
& \multicolumn{1}{c|}{$\times$2} 
& \multicolumn{1}{c|}{$\times$3} 
& $\times$4 
& \multicolumn{1}{c|}{$\times$2} 
& \multicolumn{1}{c|}{$\times$3} 
& $\times$4 \\ 
\hline

\multirow{26}*{EDSR}  
& Metric
& \multicolumn{15}{c}{PSNR$\uparrow$}
\\ \cline{2-17}

& MetaSR$^\ast $                   
& \multicolumn{1}{c|}{37.89}   & \multicolumn{1}{c|}{34.35}   & 32.08   
& \multicolumn{1}{c|}{33.57}   & \multicolumn{1}{c|}{30.29}   & 28.52   
& \multicolumn{1}{c|}{32.14}   & \multicolumn{1}{c|}{29.07}   & 27.54   
& \multicolumn{1}{c|}{32.07}   & \multicolumn{1}{c|}{28.10}   & 25.97   
& \multicolumn{1}{c|}{38.36}   & \multicolumn{1}{c|}{33.39}   & 30.37   \\ \cline{2-17}

& MetaSR$^\dag$            
& \multicolumn{1}{c|}{{\color{blue}37.85}}   
& \multicolumn{1}{c|}{{\color{red}34.40}}   
& {\color{red}32.17}   
& \multicolumn{1}{c|}{{\color{blue}33.44}}   
& \multicolumn{1}{c|}{{\color{blue}30.28}}   
& {\color{red}28.55}   
& \multicolumn{1}{c|}{{\color{blue}32.07}}   
& \multicolumn{1}{c|}{{\color{blue}29.06}}   
& {\color{red}27.56}   
& \multicolumn{1}{c|}{31.84}   
& \multicolumn{1}{c|}{{\color{blue}28.09}}   
& {\color{red}26.05}   
& \multicolumn{1}{c|}{{\color{blue}38.30}}   
& \multicolumn{1}{c|}{{\color{red}33.47}}   
& {\color{red}30.46}   \\ \cline{2-17} 

& MetaSR$^\ddag$            
& \multicolumn{1}{c|}{{\color{blue}37.89}}   & \multicolumn{1}{c|}{{\color{red}34.40}}   & {\color{red}32.17}   & \multicolumn{1}{c|}{{\color{red}33.58}}   & \multicolumn{1}{c|}{{\color{red}30.32}}   & {\color{red}28.55}   & \multicolumn{1}{c|}{{\color{blue}32.14}}   & \multicolumn{1}{c|}{{\color{red}29.08}}   & {\color{red}27.56}   & \multicolumn{1}{c|}{{\color{red}32.16}}   & \multicolumn{1}{c|}{{\color{red}28.19}}   & {\color{red}26.05}   & \multicolumn{1}{c|}{{\color{red}38.41}}   & \multicolumn{1}{c|}{{\color{red}33.50}}   & {\color{red}30.46}   \\ \cline{2-17} 

& LIIF                     & \multicolumn{1}{c|}{37.99}   & \multicolumn{1}{c|}{34.40}   & 32.24   & \multicolumn{1}{c|}{33.66}   & \multicolumn{1}{c|}{30.34}   & 28.62   & \multicolumn{1}{c|}{32.17}   & \multicolumn{1}{c|}{29.10}   & 27.60   & \multicolumn{1}{c|}{32.15}   & \multicolumn{1}{c|}{28.22}   & 26.15   & \multicolumn{1}{c|}{36.19}   & \multicolumn{1}{c|}{31.47}   & 28.81   \\ \cline{2-17} 
& LIIF$^\dag$              & \multicolumn{1}{c|}{{\color{blue}37.88}}   & \multicolumn{1}{c|}{{\color{blue}34.34}}   & {\color{blue}32.21}   & \multicolumn{1}{c|}{{\color{blue}33.53}}   & \multicolumn{1}{c|}{{\color{blue}30.32}}   & {\color{blue}28.61}   & \multicolumn{1}{c|}{{\color{blue}32.10}}   & \multicolumn{1}{c|}{{\color{blue}29.08}}   & {\color{blue}27.59}   & \multicolumn{1}{c|}{31.90}   & \multicolumn{1}{c|}{{\color{blue}28.21}}   & {\color{red}26.19}   & \multicolumn{1}{c|}{35.99}   & \multicolumn{1}{c|}{{\color{blue}31.45}}   & {\color{blue}28.80}   \\ \cline{2-17}  

& LIIF$^\ddag$              & \multicolumn{1}{c|}{{\color{blue}37.98}}   & \multicolumn{1}{c|}{{\color{blue}34.37}}   & {\color{blue}32.21}   & \multicolumn{1}{c|}{{\color{blue}33.62}}   & \multicolumn{1}{c|}{{\color{red}30.35}}   & {\color{blue}28.61}   & \multicolumn{1}{c|}{{\color{blue}32.15}}   & \multicolumn{1}{c|}{{\color{blue}29.09}}   & {\color{blue}27.59}   & \multicolumn{1}{c|}{{\color{blue}32.11}}   & \multicolumn{1}{c|}{{\color{red}28.25}}   & {\color{red}26.19}   & \multicolumn{1}{c|}{{\color{red}36.26}}   & \multicolumn{1}{c|}{{\color{red}31.53}}   & {\color{blue}28.80}   \\ \cline{2-17} 

& ArbSR$^\ast $                   & \multicolumn{1}{c|}{37.97}   & \multicolumn{1}{c|}{34.39}   & 32.07   & \multicolumn{1}{c|}{33.70}   & \multicolumn{1}{c|}{30.31}   & 28.56   & \multicolumn{1}{c|}{32.19}   & \multicolumn{1}{c|}{29.11}   & 27.55   & \multicolumn{1}{c|}{32.18}   & \multicolumn{1}{c|}{28.13}   & 25.96   & \multicolumn{1}{c|}{38.52}   & \multicolumn{1}{c|}{33.63}   & 30.46   \\ \cline{2-17} 
%
& ArbSR$^\dag$         & \multicolumn{1}{c|}{{\color{blue}37.87}}   & \multicolumn{1}{c|}{{\color{blue}34.34}}   & {\color{blue}32.05}   & \multicolumn{1}{c|}{33.46}   & \multicolumn{1}{c|}{{\color{blue}30.28}}   & {\color{blue}28.53}   & \multicolumn{1}{c|}{{\color{blue}32.10}}   & \multicolumn{1}{c|}{{\color{blue}29.07}}   &  {\color{blue}27.53}  & \multicolumn{1}{c|}{31.73}   & \multicolumn{1}{c|}{{\color{blue}27.99}}   & {\color{blue}25.92}   & \multicolumn{1}{c|}{38.33}   & \multicolumn{1}{c|}{{\color{blue}33.53}}   & {\color{blue}30.35}   \\ \cline{2-17} 

& ArbSR$^\ddag$         & \multicolumn{1}{c|}{{\color{blue}37.93}}   & \multicolumn{1}{c|}{{\color{blue}34.35}}   &  {\color{blue}32.05}  & \multicolumn{1}{c|}{{\color{blue}33.65}}   & \multicolumn{1}{c|}{{\color{blue}30.30}}   &  {\color{blue}28.53}  & \multicolumn{1}{c|}{{\color{blue}32.17}}   & \multicolumn{1}{c|}{{\color{blue}29.08}}   & {\color{blue}27.53}   & \multicolumn{1}{c|}{{\color{blue}32.13}}   & \multicolumn{1}{c|}{{\color{blue}28.07}}   & {\color{blue}25.92}   & \multicolumn{1}{c|}{{\color{blue}38.44}}   & \multicolumn{1}{c|}{{\color{blue}33.60}}   & {\color{blue}30.35}   \\ \cline{2-17}

& SRNO                  & \multicolumn{1}{c|}{38.15}   & \multicolumn{1}{c|}{34.53}   & 32.39   & \multicolumn{1}{c|}{33.83}   & \multicolumn{1}{c|}{30.50}   & 28.79   & \multicolumn{1}{c|}{32.27}   & \multicolumn{1}{c|}{29.29}   & 27.67   & \multicolumn{1}{c|}{32.63}   & \multicolumn{1}{c|}{28.58}   & 26.50   & \multicolumn{1}{c|}{39.01}   & \multicolumn{1}{c|}{33.91}   & 30.88   \\ \cline{2-17} 
%
& SRNO$^\dag$      & \multicolumn{1}{c|}{{\color{blue}38.04}}   & \multicolumn{1}{c|}{{\color{blue}34.50}}   & {\color{blue}32.38}   & \multicolumn{1}{c|}{{\color{blue}33.74}}   & \multicolumn{1}{c|}{{\color{blue}30.47}}   & {\color{blue}28.78}   & \multicolumn{1}{c|}{{\color{blue}32.19}}   & \multicolumn{1}{c|}{{\color{blue}29.15}}   & {\color{blue}27.64}   & \multicolumn{1}{c|}{32.27}   & \multicolumn{1}{c|}{{\color{blue}28.45}}   & {\color{blue}26.45}   & \multicolumn{1}{c|}{38.78}   & \multicolumn{1}{c|}{{\color{blue}33.77}}   & {\color{blue}30.83}  
  \\ \cline{2-17} 

& SRNO$^\ddag$      & \multicolumn{1}{c|}{{\color{blue}38.12}}   & \multicolumn{1}{c|}{{\color{blue}34.51}}   & {\color{blue}32.38}   & \multicolumn{1}{c|}{{\color{blue}33.81}}   & \multicolumn{1}{c|}{{\color{blue}30.49}}   & {\color{blue}28.78}   & \multicolumn{1}{c|}{{\color{blue}32.25}}   & \multicolumn{1}{c|}{{\color{blue}29.17}}   & {\color{blue}27.64}   & \multicolumn{1}{c|}{{\color{blue}32.52}}   & \multicolumn{1}{c|}{{\color{blue}28.50}}   & {\color{blue}26.45}   & \multicolumn{1}{c|}{{\color{blue}38.96}}   & \multicolumn{1}{c|}{{\color{blue}33.83}}   & {\color{blue}30.83}   \\ 

\cline{2-17}
&  Metric
 & \multicolumn{15}{c}{LPIPS$\downarrow$}
\\
\cline{2-17}

& MetaSR$^\ast $                  & \multicolumn{1}{c|}{0.022}                              & \multicolumn{1}{c|}{0.054}                        & 0.078                        & \multicolumn{1}{c|}{0.040}                        & \multicolumn{1}{c|}{0.089}                        & 0.126                        & \multicolumn{1}{c|}{0.055}                        & \multicolumn{1}{c|}{0.102}                        & 0.139                        & \multicolumn{1}{c|}{0.044}                        & \multicolumn{1}{c|}{0.100}                        & 0.147                        & \multicolumn{1}{c|}{0.012}                        & \multicolumn{1}{c|}{0.036}                        & 0.062                        \\ \cline{2-17}

& MetaSR$^\dag$           & \multicolumn{1}{c|}{{\color[HTML]{1600FF} 0.023}}       & \multicolumn{1}{c|}{{\color[HTML]{FF0000} 0.053}} & {\color[HTML]{FF0000} 0.076} & \multicolumn{1}{c|}{{\color[HTML]{1600FF} 0.041}} & \multicolumn{1}{c|}{{\color[HTML]{FF0000} 0.087}} & {\color[HTML]{FF0000} 0.125} & \multicolumn{1}{c|}{{\color[HTML]{1600FF} 0.055}} & \multicolumn{1}{c|}{{\color[HTML]{FF0000} 0.101}} & {\color[HTML]{FF0000} 0.137} & \multicolumn{1}{c|}{{\color[HTML]{1600FF} 0.044}} & \multicolumn{1}{c|}{{\color[HTML]{FF0000} 0.098}} & {\color[HTML]{FF0000} 0.144} & \multicolumn{1}{c|}{{\color[HTML]{1600FF} 0.012}} & \multicolumn{1}{c|}{{\color[HTML]{FF0000} 0.034}} & {\color[HTML]{FF0000} 0.059} \\ \cline{2-17}

& MetaSR$^\ddag$            
 & \multicolumn{1}{c|}{{\color[HTML]{1600FF} 0.022}}       & \multicolumn{1}{c|}{{\color[HTML]{FF0000} 0.053}} & {\color[HTML]{FF0000} 0.076} & \multicolumn{1}{c|}{{\color[HTML]{FF0000} 0.040}} & \multicolumn{1}{c|}{{\color[HTML]{FF0000} 0.087}} & {\color[HTML]{FF0000} 0.125} & \multicolumn{1}{c|}{{\color[HTML]{1600FF} 0.055}} & \multicolumn{1}{c|}{{\color[HTML]{FF0000} 0.101}} & {\color[HTML]{FF0000} 0.137} & \multicolumn{1}{c|}{{\color[HTML]{FF0000} 0.042}} & \multicolumn{1}{c|}{{\color[HTML]{FF0000} 0.096}} & {\color[HTML]{FF0000} 0.144} & \multicolumn{1}{c|}{{\color[HTML]{1600FF} 0.012}} & \multicolumn{1}{c|}{{\color[HTML]{FF0000} 0.034}} & {\color[HTML]{FF0000} 0.059}   \\ \cline{2-17} 

& LIIF                     
& \multicolumn{1}{c|}{0.040}                              
& \multicolumn{1}{c|}{0.093}                        
& 0.135                        
& \multicolumn{1}{c|}{0.071}                        
& \multicolumn{1}{c|}{0.164}                        
& 0.231                        
& \multicolumn{1}{c|}{0.113}                        
& \multicolumn{1}{c|}{0.231}                        
& 0.310                        
& \multicolumn{1}{c|}{0.049}                        
& \multicolumn{1}{c|}{0.124}                        
& 0.187                        
& \multicolumn{1}{c|}{0.074}                        
& \multicolumn{1}{c|}{0.104}                        
& 0.136                        
\\ \cline{2-17}

& LIIF$^\dag$             
& \multicolumn{1}{c|}{{\color[HTML]{1600FF} 0.041}} & \multicolumn{1}{c|}{{\color[HTML]{1600FF} 0.098}} & \multicolumn{1}{c|}{{\color[HTML]{1600FF} 0.143}} & \multicolumn{1}{c|}{{\color[HTML]{1600FF} 0.074}} & {\color[HTML]{1600FF} 0.169} & \multicolumn{1}{c|}{{\color[HTML]{1600FF} 0.238}} & \multicolumn{1}{c|}{{\color[HTML]{1600FF} 0.113}} & \multicolumn{1}{c|}{{\color[HTML]{1600FF} 0.237}} & \multicolumn{1}{c|}{{\color[HTML]{1600FF} 0.320}} & {\color[HTML]{1600FF} 0.053} & \multicolumn{1}{c|}{{\color[HTML]{1600FF} 0.132}} & \multicolumn{1}{c|}{{\color[HTML]{1600FF} 0.199}} & \multicolumn{1}{c|}{{\color[HTML]{1600FF} 0.075}} & \multicolumn{1}{c|}{{\color[HTML]{1600FF} 0.106}} & {\color[HTML]{1600FF} 0.141}                            \\ \cline{2-17} 

& LIIF$^\ddag$              
& \multicolumn{1}{c|}{{\color[HTML]{1600FF} 0.041}} & \multicolumn{1}{c|}{{\color[HTML]{1600FF} 0.096}} & \multicolumn{1}{c|}{{\color[HTML]{1600FF} 0.139}} & \multicolumn{1}{c|}{{\color[HTML]{1600FF} 0.072}} & {\color[HTML]{1600FF} 0.167} & \multicolumn{1}{c|}{{\color[HTML]{1600FF} 0.235}} & \multicolumn{1}{c|}{{\color[HTML]{1600FF} 0.113}} & \multicolumn{1}{c|}{{\color[HTML]{1600FF} 0.235}} & \multicolumn{1}{c|}{{\color[HTML]{1600FF} 0.316}} & {\color[HTML]{1600FF} 0.049} & \multicolumn{1}{c|}{{\color[HTML]{1600FF} 0.125}} & \multicolumn{1}{c|}{{\color[HTML]{1600FF} 0.189}} & \multicolumn{1}{c|}{{\color[HTML]{1600FF} 0.075}} & \multicolumn{1}{c|}{{\color[HTML]{1600FF} 0.105}} & {\color[HTML]{1600FF} 0.137}     \\ \cline{2-17}

& ArbSR$^\ast $                  
& \multicolumn{1}{c|}{0.022}                              & \multicolumn{1}{c|}{0.052}                        & 0.075    
& \multicolumn{1}{c|}{0.038}                        & \multicolumn{1}{c|}{0.085}                        & 0.123                        & \multicolumn{1}{c|}{0.054}                        & \multicolumn{1}{c|}{0.097}                        & 0.136                        & \multicolumn{1}{c|}{0.038}                        & \multicolumn{1}{c|}{0.088}                        & 0.138                        & \multicolumn{1}{c|}{0.011}                        & \multicolumn{1}{c|}{0.033}                        & 0.059                        \\ \cline{2-17}

& ArbSR$^\dag$         & \multicolumn{1}{c|}{{\color[HTML]{1600FF} 0.023}}       & \multicolumn{1}{c|}{{\color[HTML]{1600FF} 0.052}} & {\color[HTML]{FF0000} 0.074} & \multicolumn{1}{c|}{{\color[HTML]{1600FF} 0.041}} & \multicolumn{1}{c|}{{\color[HTML]{1600FF} 0.085}} & {\color[HTML]{1600FF} 0.123} & \multicolumn{1}{c|}{{\color[HTML]{1600FF} 0.055}} & \multicolumn{1}{c|}{{\color[HTML]{1600FF} 0.097}} & {\color[HTML]{FF0000} 0.135} & \multicolumn{1}{c|}{{\color[HTML]{1600FF} 0.042}} & \multicolumn{1}{c|}{{\color[HTML]{1600FF} 0.090}} & {\color[HTML]{1600FF} 0.139} & \multicolumn{1}{c|}{{\color[HTML]{1600FF} 0.011}} & \multicolumn{1}{c|}{{\color[HTML]{1600FF} 0.033}} & {\color[HTML]{1600FF} 0.059} \\ \cline{2-17}  

& ArbSR$^\ddag$       & \multicolumn{1}{c|}{{\color[HTML]{1600FF} 0.022}}       & \multicolumn{1}{c|}{{\color[HTML]{1600FF} 0.052}} & {\color[HTML]{FF0000} 0.074} & \multicolumn{1}{c|}{{\color[HTML]{1600FF} 0.038}} & \multicolumn{1}{c|}{{\color[HTML]{1600FF} 0.085}} & {\color[HTML]{1600FF} 0.123} & \multicolumn{1}{c|}{{\color[HTML]{1600FF} 0.054}} & \multicolumn{1}{c|}{{\color{blue} 0.097}} & {\color[HTML]{FF0000} 0.135} & \multicolumn{1}{c|}{{\color[HTML]{FF0000} 0.037}} & \multicolumn{1}{c|}{{\color[HTML]{1600FF} 0.088}} & {\color[HTML]{1600FF} 0.139} & \multicolumn{1}{c|}{{\color[HTML]{1600FF} 0.011}} & \multicolumn{1}{c|}{{\color[HTML]{1600FF} 0.033}} & {\color[HTML]{1600FF} 0.059}                             \\ \cline{2-17}

& SRNO                   & \multicolumn{1}{c|}{0.040}                              & \multicolumn{1}{c|}{0.092}                        & 0.137                        & \multicolumn{1}{c|}{0.070}                        & \multicolumn{1}{c|}{0.162}                        & 0.226                        & \multicolumn{1}{c|}{0.111}                        & \multicolumn{1}{c|}{0.229}                        & 0.303                        & \multicolumn{1}{c|}{0.045}                        & \multicolumn{1}{c|}{0.118}                        & 0.175                        & \multicolumn{1}{c|}{0.015}                        & \multicolumn{1}{c|}{0.045}                        & 0.077                        \\ \cline{2-17}

& SRNO$^\dag$     & \multicolumn{1}{c|}{{\color[HTML]{1600FF} 0.041}}       & \multicolumn{1}{c|}{{\color[HTML]{1600FF} 0.092}} & {\color[HTML]{1600FF} 0.138} & \multicolumn{1}{c|}{{\color[HTML]{1600FF} 0.072}} & \multicolumn{1}{c|}{{\color[HTML]{1600FF} 0.164}} & {\color[HTML]{1600FF} 0.227} & \multicolumn{1}{c|}{{\color[HTML]{1600FF} 0.112}} & \multicolumn{1}{c|}{{\color[HTML]{1600FF} 0.230}} & {\color[HTML]{1600FF} 0.304} & \multicolumn{1}{c|}{{\color[HTML]{1600FF} 0.048}} & \multicolumn{1}{c|}{{\color[HTML]{1600FF} 0.120}} & {\color[HTML]{1600FF} 0.176} & \multicolumn{1}{c|}{{\color[HTML]{1600FF} 0.016}} & \multicolumn{1}{c|}{{\color[HTML]{1600FF} 0.046}} & {\color[HTML]{1600FF} 0.078} \\ \cline{2-17} 

& SRNO$^\ddag$      & \multicolumn{1}{c|}{{\color[HTML]{1600FF} 0.040}}       & \multicolumn{1}{c|}{{\color[HTML]{1600FF} 0.093}} & {\color[HTML]{1600FF} 0.138} & \multicolumn{1}{c|}{{\color[HTML]{1600FF} 0.071}} & \multicolumn{1}{c|}{{\color[HTML]{1600FF} 0.163}} & {\color[HTML]{1600FF} 0.227} & \multicolumn{1}{c|}{{\color[HTML]{1600FF} 0.112}} & \multicolumn{1}{c|}{{\color[HTML]{1600FF} 0.230}} & {\color[HTML]{1600FF} 0.304} & \multicolumn{1}{c|}{{\color[HTML]{1600FF} 0.047}} & \multicolumn{1}{c|}{{\color[HTML]{1600FF} 0.119}} & {\color[HTML]{1600FF} 0.176} & \multicolumn{1}{c|}{{\color[HTML]{1600FF} 0.016}} & \multicolumn{1}{c|}{{\color[HTML]{1600FF} 0.046}} & {\color[HTML]{1600FF} 0.078}                               
\\ 
\hline

\multirow{26}*{RDN}    
& Metric
& \multicolumn{15}{c}{PSNR$\uparrow$}
\\ \cline{2-17}
& MetaSR                   & \multicolumn{1}{c|}{38.22}   & \multicolumn{1}{c|}{34.63}   & 32.38    & \multicolumn{1}{c|}{33.98}   & \multicolumn{1}{c|}{30.54}   & 28.78   & \multicolumn{1}{c|}{32.33}   & \multicolumn{1}{c|}{29.26}   & 27.71   & \multicolumn{1}{c|}{32.92}   & \multicolumn{1}{c|}{28.82}   & 26.55   & \multicolumn{1}{c|}{39.18}   & \multicolumn{1}{c|}{34.14}   & 31.03   \\ \cline{2-17}

& MetaSR$^\dag$     & \multicolumn{1}{c|}{{\color{blue}38.13}}   & \multicolumn{1}{c|}{{\color{blue}34.57}}   & {\color{red}32.51}   & \multicolumn{1}{c|}{{\color{blue}33.94}}   & \multicolumn{1}{c|}{{\color{blue}30.50}}   & {\color{red}28.83}   & \multicolumn{1}{c|}{{\color{blue}32.25}}   & \multicolumn{1}{c|}{{\color{blue}29.21}}   & {\color{red}27.74}   & \multicolumn{1}{c|}{32.69}   & \multicolumn{1}{c|}{28.63}   & {\color{red}26.69}   & \multicolumn{1}{c|}{{\color{blue}39.06}}   & \multicolumn{1}{c|}{{\color{blue}34.07}}   & {\color{red}31.27}   \\ \cline{2-17}  

& MetaSR$^\ddag$     & \multicolumn{1}{c|}{{\color{blue}38.19}}   & \multicolumn{1}{c|}{{\color{red}34.72}}   & {\color{red}32.51}   & \multicolumn{1}{c|}{{\color{blue}33.94}}   & \multicolumn{1}{c|}{{\color{red}30.59}}   & {\color{red}28.83}   & \multicolumn{1}{c|}{{\color{blue}32.30}}   & \multicolumn{1}{c|}{{\color{red}29.27}}   & {\color{red}27.74}   & \multicolumn{1}{c|}{{\color{red}32.94}}   & \multicolumn{1}{c|}{{\color{red}28.92}}   & {\color{red}26.69}   & \multicolumn{1}{c|}{{\color{red}39.23}}   & \multicolumn{1}{c|}{{\color{red}34.32}}   & {\color{red}31.27}   \\ \cline{2-17}

& LIIF                     & \multicolumn{1}{c|}{38.17}   & \multicolumn{1}{c|}{34.68}   & 32.50   & \multicolumn{1}{c|}{33.97}   & \multicolumn{1}{c|}{30.53}   & 28.80   & \multicolumn{1}{c|}{32.32}   & \multicolumn{1}{c|}{29.26}   & 27.74   & \multicolumn{1}{c|}{32.87}   & \multicolumn{1}{c|}{28.82}   & 26.68  & \multicolumn{1}{c|}{39.26}   & \multicolumn{1}{c|}{34.21}   & 31.20   \\ \cline{2-17}

& LIIF$^\dag$              & \multicolumn{1}{c|}{{\color{blue}38.15}}   & \multicolumn{1}{c|}{{\color{blue}34.58}}   & {\color{blue}32.49}   & \multicolumn{1}{c|}{{\color{blue}33.84}}   & \multicolumn{1}{c|}{{\color{blue}30.49}}   & {\color{red}28.82}   & \multicolumn{1}{c|}{{\color{blue}32.27}}   & \multicolumn{1}{c|}{{\color{blue}29.21}}   & {\color{blue}27.74}   & \multicolumn{1}{c|}{32.59}   & \multicolumn{1}{c|}{28.62}   &  {\color{blue}26.68}  & \multicolumn{1}{c|}{39.02}   & \multicolumn{1}{c|}{34.02}   & {\color{red}31.22}   \\ \cline{2-17}                        %

& LIIF$^\ddag$              & \multicolumn{1}{c|}{{\color{red}38.19}}   & \multicolumn{1}{c|}{{\color{blue}34.66}}   &  {\color{blue}32.49}  & \multicolumn{1}{c|}{{\color{red}33.99}}   & \multicolumn{1}{c|}{{\color{red}30.54}}   & {\color{red}28.82}   & \multicolumn{1}{c|}{{\color{blue}32.31}}   & \multicolumn{1}{c|}{{\color{red}29.27}}   & {\color{blue}27.74}   & \multicolumn{1}{c|}{{\color{blue}32.79}}   & \multicolumn{1}{c|}{{\color{blue}28.81}}   & {\color{blue}26.68}   & \multicolumn{1}{c|}{{\color{blue}39.19}}   & \multicolumn{1}{c|}{{\color{blue}34.20}}   & {\color{red}31.22}   \\ \cline{2-17}

& ArbSR$^\ast$                    & \multicolumn{1}{c|}{38.10}   & \multicolumn{1}{c|}{34.57}   & 32.26   & \multicolumn{1}{c|}{33.83}   & \multicolumn{1}{c|}{30.46}   & 28.68   & \multicolumn{1}{c|}{32.26}   & \multicolumn{1}{c|}{29.19}   & 27.64   & \multicolumn{1}{c|}{32.47}   & \multicolumn{1}{c|}{28.45}   & 26.23   & \multicolumn{1}{c|}{38.81}   & \multicolumn{1}{c|}{34.02}   & 30.87   \\ \cline{2-17}

& ArbSR$^\dag$              & \multicolumn{1}{c|}{{\color{blue}38.01}}   & \multicolumn{1}{c|}{{\color{blue}34.43}}   & {\color{red}32.31}   & \multicolumn{1}{c|}{33.61}   & \multicolumn{1}{c|}{{\color{blue}30.37}}   & {\color{red}28.70}   & \multicolumn{1}{c|}{{\color{blue}32.17}}   & \multicolumn{1}{c|}{{\color{blue}29.11}}   &  {\color{red}27.66}   & \multicolumn{1}{c|}{32.13}   & \multicolumn{1}{c|}{28.14}   & {\color{red}26.30}   & \multicolumn{1}{c|}{38.61}   & \multicolumn{1}{c|}{33.70}   & {\color{red}30.94}   \\ \cline{2-17}

& ArbSR$^\ddag$          & \multicolumn{1}{c|}{{\color{blue}38.05}}   & \multicolumn{1}{c|}{{\color{red}34.59}}   & {\color{red}32.31}   & \multicolumn{1}{c|}{{\color{blue}33.80}}   & \multicolumn{1}{c|}{{\color{red}30.47}}   & {\color{red}28.70}   & \multicolumn{1}{c|}{{\color{red}32.27}}   & \multicolumn{1}{c|}{{\color{red}29.20}}   & {\color{red}27.66}   & \multicolumn{1}{c|}{{\color{red}32.57}}   & \multicolumn{1}{c|}{{\color{red}28.52}}   & {\color{red}26.30}   & \multicolumn{1}{c|}{{\color{red}38.82}}   & \multicolumn{1}{c|}{{\color{blue}34.01}}   & {\color{red}30.94}   \\ \cline{2-17} 

& SRNO                   & \multicolumn{1}{c|}{38.32}   & \multicolumn{1}{c|}{34.84}   & 32.69   & \multicolumn{1}{c|}{34.27}   & \multicolumn{1}{c|}{30.71}   & 28.97   & \multicolumn{1}{c|}{32.43}   & \multicolumn{1}{c|}{29.37}   & 27.83   & \multicolumn{1}{c|}{33.33}   & \multicolumn{1}{c|}{29.14}   & 26.98   & \multicolumn{1}{c|}{39.47}   & \multicolumn{1}{c|}{34.62}   & 31.58   \\ \cline{2-17}

& SRNO$^\dag$       & \multicolumn{1}{c|}{{\color{blue}38.18}}   & \multicolumn{1}{c|}{{\color{blue}34.73}}   & {\color{red}32.72}   & \multicolumn{1}{c|}{34.03}   & \multicolumn{1}{c|}{{\color{blue}30.60}}   & {\color{red}28.98}   & \multicolumn{1}{c|}{{\color{blue}32.33}}  & \multicolumn{1}{c|}{{\color{blue}29.27}}   & {\color{blue}27.81}   & \multicolumn{1}{c|}{32.94}   & \multicolumn{1}{c|}{28.87}   & {\color{blue}26.95}   & \multicolumn{1}{c|}{39.25}   & \multicolumn{1}{c|}{34.33}   & {\color{blue}31.53}   
  \\ \cline{2-17} 

& SRNO$^\ddag$       & \multicolumn{1}{c|}{{\color{blue}38.27}}   & \multicolumn{1}{c|}{{\color{blue}34.83}}   & {\color{red}32.72}   & \multicolumn{1}{c|}{{\color{blue}34.23}}   & \multicolumn{1}{c|}{{\color{blue}30.69}}   & {\color{red}28.98}   & \multicolumn{1}{c|}{{\color{blue}32.39}}   & \multicolumn{1}{c|}{{\color{blue}29.34}}   & {\color{blue}27.81}   & \multicolumn{1}{c|}{{\color{blue}33.23}}   & \multicolumn{1}{c|}{{\color{blue}29.10}}   & {\color{blue}26.95}   & \multicolumn{1}{c|}{{\color{blue}39.43}}   & \multicolumn{1}{c|}{{\color{blue}34.58}}   & {\color{blue}31.53}   \\

\cline{2-17}
& Metric
& \multicolumn{15}{c}{LPIPS$\downarrow$}
\\ \cline{2-17}

& MetaSR                   
& \multicolumn{1}{c|}{0.021}                              & \multicolumn{1}{c|}{0.051}                        & 0.075                        & \multicolumn{1}{c|}{0.037}                        & \multicolumn{1}{c|}{0.082}                        & 0.120                        & \multicolumn{1}{c|}{0.053}                        & \multicolumn{1}{c|}{0.098}                        & 0.132                        & \multicolumn{1}{c|}{0.037}                        & \multicolumn{1}{c|}{0.088}                        & 0.130                        & \multicolumn{1}{c|}{0.037}                        & \multicolumn{1}{c|}{0.032}                        & 0.053                        \\ \cline{2-17}

& MetaSR$^\dag$     & \multicolumn{1}{c|}{{\color[HTML]{1600FF} 0.022}}       & \multicolumn{1}{c|}{{\color[HTML]{1600FF} 0.052}} & {\color[HTML]{FF0000} 0.074} & \multicolumn{1}{c|}{{\color[HTML]{1600FF} 0.038}} & \multicolumn{1}{c|}{{\color[HTML]{1600FF} 0.085}} & {\color[HTML]{1600FF} 0.121} & \multicolumn{1}{c|}{{\color[HTML]{1600FF} 0.055}} & \multicolumn{1}{c|}{{\color[HTML]{1600FF} 0.100}} & {\color[HTML]{FF0000} 0.132} & \multicolumn{1}{c|}{{\color[HTML]{1600FF} 0.038}} & \multicolumn{1}{c|}{{\color[HTML]{1600FF} 0.091}} & {\color[HTML]{FF0000} 0.129} & \multicolumn{1}{c|}{{\color[HTML]{1600FF} 0.038}} & \multicolumn{1}{c|}{{\color[HTML]{1600FF} 0.033}} & {\color[HTML]{FF0000} 0.053} \\ \cline{2-17} 

& MetaSR$^\ddag$    & \multicolumn{1}{c|}{{\color[HTML]{1600FF} 0.021}}       & \multicolumn{1}{c|}{{\color[HTML]{FF0000} 0.050}} & {\color[HTML]{FF0000} 0.074} & \multicolumn{1}{c|}{{\color[HTML]{1600FF} 0.037}} & \multicolumn{1}{c|}{{\color[HTML]{1600FF} 0.083}} & {\color[HTML]{1600FF} 0.121} & \multicolumn{1}{c|}{{\color[HTML]{1600FF} 0.053}} & \multicolumn{1}{c|}{{\color[HTML]{1600FF} 0.098}} & {\color[HTML]{FF0000} 0.132} & \multicolumn{1}{c|}{{\color[HTML]{1600FF} 0.037}} & \multicolumn{1}{c|}{{\color[HTML]{1600FF} 0.088}} & {\color[HTML]{FF0000} 0.129} & \multicolumn{1}{c|}{{\color[HTML]{1600FF} 0.037}} & \multicolumn{1}{c|}{{\color[HTML]{FF0000} 0.032}} & {\color[HTML]{FF0000} 0.053} \\ \cline{2-17} 

& LIIF                    & \multicolumn{1}{c|}{0.039}                              & \multicolumn{1}{c|}{0.092}                        & 0.133                        & \multicolumn{1}{c|}{0.067}                        & \multicolumn{1}{c|}{0.157}                        & 0.224                        & \multicolumn{1}{c|}{0.111}                        & \multicolumn{1}{c|}{0.227}                        & 0.303                        & \multicolumn{1}{c|}{0.043}                        & \multicolumn{1}{c|}{0.110}                        & 0.170                        & \multicolumn{1}{c|}{0.074}                        & \multicolumn{1}{c|}{0.101}                        & 0.130                        \\ \cline{2-17}

& LIIF$^\dag$             & \multicolumn{1}{c|}{{\color[HTML]{1600FF} 0.040}}       & \multicolumn{1}{c|}{{\color[HTML]{1600FF} 0.093}} & {\color[HTML]{1600FF} 0.136} & \multicolumn{1}{c|}{{\color[HTML]{1600FF} 0.071}} & \multicolumn{1}{c|}{{\color[HTML]{1600FF} 0.165}} & {\color[HTML]{1600FF} 0.233} & \multicolumn{1}{c|}{{\color[HTML]{1600FF} 0.113}} & \multicolumn{1}{c|}{{\color[HTML]{1600FF} 0.233}} & {\color[HTML]{1600FF} 0.312} & \multicolumn{1}{c|}{{\color[HTML]{1600FF} 0.046}} & \multicolumn{1}{c|}{{\color[HTML]{1600FF} 0.122}} & {\color[HTML]{1600FF} 0.187} & \multicolumn{1}{c|}{{\color[HTML]{1600FF} 0.074}} & \multicolumn{1}{c|}{{\color[HTML]{1600FF} 0.103}} & {\color[HTML]{1600FF} 0.135} \\ \cline{2-17}                  %

& LIIF$^\ddag$              & \multicolumn{1}{c|}{{\color[HTML]{1600FF} 0.040}}       & \multicolumn{1}{c|}{{\color[HTML]{1600FF} 0.094}} & {\color[HTML]{1600FF} 0.137} & \multicolumn{1}{c|}{{\color[HTML]{1600FF} 0.071}} & \multicolumn{1}{c|}{{\color[HTML]{1600FF} 0.165}} & {\color[HTML]{1600FF} 0.232} & \multicolumn{1}{c|}{{\color[HTML]{1600FF} 0.113}} & \multicolumn{1}{c|}{{\color[HTML]{1600FF} 0.235}} & {\color[HTML]{1600FF} 0.313} & \multicolumn{1}{c|}{{\color[HTML]{1600FF} 0.045}} & \multicolumn{1}{c|}{{\color[HTML]{1600FF} 0.119}} & {\color[HTML]{1600FF} 0.182} & \multicolumn{1}{c|}{{\color[HTML]{1600FF} 0.074}} & \multicolumn{1}{c|}{{\color[HTML]{1600FF} 0.103}} & {\color[HTML]{1600FF} 0.133} \\ \cline{2-17}

& ArbSR$^\ast$                  & \multicolumn{1}{c|}{0.021}                              & \multicolumn{1}{c|}{0.052}                        & 0.075                        & \multicolumn{1}{c|}{0.037}                        & \multicolumn{1}{c|}{0.084}                        & 0.122                        & \multicolumn{1}{c|}{0.054}                        & \multicolumn{1}{c|}{0.098}                        & 0.135                        & \multicolumn{1}{c|}{0.036}                        & \multicolumn{1}{c|}{0.087}                        & 0.134                        & \multicolumn{1}{c|}{0.011}                        & \multicolumn{1}{c|}{0.033}                        & 0.057                        \\ \cline{2-17}

& ArbSR$^\dag$            & \multicolumn{1}{c|}{{\color[HTML]{1600FF} 0.021}}       & \multicolumn{1}{c|}{{\color[HTML]{1600FF} 0.052}} & {\color[HTML]{FF0000} 0.074} & \multicolumn{1}{c|}{{\color[HTML]{1600FF} 0.039}} & \multicolumn{1}{c|}{{\color[HTML]{1600FF} 0.086}} & {\color[HTML]{1600FF} 0.122} & \multicolumn{1}{c|}{{\color[HTML]{1600FF} 0.054}} & \multicolumn{1}{c|}{{\color[HTML]{FF0000} 0.098}} & {\color[HTML]{FF0000} 0.134} & \multicolumn{1}{c|}{{\color[HTML]{1600FF} 0.038}} & \multicolumn{1}{c|}{{\color[HTML]{1600FF} 0.089}} & {\color[HTML]{FF0000} 0.128} & \multicolumn{1}{c|}{{\color[HTML]{1600FF} 0.011}} & \multicolumn{1}{c|}{{\color[HTML]{1600FF} 0.033}} & {\color[HTML]{FF0000} 0.055} \\ \cline{2-17}

& ArbSR$^\ddag$        & \multicolumn{1}{c|}{{\color[HTML]{1600FF} 0.021}}       & \multicolumn{1}{c|}{{\color[HTML]{1600FF} 0.052}} & {\color[HTML]{FF0000} 0.074} & \multicolumn{1}{c|}{{\color[HTML]{1600FF} 0.037}} & \multicolumn{1}{c|}{{\color[HTML]{1600FF} 0.085}} & {\color[HTML]{1600FF} 0.122} & \multicolumn{1}{c|}{{\color[HTML]{1600FF} 0.054}} & \multicolumn{1}{c|}{{\color[HTML]{FF0000} 0.097}} & {\color[HTML]{FF0000} 0.134} & \multicolumn{1}{c|}{{\color[HTML]{FF0000} 0.035}} & \multicolumn{1}{c|}{{\color[HTML]{FF0000} 0.083}} & {\color[HTML]{FF0000} 0.128} & \multicolumn{1}{c|}{{\color[HTML]{1600FF} 0.011}} & \multicolumn{1}{c|}{{\color[HTML]{FF0000} 0.032}} & {\color[HTML]{FF0000} 0.055}  \\ \cline{2-17} 

& SRNO                  & \multicolumn{1}{c|}{0.039}                              & \multicolumn{1}{c|}{0.090}                        & 0.133                        & \multicolumn{1}{c|}{0.066}                        & \multicolumn{1}{c|}{0.157}                        & 0.222                        & \multicolumn{1}{c|}{0.105}                        & \multicolumn{1}{c|}{0.222}                        & 0.297                        & \multicolumn{1}{c|}{0.039}                        & \multicolumn{1}{c|}{0.106}                        & 0.162                        & \multicolumn{1}{c|}{0.015}                        & \multicolumn{1}{c|}{0.043}                        & 0.073                        \\ \cline{2-17}

& SRNO$^\dag$      & \multicolumn{1}{c|}{{\color[HTML]{1600FF} 0.040}}       & \multicolumn{1}{c|}{{\color[HTML]{1600FF} 0.093}} & {\color[HTML]{1600FF} 0.133} & \multicolumn{1}{c|}{{\color[HTML]{1600FF} 0.068}} & \multicolumn{1}{c|}{{\color[HTML]{1600FF} 0.161}} & {\color[HTML]{1600FF} 0.224} & \multicolumn{1}{c|}{{\color[HTML]{1600FF} 0.106}} & \multicolumn{1}{c|}{{\color[HTML]{1600FF} 0.223}} & {\color[HTML]{1600FF} 0.298} & \multicolumn{1}{c|}{{\color[HTML]{1600FF} 0.041}} & \multicolumn{1}{c|}{{\color[HTML]{1600FF} 0.111}} & {\color[HTML]{1600FF} 0.163} & \multicolumn{1}{c|}{{\color[HTML]{1600FF} 0.015}} & \multicolumn{1}{c|}{{\color[HTML]{1600FF} 0.044}} & {\color[HTML]{1600FF} 0.074} \\ \cline{2-17} 

& SRNO$^\ddag$     & \multicolumn{1}{c|}{{\color[HTML]{1600FF} 0.039}}       & \multicolumn{1}{c|}{{\color[HTML]{1600FF} 0.091}} & {\color[HTML]{1600FF} 0.133} & \multicolumn{1}{c|}{{\color[HTML]{1600FF} 0.067}} & \multicolumn{1}{c|}{{\color[HTML]{1600FF} 0.160}} & {\color[HTML]{1600FF} 0.224} & \multicolumn{1}{c|}{{\color[HTML]{1600FF} 0.106}} & \multicolumn{1}{c|}{{\color[HTML]{1600FF} 0.224}} & {\color[HTML]{1600FF} 0.298} & \multicolumn{1}{c|}{{\color[HTML]{1600FF} 0.040}} & \multicolumn{1}{c|}{{\color[HTML]{1600FF} 0.107}} & {\color[HTML]{1600FF} 0.163} & \multicolumn{1}{c|}{{\color[HTML]{1600FF} 0.015}} & \multicolumn{1}{c|}{{\color[HTML]{1600FF} 0.043}} & {\color[HTML]{1600FF} 0.074}  \\ \bottomrule

\end{tabular}}
\end{table*}

\section{Experimental Results}
\label{sec:experimentation}
\subsection{Experiment Settings}

\subsubsection{Training Details}
We primarily rely on existing well-established arbitrary-scale SR models, including Meta-SR~\cite{hu2019meta}, LIIF~\cite{chen2021learning}, ArbSR~\cite{wang2021learning}, and SRNO~\cite{srnowei2023super}, for AnySR rebuilding.
Without loss of generality, experiments on top of two famous feature extraction backbones, including EDSR~\cite{Lim_2017_CVPR_Workshops} and RDN~\cite{RDNZhang_2018_CVPR}, are conducted to validate the universality of AnySR.

During specific implementation, we configure the network number $T = 4$ and $\{  w_i \}^T_{i=1} =  \{ 0.5, 0.7, 0.9, 1.0 \}$, that is, the smallest scale group $S_1$ enables a reduction in inference cost by up to 50\% compared with the original.
The reset probability $p$ in Line 4 of Algorithm\,\ref{alg:training} is configured at 0.6, and the hyper-parameter $\lambda$ in feature-interweaving is set as 4 for EDSR and 8 for RDN.
Accordingly, the scale groups are designated as {$\mathcal{S}_1 = \{1.1, 1.2, ..., 1.7\}$, $\mathcal{S}_2 = \{1.8, 1.9, ..., 2.5\}$, $\mathcal{S}_3 = \{2.6, 2.7, ..., 3.2\}$ and $\mathcal{S}_4 = \{3.3, 3.4, ..., 4.0\}$}.

We train AnySR for 500 epochs from the pre-trained models, with an initial learning rate $10^{-5}$ decayed by 0.5 every 200 epochs.
As pre-trained models for some existing methods (Meta-SR: EDSR; ArbSR: EDSR, RDN) are not available, we re-implement and re-train them following the experiment settings outlined in the papers~\cite{hu2019meta,wang2021learning}.
To ensure a fair performance evaluation, we keep identical settings and configurations used for the original models in our training process. The patch size is $50\times50$ for Meta-SR and ArbSR, $48\times48$ for LIIF, and $128\times128$ for SRNO, with a batch size of 8 per GPU for EDSR and 4 per GPU for RDN.
We employ an $\ell_1$ loss~\cite{Lim_2017_CVPR_Workshops} and the Adam optimizer~\cite{kingma2014adam}.
\subsubsection{Dataset}

For training, current practices~\cite{hatchen2023activating} use DF2K (including DIV2K~\cite{agustsson2017ntire} and Flicker2K~\cite{Timofte_2017_CVPR_Workshops}) as the training set, or pre-training on ImageNet1K~\cite{deng2009imagenet} to enhance performance. 
For fairness, we choose to follow the traditional training regimen directly on the DIV2K dataset to ensure a direct comparison with previous studies and to maintain consistency with the experimental protocols established in the literature.

For evaluation, we assess the PSNR~\cite{psnr} and LPIPS~\cite{lpips} on the most common evaluation datasets, including Set5~\cite{bevilacqua2012low}, Set14~\cite{zeyde2012single}, B100~\cite{timofte2013anchored}, Urban100~\cite{huang2015single}, and Manga109~\cite{aizawa2020building}.

\begin{table*}[]
\centering
\caption{PSNR (dB) and LPIPS comparison for fractional-scale SR tasks of SRNO~\cite{srnowei2023super} and their AnySR variants, highlighted by $^\dag$ (through different subnets) and $^\ddag$ (AnySR-retrained version via the largest network). AnySR performance is highlighted in {\color{blue}blue} if within 0.15dB (PSNR) / 0.01 (LPIPS) of the vanilla method, and {\color{red}red} if improved.}
\vspace{-2mm}
\label{tab:result-psnr-detail-edsr}
\setlength\tabcolsep{6.5pt}
\resizebox{\textwidth}{!}{
\renewcommand{\arraystretch}{1.5}
\begin{tabular}{c|c|ccc|ccc|ccc|ccc|ccc}
\toprule
\multirow{2}{*}{\begin{tabular}[c]{@{}c@{}}Backbone\\ Network\end{tabular}} & \multirow{2}{*}{Methods} & \multicolumn{3}{c|}{Set5}                & \multicolumn{3}{c|}{Set14}               & \multicolumn{3}{c|}{B100}                & \multicolumn{3}{c|}{Urban100}            & \multicolumn{3}{c}{Manga109}            \\
\cline{3-17}

&            & \multicolumn{1}{c|}{$\times$1.8} & \multicolumn{1}{c|}{$\times$2.6} & $\times$3.6 & \multicolumn{1}{c|}{$\times$1.9} & \multicolumn{1}{c|}{$\times$2.9} & $\times$3.8 & \multicolumn{1}{c|}{$\times$1.9} & \multicolumn{1}{c|}{$\times$2.7} & $\times$3.5 & \multicolumn{1}{c|}{$\times$1.8} & \multicolumn{1}{c|}{$\times$2.6} & $\times$3.7 & \multicolumn{1}{c|}{$\times$1.3} & \multicolumn{1}{c|}{$\times$2.6} & $\times$3.3  \\
\cline{1-17}

\multirow{8}{*}{EDSR}    

& Metric & \multicolumn{15}{c}{PSNR$\uparrow$}
\\ \cline{2-17}

& SRNO                     & \multicolumn{1}{c|}{39.12}       & \multicolumn{1}{c|}{35.72}       & 33.14       & \multicolumn{1}{c|}{34.42}       & \multicolumn{1}{c|}{30.74}       & 29.06       & \multicolumn{1}{c|}{32.79}       & \multicolumn{1}{c|}{29.87}       & 28.33       & \multicolumn{1}{c|}{33.94}       & \multicolumn{1}{c|}{29.83}       & 27.02       & \multicolumn{1}{c|}{45.29}       & \multicolumn{1}{c|}{35.71}       & 32.93        \\
\cline{2-17}

& SRNO$^\dag$              & \multicolumn{1}{c|}{{\color[HTML]{1600FF} 39.04}}       & \multicolumn{1}{c|}{{\color[HTML]{1600FF} 35.68}} & {\color[HTML]{1600FF} 33.12} & \multicolumn{1}{c|}{{\color[HTML]{1600FF} 34.34}} & \multicolumn{1}{c|}{{\color[HTML]{1600FF} 30.70}} & {\color[HTML]{1600FF} 29.04} & \multicolumn{1}{c|}{{\color[HTML]{1600FF} 32.73}} & \multicolumn{1}{c|}{{\color[HTML]{1600FF} 29.84}} & {\color[HTML]{1600FF} 28.31} & \multicolumn{1}{c|}{{\color[HTML]{000000} 33.60}} & \multicolumn{1}{c|}{{\color[HTML]{1600FF} 29.72}} & {\color[HTML]{1600FF} 26.99} & \multicolumn{1}{c|}{{\color[HTML]{000000} 45.01}} & \multicolumn{1}{c|}{{\color[HTML]{1600FF} 35.57}} & {\color[HTML]{1600FF} 32.87}      \\
\cline{2-17}

& SRNO$^\ddag$             & \multicolumn{1}{c|}{{\color[HTML]{1600FF} 39.11}}       & \multicolumn{1}{c|}{{\color[HTML]{1600FF} 35.70}} & {\color[HTML]{1600FF} 33.12} & \multicolumn{1}{c|}{{\color[HTML]{1600FF} 34.39}} & \multicolumn{1}{c|}{{\color[HTML]{1600FF} 30.71}} & {\color[HTML]{1600FF} 29.04} & \multicolumn{1}{c|}{{\color[HTML]{1600FF} 32.78}} & \multicolumn{1}{c|}{{\color[HTML]{1600FF} 29.85}} & {\color[HTML]{1600FF} 28.31} & \multicolumn{1}{c|}{{\color[HTML]{1600FF} 33.85}} & \multicolumn{1}{c|}{{\color[HTML]{1600FF} 29.77}} & {\color[HTML]{1600FF} 26.99} & \multicolumn{1}{c|}{{\color[HTML]{1600FF} 45.28}} & \multicolumn{1}{c|}{{\color[HTML]{1600FF} 35.63}} & {\color[HTML]{1600FF} 32.87}   \\
\cline{2-17}
& Metric & \multicolumn{15}{c}{LPIPS$\downarrow$}
\\ \cline{2-17}

& SRNO 
& \multicolumn{1}{c|}{0.028}                        & \multicolumn{1}{c|}{0.070}                        & \multicolumn{1}{c|}{0.119}                        & \multicolumn{1}{c|}{0.048}                        & 0.127                        & \multicolumn{1}{c|}{0.203}                        & \multicolumn{1}{c|}{0.078}                        & \multicolumn{1}{c|}{0.184}                        & \multicolumn{1}{c|}{0.274}                        & 0.031                        & \multicolumn{1}{c|}{0.089}                        & \multicolumn{1}{c|}{0.154}                        & \multicolumn{1}{c|}{0.010}                        & \multicolumn{1}{c|}{0.032}                        & 0.065    \\
\cline{2-17}

& SRNO$\dag$
& \multicolumn{1}{c|}{{\color[HTML]{1600FF} 0.028}} & \multicolumn{1}{c|}{{\color[HTML]{1600FF} 0.071}} & \multicolumn{1}{c|}{{\color[HTML]{1600FF} 0.118}} & \multicolumn{1}{c|}{{\color[HTML]{1600FF} 0.050}} & {\color[HTML]{1600FF} 0.129} & \multicolumn{1}{c|}{{\color[HTML]{1600FF} 0.204}} & \multicolumn{1}{c|}{{\color[HTML]{1600FF} 0.079}} & \multicolumn{1}{c|}{{\color[HTML]{1600FF} 0.185}} & \multicolumn{1}{c|}{{\color[HTML]{1600FF} 0.275}} & {\color[HTML]{1600FF} 0.033} & \multicolumn{1}{c|}{{\color[HTML]{1600FF} 0.091}} & \multicolumn{1}{c|}{{\color[HTML]{1600FF} 0.155}} & \multicolumn{1}{c|}{{\color[HTML]{1600FF} 0.011}} & \multicolumn{1}{c|}{{\color[HTML]{1600FF} 0.032}} & {\color[HTML]{1600FF} 0.065} \\
\cline{2-17}

& SRNO$\ddag$
& \multicolumn{1}{c|}{{\color[HTML]{1600FF} 0.028}} & \multicolumn{1}{c|}{{\color[HTML]{1600FF} 0.071}} & \multicolumn{1}{c|}{{\color[HTML]{1600FF} 0.118}} & \multicolumn{1}{c|}{{\color[HTML]{1600FF} 0.049}} & {\color[HTML]{1600FF} 0.129} & \multicolumn{1}{c|}{{\color[HTML]{1600FF} 0.204}} & \multicolumn{1}{c|}{{\color[HTML]{1600FF} 0.079}} & \multicolumn{1}{c|}{{\color[HTML]{1600FF} 0.185}} & \multicolumn{1}{c|}{{\color[HTML]{1600FF} 0.275}} & {\color[HTML]{1600FF} 0.032} & \multicolumn{1}{c|}{{\color[HTML]{1600FF} 0.090}} & \multicolumn{1}{c|}{{\color[HTML]{1600FF} 0.155}} & \multicolumn{1}{c|}{{\color[HTML]{1600FF} 0.011}} & \multicolumn{1}{c|}{{\color[HTML]{1600FF} 0.032}} & {\color[HTML]{1600FF} 0.065}\\

\hline

\multirow{2}{*}{\begin{tabular}[c]{@{}c@{}}Backbone\\ Network\end{tabular}} & \multirow{2}{*}{Methods} & \multicolumn{3}{c|}{Set5}                & \multicolumn{3}{c|}{Set14}               & \multicolumn{3}{c|}{B100}                & \multicolumn{3}{c|}{Urban100}            & \multicolumn{3}{c}{Manga109}            \\
\cline{3-17}

&    & \multicolumn{1}{c|}{$\times$1.8} & \multicolumn{1}{c|}{$\times$2.6} & $\times$3.1 & \multicolumn{1}{c|}{$\times$1.8} & \multicolumn{1}{c|}{$\times$2.7} & $\times$3.5 & \multicolumn{1}{c|}{$\times$1.8} & \multicolumn{1}{c|}{$\times$2.6} & $\times$3.7 & \multicolumn{1}{c|}{$\times$1.8} & \multicolumn{1}{c|}{$\times$3.1} & $\times$3.8 & \multicolumn{1}{c|}{$\times$1.6} & \multicolumn{1}{c|}{$\times$1.8} & $\times$3.4  
\\ 
\cline{1-17}

\multirow{8}{*}{RDN}  
& Metric & \multicolumn{15}{c}{PSNR$\uparrow$}
\\ \cline{2-17}

& SRNO                    & \multicolumn{1}{c|}{39.31}       & \multicolumn{1}{c|}{35.92}       & 34.65       & \multicolumn{1}{c|}{35.35}       & \multicolumn{1}{c|}{31.50}       & 29.73       & \multicolumn{1}{c|}{33.54}       & \multicolumn{1}{c|}{30.28}       & 28.21       & \multicolumn{1}{c|}{34.67}       & \multicolumn{1}{c|}{28.88}       & 27.35       & \multicolumn{1}{c|}{42.39}       & \multicolumn{1}{c|}{40.80}       & 33.30           \\
\cline{2-17}

& SRNO$^\dag$               & \multicolumn{1}{c|}{{\color[HTML]{1600FF} 39.21}}       & \multicolumn{1}{c|}{{\color[HTML]{1600FF} 35.86}} & {\color[HTML]{1600FF} 34.51} & \multicolumn{1}{c|}{{\color[HTML]{1600FF} 35.24}} & \multicolumn{1}{c|}{{\color[HTML]{1600FF} 31.40}} & {\color[HTML]{1600FF} 29.73} & \multicolumn{1}{c|}{{\color[HTML]{1600FF} 33.46}} & \multicolumn{1}{c|}{{\color[HTML]{1600FF} 30.20}} & {\color[HTML]{1600FF} 28.20} & \multicolumn{1}{c|}{{\color[HTML]{000000} 34.31}} & \multicolumn{1}{c|}{{\color[HTML]{000000} 28.63}} & {\color[HTML]{1600FF} 27.32} & \multicolumn{1}{c|}{{\color[HTML]{000000} 42.20}} & \multicolumn{1}{c|}{{\color[HTML]{000000} 40.61}} & {\color[HTML]{1600FF} 33.27}    \\
\cline{2-17}

& SRNO$^\ddag$           & \multicolumn{1}{c|}{{\color[HTML]{1600FF} 39.29}}       & \multicolumn{1}{c|}{{\color[HTML]{1600FF} 35.92}} & {\color[HTML]{FF0000} 34.67} & \multicolumn{1}{c|}{{\color[HTML]{FF0000} 35.37}} & \multicolumn{1}{c|}{{\color[HTML]{FF0000} 31.53}} & {\color[HTML]{1600FF} 29.73} & \multicolumn{1}{c|}{{\color[HTML]{1600FF} 33.52}} & \multicolumn{1}{c|}{{\color[HTML]{1600FF} 30.26}} & {\color[HTML]{1600FF} 28.20} & \multicolumn{1}{c|}{{\color[HTML]{1600FF} 34.59}} & \multicolumn{1}{c|}{{\color[HTML]{1600FF} 28.85}} & {\color[HTML]{1600FF} 27.32} & \multicolumn{1}{c|}{{\color[HTML]{1600FF} 42.39}} & \multicolumn{1}{c|}{{\color[HTML]{1600FF} 40.78}} & {\color[HTML]{1600FF} 33.27}     \\

\cline{2-17}
& Metric & \multicolumn{15}{c}{LPIPS$\downarrow$}
\\ \cline{2-17}

& SRNO 
& \multicolumn{1}{c|}{0.028}                        & \multicolumn{1}{c|}{0.068}                        & \multicolumn{1}{c|}{0.115}                        & \multicolumn{1}{c|}{0.047}                        & 0.123                        & \multicolumn{1}{c|}{0.198}                        & \multicolumn{1}{c|}{0.075}                        & \multicolumn{1}{c|}{0.178}                        & \multicolumn{1}{c|}{0.268}                        & 0.026                        & \multicolumn{1}{c|}{0.080}                        & \multicolumn{1}{c|}{0.141}                        & \multicolumn{1}{c|}{0.010}                        & \multicolumn{1}{c|}{0.030}                        & 0.062                       
\\
\cline{2-17}

& SRNO$\dag$ 
& \multicolumn{1}{c|}{{\color[HTML]{1600FF} 0.028}} & \multicolumn{1}{c|}{{\color[HTML]{1600FF} 0.070}} & \multicolumn{1}{c|}{{\color[HTML]{1600FF} 0.115}} & \multicolumn{1}{c|}{{\color[HTML]{1600FF} 0.048}} & {\color[HTML]{1600FF} 0.127} & \multicolumn{1}{c|}{{\color[HTML]{1600FF} 0.200}} & \multicolumn{1}{c|}{{\color[HTML]{1600FF} 0.076}} & \multicolumn{1}{c|}{{\color[HTML]{1600FF} 0.180}} & \multicolumn{1}{c|}{{\color[HTML]{1600FF} 0.270}} & {\color[HTML]{1600FF} 0.028} & \multicolumn{1}{c|}{{\color[HTML]{1600FF} 0.084}} & \multicolumn{1}{c|}{{\color[HTML]{1600FF} 0.142}} & \multicolumn{1}{c|}{{\color[HTML]{1600FF} 0.011}} & \multicolumn{1}{c|}{{\color[HTML]{1600FF} 0.032}} & {\color[HTML]{1600FF} 0.062}
\\ \cline{2-17}

& SRNO$\ddag$

& \multicolumn{1}{c|}{{\color[HTML]{1600FF} 0.028}} & \multicolumn{1}{c|}{{\color[HTML]{1600FF} 0.069}} & \multicolumn{1}{c|}{{\color[HTML]{1600FF} 0.115}} & \multicolumn{1}{c|}{{\color[HTML]{1600FF} 0.047}} & {\color[HTML]{1600FF} 0.124} & \multicolumn{1}{c|}{{\color[HTML]{1600FF} 0.200}} & \multicolumn{1}{c|}{{\color[HTML]{1600FF} 0.076}} & \multicolumn{1}{c|}{{\color[HTML]{1600FF} 0.180}} & \multicolumn{1}{c|}{{\color[HTML]{1600FF} 0.270}} & {\color[HTML]{1600FF} 0.027} & \multicolumn{1}{c|}{{\color[HTML]{1600FF} 0.081}} & \multicolumn{1}{c|}{{\color[HTML]{1600FF} 0.142}} & \multicolumn{1}{c|}{{\color[HTML]{1600FF} 0.010}} & \multicolumn{1}{c|}{{\color[HTML]{1600FF} 0.031}} & {\color[HTML]{1600FF} 0.062}

\\ 

\bottomrule

\end{tabular}}
\end{table*}

\begin{figure*}[!t]
    \centering
    \includegraphics[width=1.0\textwidth]{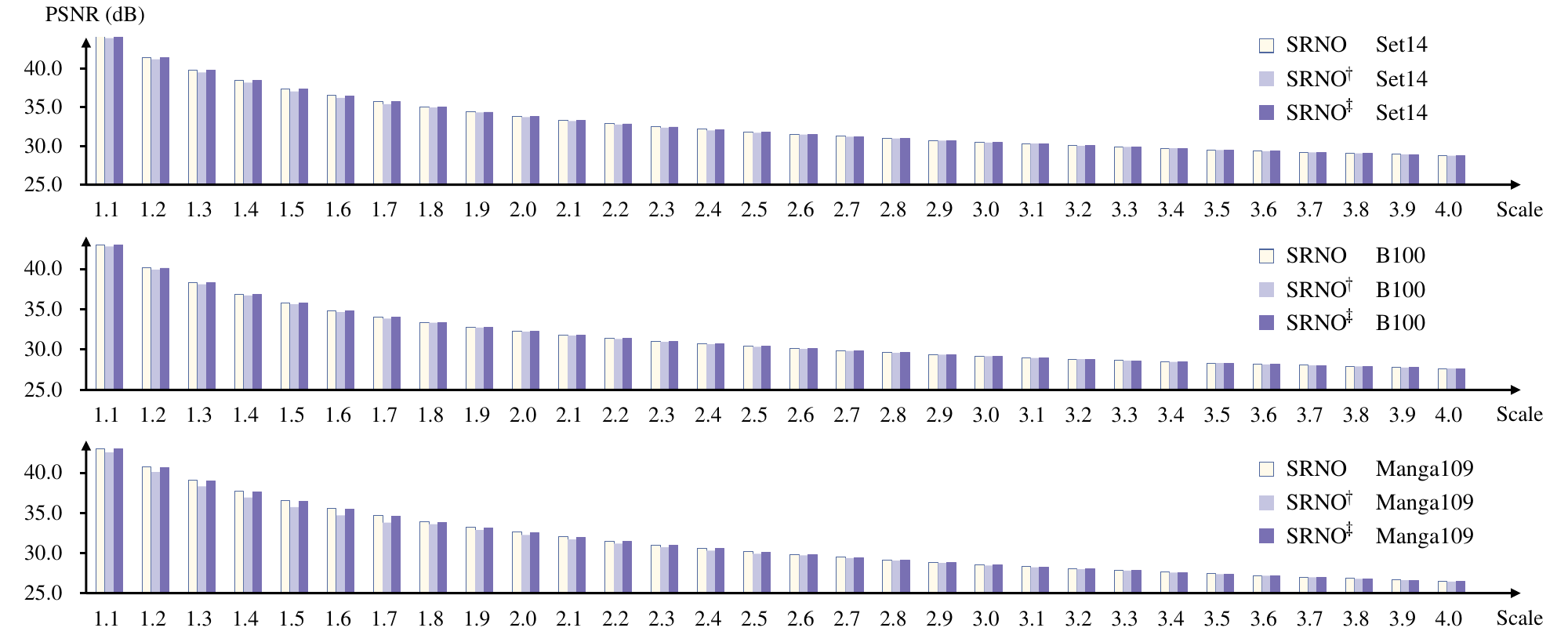}
    \caption{PSNR(dB) comparisons across all scales on different datasets of arbitrary-scale SR model SRNO~\cite{srnowei2023super}, its AnySR variants (through different subnets) highlighted by $^\dag$, and AnySR-retrained version (through the largest network) denoted by $^\ddag$.}
    \label{fig:performance}
\end{figure*}

\begin{table*}[]
\caption{PSNR (dB) and LPIPS comparison for out-of-distribution scales of arbitrary-scale SR methods and their AnySR variants, highlighted by $^\dag$ (through different subnets) and $^\ddag$ (AnySR-retrained version). AnySR performance is marked in {\color{blue}blue} if within 0.15dB (PSNR) / 0.01 (LPIPS) of the baseline, and {\color{red}red} if improved.}
\vspace{-2mm}
\label{tab:out-of-scale-psnr}
\centering
\setlength\tabcolsep{6.5pt}
\resizebox{\textwidth}{!}{
\renewcommand{\arraystretch}{1.5}
\begin{tabular}{c|c|ccccc|ccccc|ccccc}
\toprule
&       & \multicolumn{5}{c|}{Set5}                                                                                               & \multicolumn{5}{c|}{  Set14}                                                                                                         & \multicolumn{5}{c}{Urban100}                         \\ \cline{3-17} 
\multirow{-2}{*}{  Backbone} & \multirow{-2}{*}{  Method} & \multicolumn{1}{c|}{   $\times$6}    & \multicolumn{1}{c|}{   $\times$12}   & \multicolumn{1}{c|}{   $\times$18}   & \multicolumn{1}{c|}{   $\times$24}   &    $\times$30   & \multicolumn{1}{c|}{   $\times$6}    & \multicolumn{1}{c|}{   $\times$12}   & \multicolumn{1}{c|}{   $\times$18}   & \multicolumn{1}{c|}{   $\times$24}   &    $\times$30   & \multicolumn{1}{c|}{   $\times$6}    & \multicolumn{1}{c|}{   $\times$12}   & \multicolumn{1}{c|}{   $\times$18}   & \multicolumn{1}{c|}{   $\times$24}   &    $\times$30   \\ \hline

\multirow{18}{*}{EDSR}   
& Metric & \multicolumn{15}{c}{PSNR$\uparrow$}
\\
\cline{2-17}
                            
&   MetaSR                   & \multicolumn{1}{c|}{28.48} & \multicolumn{1}{c|}{  24.17} & \multicolumn{1}{c|}{  22.01} & \multicolumn{1}{c|}{  20.97} &   20.06 & \multicolumn{1}{c|}{  26.30} & \multicolumn{1}{c|}{  22.99} & \multicolumn{1}{c|}{  21.44} & \multicolumn{1}{c|}{  20.43} &   19.60 & \multicolumn{1}{c|}{  23.57} & \multicolumn{1}{c|}{  20.74} & \multicolumn{1}{c|}{  19.47} & \multicolumn{1}{c|}{  18.71} &   18.17 \\ \cline{2-17}

&   MetaSR$\ddag$                  & \multicolumn{1}{c|}{{\color[HTML]{FF0000} 28.50}} & \multicolumn{1}{c|}{{\color[HTML]{FF0000} 24.20}} & \multicolumn{1}{c|}{  {\color{blue}21.96}} & \multicolumn{1}{c|}{{\color{blue}  20.94}} &   {\color{blue}20.01} & \multicolumn{1}{c|}{{\color[HTML]{FF0000} 26.31}} & \multicolumn{1}{c|}{\color{blue}  22.99} & \multicolumn{1}{c|}{{\color[HTML]{FF0000} 21.45}} & \multicolumn{1}{c|}{ {\color{blue} 20.42}} &   {\color{blue}19.57} & \multicolumn{1}{c|}{{\color[HTML]{FF0000} 23.61}} & \multicolumn{1}{c|}{{\color[HTML]{FF0000} 20.75}} & \multicolumn{1}{c|}{ {\color{blue} 19.47}} & \multicolumn{1}{c|}{ {\color{blue} 18.71}} &   {\color{blue}18.15} \\ \cline{2-17}

 &   LIIF                     & \multicolumn{1}{c|}{  28.96} & \multicolumn{1}{c|}{  24.56} & \multicolumn{1}{c|}{  22.48} & \multicolumn{1}{c|}{  21.37} &   20.58 & \multicolumn{1}{c|}{  26.45} & \multicolumn{1}{c|}{  23.16} & \multicolumn{1}{c|}{  21.61} & \multicolumn{1}{c|}{  20.66} &   19.81 & \multicolumn{1}{c|}{  23.79} & \multicolumn{1}{c|}{  20.89} & \multicolumn{1}{c|}{  19.62} & \multicolumn{1}{c|}{  18.84} &   18.29 \\ \cline{2-17}

&   LIIF$\ddag$                
& \multicolumn{1}{c|}{{\color[HTML]{1600FF} 28.94}} & \multicolumn{1}{c|}{{\color[HTML]{FF0000} 24.59}} & \multicolumn{1}{c|}{{\color[HTML]{FF0000} 22.51}} & \multicolumn{1}{c|}{{\color[HTML]{FF0000} 21.39}} & {\color[HTML]{1600FF} 20.56} & \multicolumn{1}{c|}{{\color[HTML]{1600FF} 26.45}} & \multicolumn{1}{c|}{{\color[HTML]{FF0000} 23.19}} & \multicolumn{1}{c|}{{\color[HTML]{FF0000} 21.61}} & \multicolumn{1}{c|}{{\color[HTML]{1600FF} 20.65}} & {\color[HTML]{FF0000} 19.83} & \multicolumn{1}{c|}{{\color[HTML]{1600FF} 23.79}} & \multicolumn{1}{c|}{{\color[HTML]{1600FF} 20.89}} & \multicolumn{1}{c|}{{\color[HTML]{1600FF} 19.61}} & \multicolumn{1}{c|}{{\color[HTML]{1600FF} 18.84}} & {\color[HTML]{1600FF} 18.29}\\ \cline{2-17}

&   ArbSR                    & \multicolumn{1}{c|}{  28.39} & \multicolumn{1}{c|}{  23.76} & \multicolumn{1}{c|}{  21.69} & \multicolumn{1}{c|}{  20.64} &   19.75 & \multicolumn{1}{c|}{  26.19} & \multicolumn{1}{c|}{  22.57} & \multicolumn{1}{c|}{  20.99} & \multicolumn{1}{c|}{  19.93} &   19.22 & \multicolumn{1}{c|}{  23.47} & \multicolumn{1}{c|}{  20.37} & \multicolumn{1}{c|}{  19.08} & \multicolumn{1}{c|}{  18.33} &   17.82 \\ \cline{2-17}

&   ArbSR$\ddag$                   & \multicolumn{1}{c|}{  {\color{blue}28.37}} & \multicolumn{1}{c|}{  {\color{blue}23.73}} & \multicolumn{1}{c|}{  {\color{blue}21.66}} & \multicolumn{1}{c|}{  {\color{blue}20.62}} &   {\color{blue}19.73} & \multicolumn{1}{c|}{{\color{blue}  26.15}} & \multicolumn{1}{c|}{  {\color{blue}22.56}} & \multicolumn{1}{c|}{  {\color{blue}20.96}} & \multicolumn{1}{c|}{  {\color{blue}19.91}} &   {\color{blue}19.19} & \multicolumn{1}{c|}{  {\color{blue}23.46}} & \multicolumn{1}{c|}{  {\color{blue}20.36}} & \multicolumn{1}{c|}{  {\color{blue}19.06}} & \multicolumn{1}{c|}{ {\color{blue} 18.32}} &   {\color{blue}17.80} \\ \cline{2-17}

  &   SRNO                     & \multicolumn{1}{c|}{  29.06} & \multicolumn{1}{c|}{  24.62} & \multicolumn{1}{c|}{  22.52} & \multicolumn{1}{c|}{  21.37} &   20.54 & \multicolumn{1}{c|}{  26.55} & \multicolumn{1}{c|}{  23.19} & \multicolumn{1}{c|}{  21.65} & \multicolumn{1}{c|}{  20.68} &   19.77 & \multicolumn{1}{c|}{  24.08} & \multicolumn{1}{c|}{  21.10} & \multicolumn{1}{c|}{  19.75} & \multicolumn{1}{c|}{  18.96} &   18.40 \\ \cline{2-17}

&   SRNO$\ddag$                     
& \multicolumn{1}{c|}{{\color[HTML]{1600FF} 29.05}} & \multicolumn{1}{c|}{{\color[HTML]{1600FF} 24.57}} & \multicolumn{1}{c|}{{\color[HTML]{1600FF} 22.51}} & \multicolumn{1}{c|}{{\color[HTML]{1600FF} 21.31}} & {\color[HTML]{1600FF} 20.52} & \multicolumn{1}{c|}{{\color[HTML]{1600FF} 26.55}} & \multicolumn{1}{c|}{{\color[HTML]{1600FF} 23.15}} & \multicolumn{1}{c|}{{\color[HTML]{1600FF} 21.61}} & \multicolumn{1}{c|}{{\color[HTML]{1600FF} 20.65}} & {\color[HTML]{1600FF} 19.73} & \multicolumn{1}{c|}{{\color[HTML]{1600FF} 24.01}} & \multicolumn{1}{c|}{{\color[HTML]{1600FF} 21.01}} & \multicolumn{1}{c|}{{\color[HTML]{1600FF} 19.67}} & \multicolumn{1}{c|}{{\color[HTML]{1600FF} 18.88}} & {\color[HTML]{1600FF} 18.33} \\ 
\cline{2-17}
& Metric & \multicolumn{15}{c}{LPIPS$\downarrow$}
\\
\cline{2-17}

& MetaSR                   & \multicolumn{1}{c|}{0.123}                        & \multicolumn{1}{c|}{0.211}                        & \multicolumn{1}{c|}{0.263}                        & \multicolumn{1}{c|}{0.281}                        & 0.297                        & \multicolumn{1}{c|}{0.194}                        & \multicolumn{1}{c|}{0.266}                        & \multicolumn{1}{c|}{0.302}                        & \multicolumn{1}{c|}{0.319}                        & 0.333                        & \multicolumn{1}{c|}{0.224}                        & \multicolumn{1}{c|}{0.340}                        & \multicolumn{1}{c|}{0.383}                        & \multicolumn{1}{c|}{0.400}                        & 0.410                        \\ \cline{2-17}

& MetaSR$\ddag$                   & \multicolumn{1}{c|}{{\color[HTML]{1600FF} 0.123}} & \multicolumn{1}{c|}{{\color[HTML]{FF0000} 0.207}} & \multicolumn{1}{c|}{{\color[HTML]{FF0000} 0.262}} & \multicolumn{1}{c|}{{\color[HTML]{1600FF} 0.282}} & {\color[HTML]{1600FF} 0.297} & \multicolumn{1}{c|}{{\color[HTML]{FF0000} 0.193}} & \multicolumn{1}{c|}{{\color[HTML]{FF0000} 0.264}} & \multicolumn{1}{c|}{{\color[HTML]{FF0000} 0.299}} & \multicolumn{1}{c|}{{\color[HTML]{FF0000} 0.315}} & {\color[HTML]{FF0000} 0.330} & \multicolumn{1}{c|}{{\color[HTML]{FF0000} 0.222}} & \multicolumn{1}{c|}{{\color[HTML]{FF0000} 0.338}} & \multicolumn{1}{c|}{{\color[HTML]{FF0000} 0.381}} & \multicolumn{1}{c|}{{\color[HTML]{FF0000} 0.396}} & {\color[HTML]{FF0000} 0.407} \\ \cline{2-17}

& LIIF                     & \multicolumn{1}{c|}{0.204}                        & \multicolumn{1}{c|}{0.379}                        & \multicolumn{1}{c|}{0.510}                        & \multicolumn{1}{c|}{0.587}                        & 0.626                        & \multicolumn{1}{c|}{0.330}                        & \multicolumn{1}{c|}{0.487}                        & \multicolumn{1}{c|}{0.567}                        & \multicolumn{1}{c|}{0.625}                        & 0.672                        & \multicolumn{1}{c|}{0.287}                        & \multicolumn{1}{c|}{0.489}                        & \multicolumn{1}{c|}{0.603}                        & \multicolumn{1}{c|}{0.668}                        & 0.709                        \\ \cline{2-17}

 & LIIF$\ddag$                  & \multicolumn{1}{c|}{{\color[HTML]{1600FF} 0.210}} & \multicolumn{1}{c|}{{\color[HTML]{1600FF} 0.389}} & \multicolumn{1}{c|}{{\color[HTML]{1600FF} 0.513}} & \multicolumn{1}{c|}{{\color[HTML]{1600FF} 0.598}} & {\color[HTML]{1600FF} 0.635} & \multicolumn{1}{c|}{{\color[HTML]{1600FF} 0.336}} & \multicolumn{1}{c|}{{\color[HTML]{1600FF} 0.493}} & \multicolumn{1}{c|}{{\color[HTML]{1600FF} 0.573}} & \multicolumn{1}{c|}{{\color[HTML]{1600FF} 0.630}} & {\color[HTML]{1600FF} 0.676} & \multicolumn{1}{c|}{{\color[HTML]{1600FF} 0.292}} & \multicolumn{1}{c|}{{\color[HTML]{1600FF} 0.495}} & \multicolumn{1}{c|}{{\color[HTML]{1600FF} 0.609}} & \multicolumn{1}{c|}{{\color[HTML]{1600FF} 0.672}} & {\color[HTML]{1600FF} 0.712}                             \\ \cline{2-17}

& ArbSR                    & \multicolumn{1}{c|}{0.133}                        & \multicolumn{1}{c|}{0.265}                        & \multicolumn{1}{c|}{0.319}                        & \multicolumn{1}{c|}{0.333}                        & 0.343                        & \multicolumn{1}{c|}{0.180}                        & \multicolumn{1}{c|}{0.279}                        & \multicolumn{1}{c|}{0.318}                        & \multicolumn{1}{c|}{0.330}                        & 0.339                        & \multicolumn{1}{c|}{0.229}                        & \multicolumn{1}{c|}{0.350}                        & \multicolumn{1}{c|}{0.389}                        & \multicolumn{1}{c|}{0.401}                        & 0.404                        \\ \cline{2-17}

& ArbSR$\ddag$                  & \multicolumn{1}{c|}{{\color[HTML]{1600FF} 0.136}} & \multicolumn{1}{c|}{{\color[HTML]{1600FF} 0.269}} & \multicolumn{1}{c|}{{\color[HTML]{1600FF} 0.321}} & \multicolumn{1}{c|}{{\color[HTML]{1600FF} 0.337}} & {\color[HTML]{1600FF} 0.348} & \multicolumn{1}{c|}{{\color[HTML]{1600FF} 0.183}} & \multicolumn{1}{c|}{{\color[HTML]{1600FF} 0.281}} & \multicolumn{1}{c|}{{\color[HTML]{1600FF} 0.319}} & \multicolumn{1}{c|}{{\color[HTML]{1600FF} 0.331}} & {\color[HTML]{1600FF} 0.341} & \multicolumn{1}{c|}{{\color[HTML]{1600FF} 0.233}} & \multicolumn{1}{c|}{{\color[HTML]{1600FF} 0.352}} & \multicolumn{1}{c|}{{\color[HTML]{1600FF} 0.392}} & \multicolumn{1}{c|}{{\color[HTML]{1600FF} 0.403}} & {\color[HTML]{1600FF} 0.406}                     \\ \cline{2-17}

 & SRNO                     & \multicolumn{1}{c|}{0.210}                        & \multicolumn{1}{c|}{0.394}                        & \multicolumn{1}{c|}{0.525}                        & \multicolumn{1}{c|}{0.619}                        & 0.636                        & \multicolumn{1}{c|}{0.332}                        & \multicolumn{1}{c|}{0.496}                        & \multicolumn{1}{c|}{0.579}                        & \multicolumn{1}{c|}{0.637}                        & 0.687                        & \multicolumn{1}{c|}{0.281}                        & \multicolumn{1}{c|}{0.491}                        & \multicolumn{1}{c|}{0.610}                        & \multicolumn{1}{c|}{0.677}                        & 0.719                        \\ \cline{2-17}

& SRNO$\ddag$                    & \multicolumn{1}{c|}{{\color[HTML]{FF0000} 0.209}} & \multicolumn{1}{c|}{{\color[HTML]{FF0000} 0.392}} & \multicolumn{1}{c|}{{\color[HTML]{1600FF} 0.529}} & \multicolumn{1}{c|}{{\color[HTML]{1600FF} 0.620}} & {\color[HTML]{1600FF} 0.641} & \multicolumn{1}{c|}{{\color[HTML]{FF0000} 0.329}} & \multicolumn{1}{c|}{{\color[HTML]{FF0000} 0.491}} & \multicolumn{1}{c|}{{\color[HTML]{FF0000} 0.578}} & \multicolumn{1}{c|}{{\color[HTML]{FF0000} 0.636}} & {\color[HTML]{1600FF} 0.687} & \multicolumn{1}{c|}{{\color[HTML]{FF0000} 0.279}} & \multicolumn{1}{c|}{{\color[HTML]{FF0000} 0.485}} & \multicolumn{1}{c|}{{\color[HTML]{FF0000} 0.605}} & \multicolumn{1}{c|}{{\color[HTML]{FF0000} 0.674}} & {\color[HTML]{FF0000} 0.719}         
\\ \hline

\multirow{18}{*}{  RDN}  
& Metric & \multicolumn{15}{c}{PSNR$\uparrow$}
\\
\cline{2-17}

  &   MetaSR                   & \multicolumn{1}{c|}{  28.95} & \multicolumn{1}{c|}{  24.36} & \multicolumn{1}{c|}{  22.04} & \multicolumn{1}{c|}{  20.99} &   20.01 & \multicolumn{1}{c|}{  26.57} & \multicolumn{1}{c|}{  23.10} & \multicolumn{1}{c|}{  21.50} & \multicolumn{1}{c|}{  20.51} &   19.55 & \multicolumn{1}{c|}{  24.06} & \multicolumn{1}{c|}{  20.96} & \multicolumn{1}{c|}{  19.62} & \multicolumn{1}{c|}{  18.83} &   18.25 \\ \cline{2-17}

  &   MetaSR$\ddag$                   & \multicolumn{1}{c|}{{\color[HTML]{FF0000} 28.96}} & \multicolumn{1}{c|}{{\color[HTML]{1600FF} 24.28}} & \multicolumn{1}{c|}{{\color[HTML]{1600FF} 22.02}} & \multicolumn{1}{c|}{{\color[HTML]{1600FF} 20.98}} & {\color[HTML]{FF0000} 20.04} & \multicolumn{1}{c|}{{\color[HTML]{1600FF} 26.55}} & \multicolumn{1}{c|}{{\color[HTML]{1600FF} 23.03}} & \multicolumn{1}{c|}{{\color[HTML]{1600FF} 21.44}} & \multicolumn{1}{c|}{{\color[HTML]{1600FF} 20.47}} & {\color[HTML]{1600FF} 19.51} & \multicolumn{1}{c|}{{\color[HTML]{1600FF} 24.03}} & \multicolumn{1}{c|}{{\color[HTML]{1600FF} 20.91}} & \multicolumn{1}{c|}{{\color[HTML]{1600FF} 19.52}} & \multicolumn{1}{c|}{{\color[HTML]{1600FF} 18.73}} & {\color[HTML]{1600FF} 18.15} \\ \cline{2-17}

&   LIIF                     & \multicolumn{1}{c|}{  29.15} & \multicolumn{1}{c|}{  24.86} & \multicolumn{1}{c|}{  22.66} & \multicolumn{1}{c|}{  21.50} &   20.57 & \multicolumn{1}{c|}{  26.64} & \multicolumn{1}{c|}{  23.24} & \multicolumn{1}{c|}{  21.73} & \multicolumn{1}{c|}{  20.78} &   19.85 & \multicolumn{1}{c|}{  24.20} & \multicolumn{1}{c|}{  21.15} & \multicolumn{1}{c|}{  19.80} & \multicolumn{1}{c|}{  19.00} &   18.44 \\ \cline{2-17}

  &   LIIF$\ddag$                    & \multicolumn{1}{c|}{{\color[HTML]{1600FF} 29.14}} & \multicolumn{1}{c|}{{\color[HTML]{FF0000} 24.89}} & \multicolumn{1}{c|}{{\color[HTML]{1600FF} 22.64}} & \multicolumn{1}{c|}{{\color[HTML]{1600FF} 21.49}} & {\color[HTML]{FF0000} 20.62} & \multicolumn{1}{c|}{{\color[HTML]{FF0000} 26.65}} & \multicolumn{1}{c|}{{\color[HTML]{FF0000} 23.28}} & \multicolumn{1}{c|}{{\color[HTML]{FF0000} 21.76}} & \multicolumn{1}{c|}{{\color[HTML]{1600FF} 20.76}} & {\color[HTML]{FF0000} 19.90} & \multicolumn{1}{c|}{{\color[HTML]{1600FF} 24.18}} & \multicolumn{1}{c|}{{\color[HTML]{1600FF} 21.15}} & \multicolumn{1}{c|}{{\color[HTML]{FF0000} 19.81}} & \multicolumn{1}{c|}{{\color[HTML]{FF0000} 19.02}} & {\color[HTML]{FF0000} 18.44} \\ \cline{2-17}

&   ArbSR                    & \multicolumn{1}{c|}{  28.45} & \multicolumn{1}{c|}{  23.57} & \multicolumn{1}{c|}{  21.53} & \multicolumn{1}{c|}{  20.45} &   19.55 & \multicolumn{1}{c|}{  26.18} & \multicolumn{1}{c|}{  22.44} & \multicolumn{1}{c|}{  20.83} & \multicolumn{1}{c|}{  19.79} &   19.09 & \multicolumn{1}{c|}{  23.55} & \multicolumn{1}{c|}{  20.27} & \multicolumn{1}{c|}{  18.97} & \multicolumn{1}{c|}{  18.24} &   17.73 \\ \cline{2-17}

  &   ArbSR$\ddag$                  & \multicolumn{1}{c|}{{\color[HTML]{FF0000} 28.51}} & \multicolumn{1}{c|}{{\color[HTML]{1600FF} 23.57}} & \multicolumn{1}{c|}{{\color[HTML]{1600FF} 21.52}} & \multicolumn{1}{c|}{{\color[HTML]{1600FF} 20.43}} & {\color[HTML]{FF0000} 19.56} & \multicolumn{1}{c|}{{\color[HTML]{FF0000} 26.22}} & \multicolumn{1}{c|}{{\color[HTML]{FF0000} 22.46}} & \multicolumn{1}{c|}{{\color[HTML]{FF0000} 20.84}} & \multicolumn{1}{c|}{{\color[HTML]{1600FF} 19.78}} & {\color[HTML]{1600FF} 19.09} & \multicolumn{1}{c|}{{\color[HTML]{FF0000} 23.60}} & \multicolumn{1}{c|}{{\color[HTML]{FF0000} 20.29}} & \multicolumn{1}{c|}{{\color[HTML]{1600FF} 18.97}} & \multicolumn{1}{c|}{{\color[HTML]{1600FF} 18.23}} & {\color[HTML]{1600FF} 17.73}  \\ \cline{2-17}

 &   SRNO                     & \multicolumn{1}{c|}{  29.40} & \multicolumn{1}{c|}{  24.94} & \multicolumn{1}{c|}{  22.73} & \multicolumn{1}{c|}{  21.50} &   20.65 & \multicolumn{1}{c|}{  26.76} & \multicolumn{1}{c|}{  23.33} & \multicolumn{1}{c|}{  21.82} & \multicolumn{1}{c|}{  20.78} &   19.96 & \multicolumn{1}{c|}{  24.42} & \multicolumn{1}{c|}{  21.35} & \multicolumn{1}{c|}{  19.95} & \multicolumn{1}{c|}{  19.15} &   18.55 \\ \cline{2-17}

&   SRNO$\ddag$                    & \multicolumn{1}{c|}{{\color[HTML]{FF0000} 29.45}} & \multicolumn{1}{c|}{{\color[HTML]{1600FF} 24.87}} & \multicolumn{1}{c|}{{\color[HTML]{FF0000} 22.76}} & \multicolumn{1}{c|}{{\color[HTML]{1600FF} 21.45}} & {\color[HTML]{1600FF} 20.62} & \multicolumn{1}{c|}{{\color[HTML]{1600FF} 26.74}} & \multicolumn{1}{c|}{{\color[HTML]{1600FF} 23.26}} & \multicolumn{1}{c|}{{\color[HTML]{1600FF} 21.75}} & \multicolumn{1}{c|}{{\color[HTML]{FF0000} 20.79}} & {\color[HTML]{1600FF} 19.88} & \multicolumn{1}{c|}{{\color[HTML]{1600FF} 24.38}} & \multicolumn{1}{c|}{{\color[HTML]{1600FF} 21.27}} & \multicolumn{1}{c|}{{\color[HTML]{1600FF} 19.86}} & \multicolumn{1}{c|}{{\color[HTML]{1600FF} 19.06}} & {\color[HTML]{1600FF} 18.46} \\ \cline{2-17}
& Metric & \multicolumn{15}{c}{LPIPS$\downarrow$}
\\
\cline{2-17}

 & MetaSR                   & \multicolumn{1}{c|}{0.114}                        & \multicolumn{1}{c|}{0.195}                        & \multicolumn{1}{c|}{0.248}                        & \multicolumn{1}{c|}{0.272}                        & 0.295                        & \multicolumn{1}{c|}{0.184}                        & \multicolumn{1}{c|}{0.257}                        & \multicolumn{1}{c|}{0.293}                        & \multicolumn{1}{c|}{0.308}                        & 0.326                        & \multicolumn{1}{c|}{0.199}                        & \multicolumn{1}{c|}{0.320}                        & \multicolumn{1}{c|}{0.370}                        & \multicolumn{1}{c|}{0.389}                        & 0.403                        \\ \cline{2-17}

& MetaSR$\ddag$                  & \multicolumn{1}{c|}{{\color[HTML]{FF0000} 0.106}} & \multicolumn{1}{c|}{{\color[HTML]{FF0000} 0.192}} & \multicolumn{1}{c|}{{\color[HTML]{FF0000} 0.244}} & \multicolumn{1}{c|}{{\color[HTML]{FF0000} 0.267}} & {\color[HTML]{FF0000} 0.292} & \multicolumn{1}{c|}{{\color[HTML]{FF0000} 0.182}} & \multicolumn{1}{c|}{{\color[HTML]{FF0000} 0.257}} & \multicolumn{1}{c|}{{\color[HTML]{FF0000} 0.292}} & \multicolumn{1}{c|}{{\color[HTML]{FF0000} 0.306}} & {\color[HTML]{FF0000} 0.326} & \multicolumn{1}{c|}{{\color[HTML]{FF0000} 0.196}} & \multicolumn{1}{c|}{{\color[HTML]{FF0000} 0.317}} & \multicolumn{1}{c|}{{\color[HTML]{FF0000} 0.368}} & \multicolumn{1}{c|}{{\color[HTML]{FF0000} 0.388}} & {\color[HTML]{FF0000} 0.402} \\ \cline{2-17}

& LIIF                     & \multicolumn{1}{c|}{0.197}                        & \multicolumn{1}{c|}{0.361}                        & \multicolumn{1}{c|}{0.491}                        & \multicolumn{1}{c|}{0.566}                        & 0.601                        & \multicolumn{1}{c|}{0.319}                        & \multicolumn{1}{c|}{0.472}                        & \multicolumn{1}{c|}{0.551}                        & \multicolumn{1}{c|}{0.604}                        & 0.657                        & \multicolumn{1}{c|}{0.263}                        & \multicolumn{1}{c|}{0.457}                        & \multicolumn{1}{c|}{0.574}                        & \multicolumn{1}{c|}{0.644}                        & 0.690                        \\ \cline{2-17}

& LIIF$\ddag$                    & \multicolumn{1}{c|}{0.210}                        & \multicolumn{1}{c|}{0.372}                        & \multicolumn{1}{c|}{{\color[HTML]{1600FF} 0.501}} & \multicolumn{1}{c|}{0.579}                        & {\color[HTML]{1600FF} 0.610} & \multicolumn{1}{c|}{0.330}                        & \multicolumn{1}{c|}{0.485}                        & \multicolumn{1}{c|}{{\color[HTML]{1600FF} 0.560}} & \multicolumn{1}{c|}{0.615}                        & {\color[HTML]{1600FF} 0.666} & \multicolumn{1}{c|}{0.278}                        & \multicolumn{1}{c|}{0.475}                        & \multicolumn{1}{c|}{0.590}                        & \multicolumn{1}{c|}{0.657}                        & {\color[HTML]{1600FF} 0.701}      \\ \cline{2-17}

& ArbSR                    & \multicolumn{1}{c|}{0.132}                        & \multicolumn{1}{c|}{0.267}                        & \multicolumn{1}{c|}{0.316}                        & \multicolumn{1}{c|}{0.333}                        & 0.340                        & \multicolumn{1}{c|}{0.179}                        & \multicolumn{1}{c|}{0.278}                        & \multicolumn{1}{c|}{0.316}                        & \multicolumn{1}{c|}{0.327}                        & 0.335                        & \multicolumn{1}{c|}{0.229}                        & \multicolumn{1}{c|}{0.351}                        & \multicolumn{1}{c|}{0.388}                        & \multicolumn{1}{c|}{0.398}                        & 0.401                        \\ \cline{2-17}

& ArbSR$\ddag$                  & \multicolumn{1}{c|}{{\color[HTML]{FF0000} 0.128}} & \multicolumn{1}{c|}{{\color[HTML]{FF0000} 0.266}} & \multicolumn{1}{c|}{{\color[HTML]{1600FF} 0.317}} & \multicolumn{1}{c|}{{\color[HTML]{1600FF} 0.334}} & {\color[HTML]{1600FF} 0.341} & \multicolumn{1}{c|}{{\color[HTML]{FF0000} 0.176}} & \multicolumn{1}{c|}{{\color[HTML]{FF0000} 0.278}} & \multicolumn{1}{c|}{{\color[HTML]{1600FF} 0.316}} & \multicolumn{1}{c|}{{\color[HTML]{1600FF} 0.327}} & {\color[HTML]{1600FF} 0.336} & \multicolumn{1}{c|}{{\color[HTML]{FF0000} 0.224}} & \multicolumn{1}{c|}{{\color[HTML]{FF0000} 0.349}} & \multicolumn{1}{c|}{{\color[HTML]{FF0000} 0.387}} & \multicolumn{1}{c|}{{\color[HTML]{FF0000} 0.398}} & {\color[HTML]{FF0000} 0.400} \\ \cline{2-17}

& SRNO                     & \multicolumn{1}{c|}{0.206}                        & \multicolumn{1}{c|}{0.387}                        & \multicolumn{1}{c|}{0.514}                        & \multicolumn{1}{c|}{0.591}                        & 0.628                        & \multicolumn{1}{c|}{0.323}                        & \multicolumn{1}{c|}{0.485}                        & \multicolumn{1}{c|}{0.565}                        & \multicolumn{1}{c|}{0.623}                        & 0.674                        & \multicolumn{1}{c|}{0.267}                        & \multicolumn{1}{c|}{0.474}                        & \multicolumn{1}{c|}{0.595}                        & \multicolumn{1}{c|}{0.665}                        & 0.711                        \\ \cline{2-17}

& SRNO$\ddag$                    & \multicolumn{1}{c|}{{\color[HTML]{FF0000} 0.203}} & \multicolumn{1}{c|}{{\color[HTML]{FF0000} 0.378}} & \multicolumn{1}{c|}{{\color[HTML]{FF0000} 0.503}} & \multicolumn{1}{c|}{{\color[HTML]{FF0000} 0.582}} & {\color[HTML]{FF0000} 0.624} & \multicolumn{1}{c|}{{\color[HTML]{FF0000} 0.322}} & \multicolumn{1}{c|}{{\color[HTML]{FF0000} 0.479}} & \multicolumn{1}{c|}{{\color[HTML]{FF0000} 0.559}} & \multicolumn{1}{c|}{{\color[HTML]{FF0000} 0.612}} & {\color[HTML]{FF0000} 0.663} & \multicolumn{1}{c|}{{\color[HTML]{FF0000} 0.263}} & \multicolumn{1}{c|}{{\color[HTML]{FF0000} 0.461}} & \multicolumn{1}{c|}{{\color[HTML]{FF0000} 0.582}} & \multicolumn{1}{c|}{{\color[HTML]{FF0000} 0.654}} & {\color[HTML]{FF0000} 0.703} \\ \bottomrule

\end{tabular}}
\end{table*}

\subsection{Main Results}
\subsubsection{Quantitative Results}
Extensive experiments are conducted using in-distribution and out-of-distribution scales, and we analyze the quantitative performance below to show the efficacy of our proposed AnySR.

\textbf{In-Distribution Scales.}
Table\,\ref{tab:psnr} presents the PSNR and LPIPS performance on integer scales, showing the comparison between AnySR variants with different subnets dealing with individual scales and AnySR-retrained largest network for all scales. 
Additionally, Table\,\ref{tab:result-psnr-detail-edsr} shows the quantitative results on fractional scales, with three fractional scales randomly selected for each dataset on the SRNO model~\cite{srnowei2023super}.
More PSNR results of AnySR across all 30 scales ($\times$1.1–$\times$4.0) are visualized in Fig.\,\ref{fig:performance}, to provide a comprehensive overview of the results.

From Table\,\ref{tab:psnr} and Table\,\ref{tab:result-psnr-detail-edsr}, there is a limited performance drop at smaller scales with an average decrease of 0.15dB (PSNR) / 0.05 (LPIPS). With only this minimal performance loss, we can transform existing methods into an any-scale, any-resource implementation by dynamically selecting the network based on available computational resources, and we believe this trade-off is worthwhile.
When selecting the entire network, AnySR shows only a negligible performance drop across all scales, with an average PSNR difference of merely 0.05dB, and the average LPIPS difference is even less than 0.01. 
This indicates that AnySR-retrained models can effectively preserve the original model's performance during the rebuilding process, which can also be visually observed in Fig.\,\ref{fig:performance}. 
Such good performance is closely related to the reset probability $p$ in Line 4 of Algorithm\,\ref{alg:training}, investigated in Sec.\,\ref{sec:ablation}.

\textbf{Out-of-Distribution Scales.}
Table\,\ref{tab:out-of-scale-psnr} presents the performance of out-of-distribution scales. 
For scales near the training range ($\times$6, $\times$12), AnySR shows overall metric improvements. However, as the scales deviate from the training range, there is a noticeable performance drop.
Nevertheless, we observe across all scales, the AnySR-retrained version achieves better LPIPS scores, indicating that the images generated by AnySR maintain a high level of visual quality.
Overall, these results demonstrate that our AnySR preserves, and in some cases surpasses, the reconstruction capability of the original network, allowing for stable, high-quality outputs across devices with varying computational resources through the dynamic selection of different reconstruction networks.

%
%
%

\begin{figure*}[!t]
    \centering
    \includegraphics[width=0.98\textwidth]{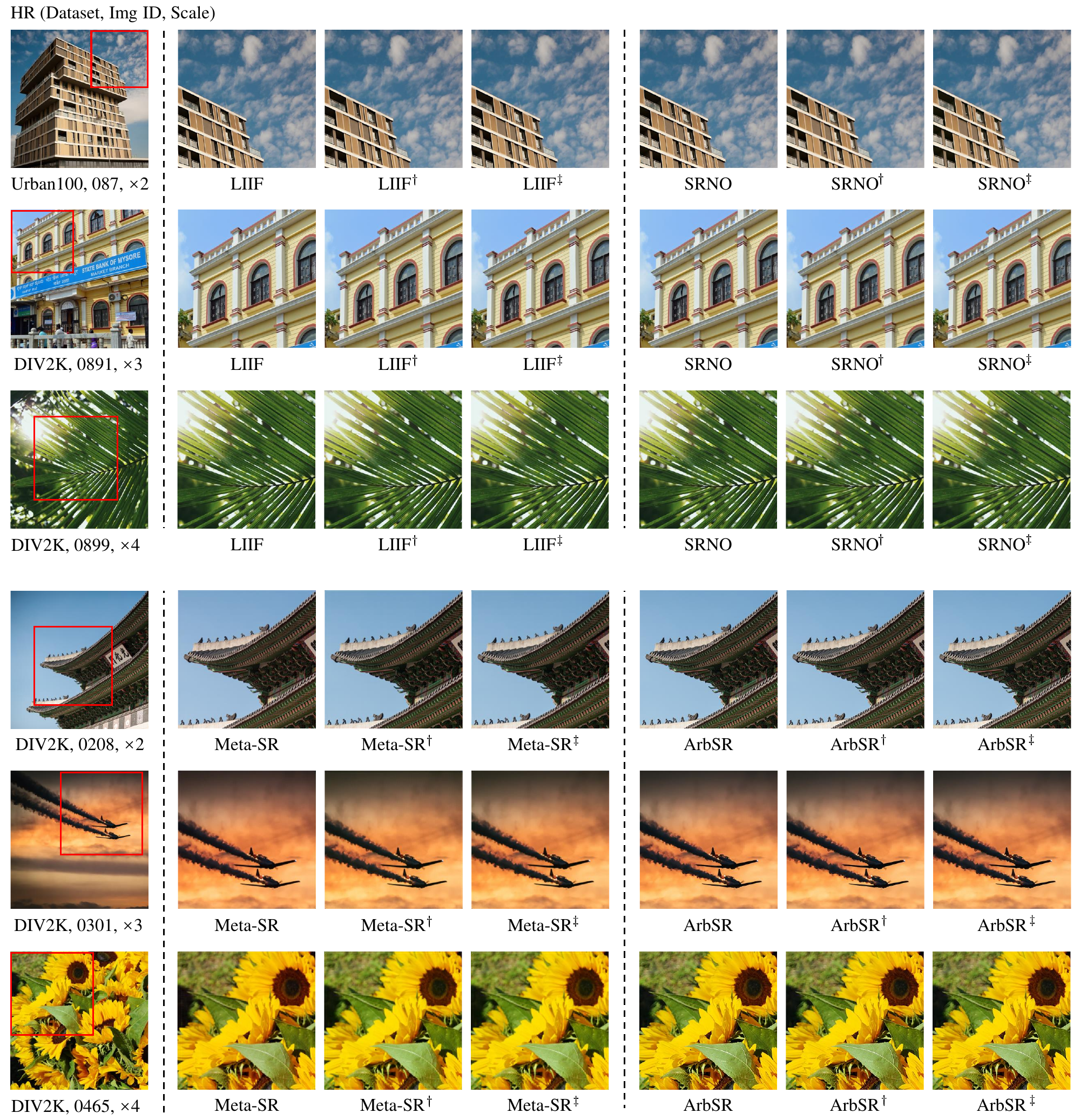}
    \caption{Visualization of existing arbitrary-scale SR models LIIF~\cite{chen2021learning}, SRNO~\cite{srnowei2023super}, Meta-SR~\cite{hu2019meta}, ArbSR~\cite{wang2021learning}, their AnySR variants (through different subnets) highlighted by $^\dag$, and AnySR-retrained version (through the largest network) denoted as $^\ddag$.}
    \label{fig:visual}
\end{figure*}

\begin{figure*}[!t]
    \centering
    \includegraphics[width=\textwidth]{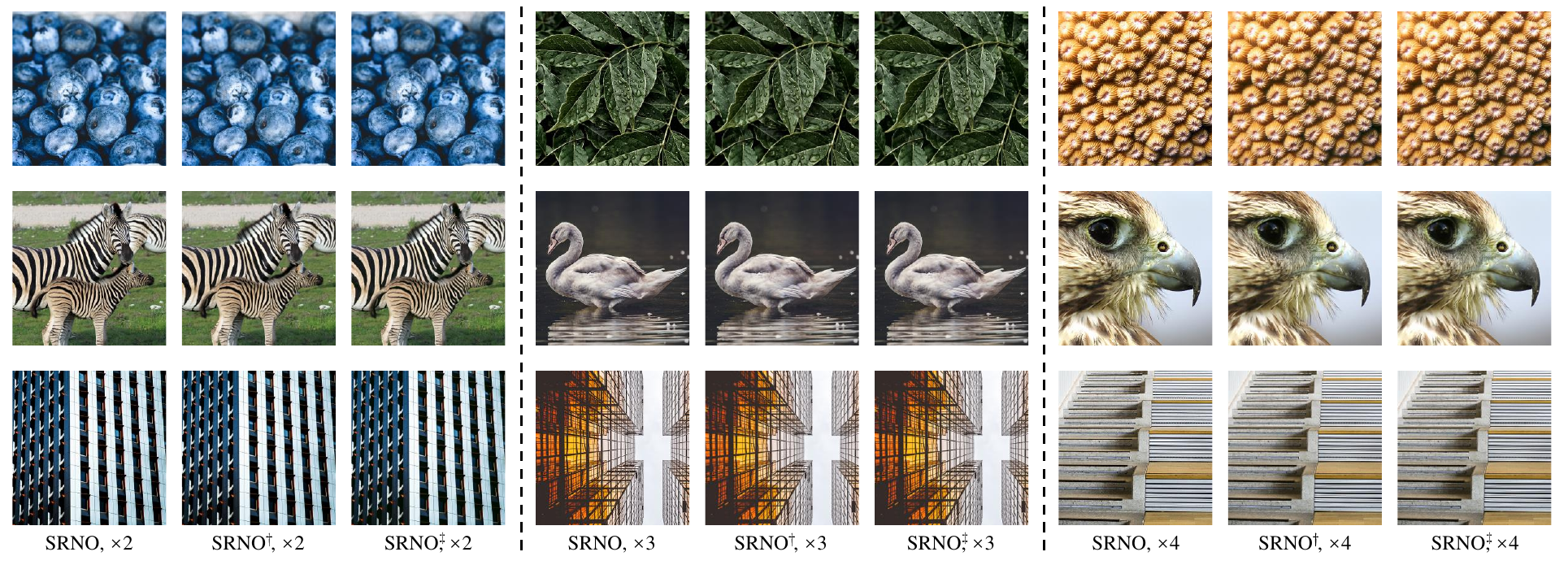}
    \caption{Visualization comparison of SRNO~\cite{srnowei2023super}, its AnySR variant (highlighted by $^\dag$), and AnySR-retrained version (denoted by $^\ddag$) in representative scenarios. Line 1: dense and fine grains/textures; Line 2: animal fur textures; Line 3: dense and complex lines.}
    \label{fig:special_scene}
\end{figure*}

\begin{figure*}[!t]
    \centering
    \includegraphics[width=0.92\textwidth]{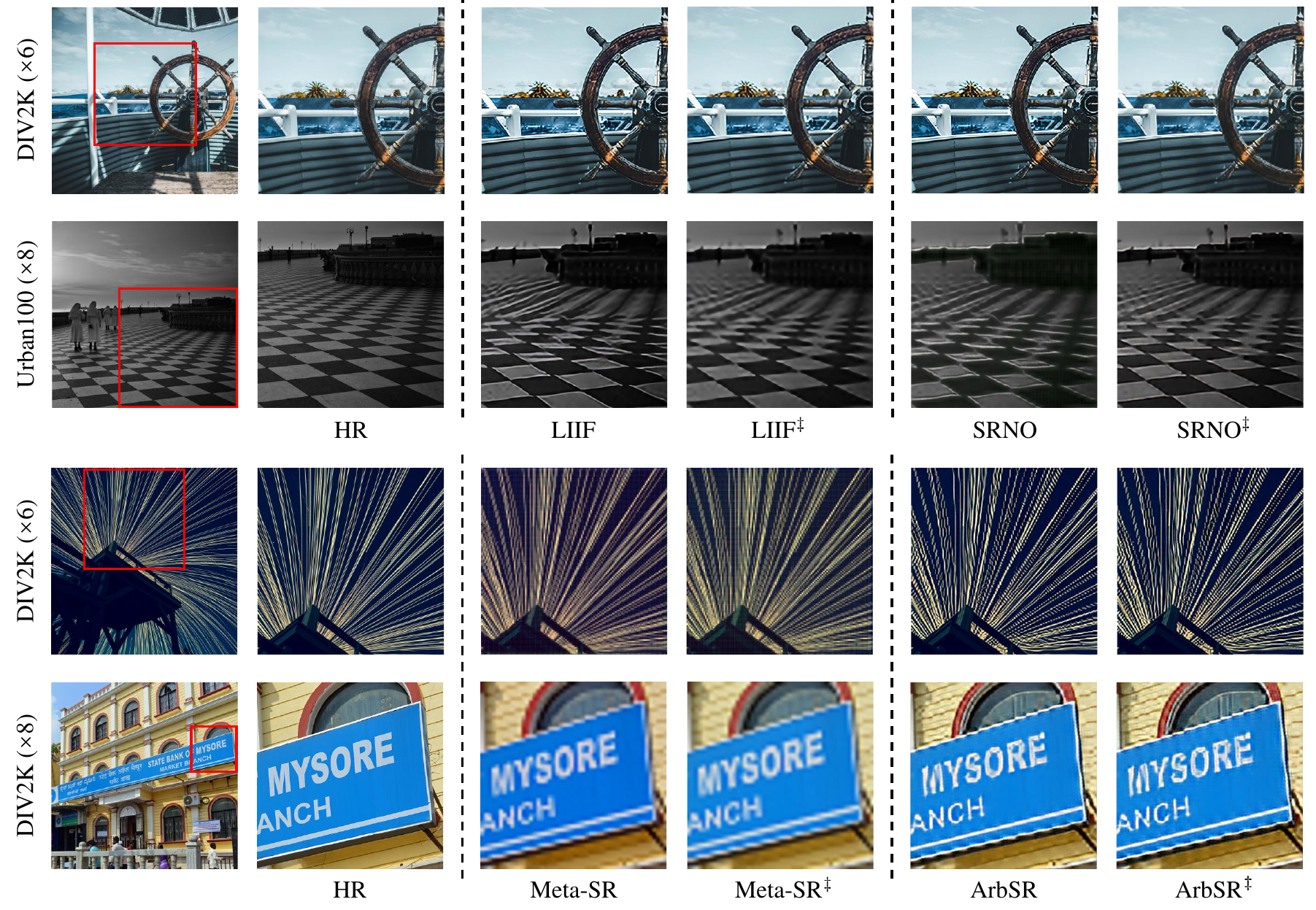}
    \caption{Visualization results for out-of-distribution scales of existing arbitrary-scale SR models and their AnySR-retrained version (denoted by $^\ddag$).}
    \label{fig:out_of_scale}
    \vspace{-0.5em}
\end{figure*}

\subsubsection{Qualitative Results.}
A large number of qualitative experiments have been carried out, demonstrating that AnySR can generate competitive, high-fidelity visual images while reducing computational costs.

\textbf{In-Distribution Scales.}
Fig.\,\ref{fig:visual} makes a comparison between off-the-shelf arbitrary-scale SR models and their corresponding AnySR variants on DIV2K~\cite{agustsson2017ntire} and Urban100~\cite{huang2015single} under the two specific settings including: different subnets for different scales and the entire network for all scales.

Qualitative results of the models are satisfactory, as both AnySR$^\dag$ and AnySR$^\ddag$ avoid generating artifacts or distortions. 
As shown in Line 2 and Line 3 of Fig.\,\ref{fig:visual}, both clear building edges and rich textural details on the leaves are retained.

We conduct evaluations of the model's reconstruction ability across different scales using representative scenarios. 
In Line 1 of Fig.\,\ref{fig:special_scene}, we task our AnySR with reconstructing fine-grained, densely arranged objects to assess its handling of detail, yielding commendable results. 
We can clearly observe the frost on blueberries, the dew on leaves, and the texture on coral surfaces. 
In Line 2, AnySR successfully reconstructs complex animal patterns (zebra stripes) and realistic, delicate animal fur (feathers of the swan and the eagle). 
In Line 3, our model accurately reconstructs the intricate and intertwined lines of buildings, resulting in clear and organized outputs.

All reconstructed results demonstrate the robust reconstruction capability of AnySR. Even at the $\times$2 scale, where computational costs are reduced by nearly 30\%, the model still achieves visual results on par with the original model. This reflects AnySR’s ability to deliver high-quality reconstructed images, even under varying computational resource conditions.

\textbf{Out-of-Distribution Scales.}
Images from the test set are downsampled and subsequently input into both the original model and its AnySR-retrained version for reconstruction, with the results compared against the corresponding high-resolution (HR) images, as illustrated in Fig.\,\ref{fig:out_of_scale}. Since the entire network is chosen under the out-of-distribution scale setting, the AnySR$\dag$ results are omitted.
For LIIF and SRNO, which perform well on out-of-distribution scales, our AnySR achieves notable improvements while fully retaining their reconstruction capabilities. For example, in Line 2 of Fig.\,\ref{fig:out_of_scale}, AnySR reconstructs more regular paving stones with fewer distortions and deformations.
For Meta-SR and ArbSR, which originally shows mediocre performance on out-of-distribution scales, AnySR training leads to performance improvements. This is particularly evident in Meta-SR, as shown in Lines 3 and 4 of Fig.\,\ref{fig:out_of_scale}. In the out-of-distribution scale setting, Meta-SR’s reconstruction results exhibit noticeable color discrepancies, but after retraining with AnySR, the model corrects the color inconsistencies and reduces image blurring.

%
%

\begin{table*}[!]
\centering
\caption{Comparison of the numbers of Parameters, PSNR performances, and FLOPs count for existing arbitrary-scale SR methods and their AnySR variants (through different subnets) highlighted by $^\dag$, AnySR-retrained version (through the largest network) denoted by $^\ddag$.
Parameters and FLOPs are measured with consistent pixel size for upsampling SR images across all scales.
}
\label{tab:flops}
\setlength\tabcolsep{6pt}
\resizebox{\textwidth}{!}{
\renewcommand{\arraystretch}{1.5}
\begin{tabular}{c|c|c|c|c|c|c|c}
\toprule
\multirow{2}*{Model} & \multirow{2}*{Params (M)} & \multicolumn{2}{c}{$\times 2$} & \multicolumn{2}{|c}{$\times 3$} & \multicolumn{2}{|c}{$\times 4$} \\
\cline{3-8}
&     & PSNR (dB)  &  FLOPs (G) &    PSNR (dB)  &  FLOPs (G)   &  PSNR (dB)    &   FLOPs (G)              \\
\cline{1-8}

MetaSR        & 1.22 (100.00\%) & 33.57 ($+$0.00) & 140.20  (100.00\%) & 30.29 ($+$0.00) & 62.03 (100.00\%) & 28.52 ($+$0.00) & 35.01 (100.00\%) \\
\cline{1-8}
MetaSR$^\dag$ & 1.50 (122.95\%) & 33.44 ($-$0.13) & 97.73 (69.71\%) & 30.28 ($-$0.01)  & 55.45 (89.40\%) & 28.55 ($+$0.03) & 35.01 (100.00\%) \\
\cline{1-8}
 MetaSR$^\ddag$& 1.50 (122.95\%) & 33.58 ($+$0.01) & 140.20 (100.00\%) & 30.32 ($+$0.03) & 62.03 (100.00\%) & 28.55 ($+$0.03) & 35.01 (100.00\%) \\  
\cline{1-8}

LIIF        & 1.22 (100.00\%) & 33.66 ($+$0.00) & 140.20 (100.00\%) & 30.34 ($+$0.00) & 53.03 (100.00\%) & 28.62 ($+$0.00) & 35.01 (100.00\%) \\
\cline{1-8}
 LIIF$^\dag$ & 1.50 (122.95\%) & 33.53 ($-$0.13) & 97.73 (69.71\%) & 30.32 ($-$0.02)  & 45.23 (85.30\%) & 28.61 ($-$0.01) & 35.01 (100.00\%) \\
\cline{1-8}
LIIF$^\ddag$& 1.50 (122.95\%) & 33.62 ($-$0.04) & 140.20 (100.00\%) & 30.35 ($+$0.01) & 53.02 (100.00\%) & 28.61 ($-$0.01) & 35.01 (100.00\%) \\  
\cline{1-8}

ArbSR        & 1.22 (100.00\%) & 33.70 ($+$0.00) & 140.92 (100.00\%) & 30.31 ($+$0.00) & 62.34 (100.00\%) & 28.56 ($+$0.00) & 35.14 (100.00\%) \\
\cline{1-8}
 ArbSR$^\dag$ & 1.50 (122.95\%) & 33.46 ($-$0.24) & 101.86 (72.28\%) & 30.28 ($-$0.03)  & 57.26 (91.85\%) & 28.53 ($-$0.03) & 35.14 (100.00\%) \\
\cline{1-8}
 ArbSR$^\ddag$& 1.50 (122.95\%) & 33.65 ($-$0.05) & 140.92 (100.00\%) & 30.30 ($-$0.01) & 62.34 (100.00\%) & 28.53 ($-$0.03) & 35.14 (100.00\%) \\  
\cline{1-8}

SRNO        & 1.22 (100.00\%) & 33.83 ($+$0.00) & 141.28 (100.00\%) & 30.50 ($+$0.00) & 62.50 (100.00\%) & 28.79 ($+$0.00) & 35.28 (100.00\%) \\
\cline{1-8}
 SRNO$^\dag$ & 1.50 (122.95\%) & 33.74 ($-$0.09) & 97.93 (69.25\%) & 30.47 ($-$0.03)  & 55.54 (88.86\%) & 28.78 ($-$0.01) & 35.28 (100.00\%) \\
\cline{1-8}
 SRNO$^\ddag$& 1.50 (122.95\%) & 33.81 ($-$0.02) & 141.28 (100.00\%) & 30.49 ($-$0.01) & 62.50 (100.00\%) & 28.78 ($-$0.01) & 35.28 (100.00\%) \\  
 \bottomrule

\end{tabular}
}
\end{table*}

\begin{table}[]
\centering
\caption{Comparison of the number of parameters and FLOPs of a single ResNet Block of EDSR~\cite{Lim_2017_CVPR_Workshops} in SRNO~\cite{srnowei2023super}, its AnySR variants and its AnySR-retrained version.}
\label{tab:detail_flops}
\setlength\tabcolsep{6pt}
\resizebox{0.48\textwidth}{!}{
\renewcommand{\arraystretch}{1.2}
\begin{tabular}{c|c|c|c|c|c|c}
\toprule

\multicolumn{1}{c|}{Setting} & Metric  & Conv  & Relu & GAP     & FC      & ALL      \\ \hline

\multirow{2}*{$\times$2, SRNO}  & Params (K) & 73.86 & 0.00      & -       & -    & 73.86    
\\ \cline{2-7}
 & FLOPs (M)  & 8860.00  & 3.84   & - & - & 8863.84 \\
\hline

\multirow{2}*{$\times$2, SRNO$\dag$}  & Params (K) & 73.86 & 0.00      & 0.00       & 17.60    & 91.46    
\\ \cline{2-7}
 & FLOPs (M)  & 6090.00  & 2.64   & 2.64 & 0.024 & 6095.30 \\
 \hline
 
\multirow{2}*{$\times$2, SRNO$\ddag$}  & Params (K) & 73.86 & 0.00      & 0.00       & 17.60    & 91.46    
\\ \cline{2-7}
 & FLOPs (M)  & 8860.00  & 3.84   & 3.84 & 0.035 & 8867.72 \\
 \hline


 \multirow{2}*{$\times$3, SRNO}  & Params (K) & 73.86 & 0.00      & -       & -    & 73.86    
\\ \cline{2-7}
 & FLOPs (M)  & 3920.00  & 1.70   & - & - & 3921.70 \\
\hline

\multirow{2}*{$\times$3, SRNO$\dag$}  & Params (K) & 73.86 & 0.00      & 0.00       & 17.60    & 91.46    
\\ \cline{2-7}
 & FLOPs (M)  & 3500.00  & 1.51   & 1.51 & 0.031 & 3503.05 \\
 \hline

\multirow{2}*{$\times$3, SRNO$\ddag$}  & Params (K) & 73.86 & 0.00      & 0.00       & 17.60    & 91.46    
\\ \cline{2-7}
 & FLOPs (M)  & 3920.00  & 1.70   & 1.70 & 0.035 & 3923.44 \\
 \hline

 
\multirow{2}*{$\times$4, SRNO}  & Params (K) & 73.86 & 0.00      & -       & -    & 73.86    
\\ \cline{2-7}
 & FLOPs (M)  & 2220.00  & 0.96   & - & - & 2220.96 \\
\hline

\multirow{2}*{$\times$4,SRNO$\dag$}  & Params (K) & 73.86 & 0.00      & 0.00      & 17.60    & 91.46    
\\ \cline{2-7}
 & FLOPs (M)  & 2220.00  & 0.96   & 0.96 & 0.035 & 2221.96 \\
 \hline
 
\multirow{2}*{$\times$4,SRNO$\ddag$}  & Params (K) & 73.86 & 0.00      & 0.00       & 17.60    & 91.46    
\\ \cline{2-7}
 & FLOPs (M)  & 2220.00  & 0.96   & 0.96 & 0.035 & 2221.96 \\ \bottomrule
\end{tabular}
}
\end{table}

\subsection{Complexity Analysis}
\label{sec:computation}
We perform a thorough comparison on SR tasks at different scales, encompassing the number of parameters, PSNR performance, and FLOPs, for existing arbitrary-scale SR methods and their AnySR variants of different subnets to solve different scales and the entire network to solve all scales.
All results are evaluated on top of the EDSR~\cite{Lim_2017_CVPR_Workshops} backbone network on the Set14 dataset~\cite{zeyde2012single} as shown in Table\,\ref{tab:flops}.
Since the encoder contains 16 ResNet blocks, we calculate the parameters and FLOPs for a single image, for a single ResNet block under different settings, as shown in Table\,\ref{tab:detail_flops}.
Taking SRNO as an example, we conduct a thorough and detailed analysis of the parameter count and computational cost.

In terms of parameters, our AnySR variant introduces only 0.28M additional weights (22.95\% of the original 1.22M parameters). As shown in Table\,\ref{tab:detail_flops}, these additional weights come from the fully connected layers, which are primarily dedicated to any-scale enhancement.

As for computational cost, as shown in Table\,\ref{tab:detail_flops}, despite the introduction of fully connected layers and GAP operations, the FLOPs are almost negligible compared to those required by the convolutional layers of the original model. Instead, the selection of smaller subnets leads to significant savings in FLOPs.
For example, when performing the $\times2$ task with selected subnetworks, each ResNet block saves around 2800M FLOPs in the convolutional layers, while FLOPs introduced by the GAP operation and fully connected layers are less than 3M. The total number of FLOPs (97.93G) constitutes only 69.25\% of the original 141.28G, with a minimal performance loss of 0.09 dB, making it suitable for devices with limited computational power to output high-quality results.

In case of ample computing resources, AnySR maintains a good performance-to-consumption balance. When performing the $\times4$ task with the entire network, Table\,\ref{tab:detail_flops} shows that each ResNet block introduces negligible 0.995M FLOPs. For the entire backbone, the FLOPs of the AnySR-retrained version (35.28G) are approximately equal to those of the original model, implying no computing introduction.

\begin{table*}[!t]
\caption{PSNR (dB) and LPIPS comparison of SRNO using smaller subnets (denoted as $F_i$) to perform out-of-distribution scale SR tasks. \underline{Underlines} indicates the best performance.}
\vspace{-2mm}
\label{tab:subnet_outofscale}
\centering
\setlength\tabcolsep{6.5pt}
\resizebox{\textwidth}{!}{
\renewcommand{\arraystretch}{1.5}
\begin{tabular}{c|c|ccccc|ccccc|ccccc}
\toprule
&       & \multicolumn{5}{c|}{Set5}                                                                                               & \multicolumn{5}{c|}{  Set14}                                                                                                         & \multicolumn{5}{c}{Urban100}                         \\ \cline{3-17} 
\multirow{-2}{*}{  Backbone} & \multirow{-2}{*}{  Method} & \multicolumn{1}{c|}{   $\times$6}    & \multicolumn{1}{c|}{   $\times$12}   & \multicolumn{1}{c|}{   $\times$18}   & \multicolumn{1}{c|}{   $\times$24}   &    $\times$30   & \multicolumn{1}{c|}{   $\times$6}    & \multicolumn{1}{c|}{   $\times$12}   & \multicolumn{1}{c|}{   $\times$18}   & \multicolumn{1}{c|}{   $\times$24}   &    $\times$30   & \multicolumn{1}{c|}{   $\times$6}    & \multicolumn{1}{c|}{   $\times$12}   & \multicolumn{1}{c|}{   $\times$18}   & \multicolumn{1}{c|}{   $\times$24}   &    $\times$30   \\ \hline

\multirow{12}{*}{EDSR}   
& Metric & \multicolumn{15}{c}{PSNR$\uparrow$}
\\
\cline{2-17}

 & SRNO                                            & \multicolumn{1}{c|}{\underline{29.06}} & \multicolumn{1}{c|}{\underline {24.62}} & \multicolumn{1}{c|}{\underline {22.52}} & \multicolumn{1}{c|}{\underline {21.37}} & \underline {20.54} & \multicolumn{1}{c|}{\underline {26.55}} & \multicolumn{1}{c|}{\underline {23.19}} & \multicolumn{1}{c|}{\underline {21.65}} & \multicolumn{1}{c|}{\underline {20.68}} & \underline {19.77} & \multicolumn{1}{c|}{\underline {24.08}} & \multicolumn{1}{c|}{\underline {21.10}} & \multicolumn{1}{c|}{\underline {19.75}} & \multicolumn{1}{c|}{\underline {18.96}} & \underline {18.40} \\ \cline{2-17}

& $F_1$                                             & \multicolumn{1}{c|}{28.64}                        & \multicolumn{1}{c|}{24.32}                        & \multicolumn{1}{c|}{22.31}                        & \multicolumn{1}{c|}{21.09}                        & 20.18                        & \multicolumn{1}{c|}{26.26}                        & \multicolumn{1}{c|}{22.98}                        & \multicolumn{1}{c|}{21.50}                        & \multicolumn{1}{c|}{20.51}                        & 19.63                        & \multicolumn{1}{c|}{23.53}                        & \multicolumn{1}{c|}{20.74}                        & \multicolumn{1}{c|}{19.47}                        & \multicolumn{1}{c|}{18.73}                        & 18.20                        \\ \cline{2-17}

 & $F_2$                                              & \multicolumn{1}{c|}{28.92}                        & \multicolumn{1}{c|}{24.55}                        & \multicolumn{1}{c|}{22.44}                        & \multicolumn{1}{c|}{21.26}                        & 20.41                        & \multicolumn{1}{c|}{26.46}                        & \multicolumn{1}{c|}{23.10}                        & \multicolumn{1}{c|}{21.58}                        & \multicolumn{1}{c|}{20.62}                        & 19.72                        & \multicolumn{1}{c|}{23.83}                        & \multicolumn{1}{c|}{20.92}                        & \multicolumn{1}{c|}{19.61}                        & \multicolumn{1}{c|}{18.82}                        & 18.28                        \\ \cline{2-17}

    & $F_3$                                               & \multicolumn{1}{c|}{29.00}                        & \multicolumn{1}{c|}{24.57}                        & \multicolumn{1}{c|}{22.45}                        & \multicolumn{1}{c|}{21.29}                        & 20.51                        & \multicolumn{1}{c|}{26.52}                        & \multicolumn{1}{c|}{23.14}                        & \multicolumn{1}{c|}{21.61}                        & \multicolumn{1}{c|}{20.66}                        & 19.74                        & \multicolumn{1}{c|}{23.96}                        & \multicolumn{1}{c|}{20.99}                        & \multicolumn{1}{c|}{19.66}                        & \multicolumn{1}{c|}{18.87}                        & 18.33                        \\ \cline{2-17}

   & $F_4$ (SRNO$\ddag$)                                         & \multicolumn{1}{c|}{29.05}                        & \multicolumn{1}{c|}{24.57}                        & \multicolumn{1}{c|}{22.51}                        & \multicolumn{1}{c|}{21.31}                        & 20.52                        & \multicolumn{1}{c|}{\underline {26.55}} & \multicolumn{1}{c|}{23.15}                        & \multicolumn{1}{c|}{21.61}                        & \multicolumn{1}{c|}{20.65}                        & 19.73                        & \multicolumn{1}{c|}{24.01}                        & \multicolumn{1}{c|}{21.01}                        & \multicolumn{1}{c|}{19.67}                        & \multicolumn{1}{c|}{18.88}                        & 18.33                        \\ \cline{2-17}

& Metric & \multicolumn{15}{c}{LPIPS$\downarrow$}
\\
\cline{2-17}

 & SRNO                                            & \multicolumn{1}{c|}{0.210}                        & \multicolumn{1}{c|}{0.394}                        & \multicolumn{1}{c|}{\underline{0.525}} & \multicolumn{1}{c|}{0.619}                        & \underline{0.636} & \multicolumn{1}{c|}{0.332}                        & \multicolumn{1}{c|}{0.496}                        & \multicolumn{1}{c|}{0.579}                        & \multicolumn{1}{c|}{0.637}                        & 0.687                        & \multicolumn{1}{c|}{0.281}                        & \multicolumn{1}{c|}{0.491}                        & \multicolumn{1}{c|}{0.610}                        & \multicolumn{1}{c|}{0.677}                        & \underline{0.719} \\ \cline{2-17}

 & $F_1$                                           & \multicolumn{1}{c|}{0.220}                        & \multicolumn{1}{c|}{0.403}                        & \multicolumn{1}{c|}{0.548}                        & \multicolumn{1}{c|}{0.625}                        & 0.641                        & \multicolumn{1}{c|}{0.342}                        & \multicolumn{1}{c|}{0.508}                        & \multicolumn{1}{c|}{0.595}                        & \multicolumn{1}{c|}{0.652}                        & 0.698                        & \multicolumn{1}{c|}{0.307}                        & \multicolumn{1}{c|}{0.523}                        & \multicolumn{1}{c|}{0.641}                        & \multicolumn{1}{c|}{0.701}                        & 0.737                        \\ \cline{2-17}

 & $F_2$                                           & \multicolumn{1}{c|}{0.213}                        & \multicolumn{1}{c|}{0.398}                        & \multicolumn{1}{c|}{0.526}                        & \multicolumn{1}{c|}{0.624}                        & 0.650                        & \multicolumn{1}{c|}{0.333}                        & \multicolumn{1}{c|}{0.498}                        & \multicolumn{1}{c|}{0.581}                        & \multicolumn{1}{c|}{0.640}                        & 0.689                        & \multicolumn{1}{c|}{0.289}                        & \multicolumn{1}{c|}{0.497}                        & \multicolumn{1}{c|}{0.617}                        & \multicolumn{1}{c|}{0.683}                        & 0.725                        \\ \cline{2-17}

& $F_3$                                           & \multicolumn{1}{c|}{\underline{0.209}} & \multicolumn{1}{c|}{0.394}                        & \multicolumn{1}{c|}{0.527}                        & \multicolumn{1}{c|}{\underline{0.618}} & 0.641                        & \multicolumn{1}{c|}{0.331}                        & \multicolumn{1}{c|}{0.494}                        & \multicolumn{1}{c|}{0.579}                        & \multicolumn{1}{c|}{0.637}                        & \underline{0.686} & \multicolumn{1}{c|}{0.282}                        & \multicolumn{1}{c|}{0.489}                        & \multicolumn{1}{c|}{0.608}                        & \multicolumn{1}{c|}{0.676}                        & 0.720                        \\ \cline{2-17}

 & $F_4$ (SRNO$\ddag$)                             & \multicolumn{1}{c|}{\underline{0.209}} & \multicolumn{1}{c|}{\underline{0.392}} & \multicolumn{1}{c|}{0.529}                        & \multicolumn{1}{c|}{0.620}                        & 0.641                        & \multicolumn{1}{c|}{\underline{0.329}} & \multicolumn{1}{c|}{\underline{0.491}} & \multicolumn{1}{c|}{\underline{0.578}} & \multicolumn{1}{c|}{\underline{0.636}} & 0.687                        & \multicolumn{1}{c|}{\underline{0.279}} & \multicolumn{1}{c|}{\underline{0.485}} & \multicolumn{1}{c|}{\underline{0.605}} & \multicolumn{1}{c|}{\underline{0.674}} & \underline{0.719} \\ 

 \midrule

\multirow{12}{*}{  RDN}  
& Metric & \multicolumn{15}{c}{PSNR$\uparrow$}
\\
\cline{2-17}

 & SRNO                                            & \multicolumn{1}{c|}{29.40}                        & \multicolumn{1}{c|}{\underline{24.94}} & \multicolumn{1}{c|}{22.73}                        & \multicolumn{1}{c|}{\underline{21.50}} & \underline{20.65} & \multicolumn{1}{c|}{\underline{26.76}} & \multicolumn{1}{c|}{\underline{23.33}} & \multicolumn{1}{c|}{\underline{21.82}} & \multicolumn{1}{c|}{20.78}                        & \underline{19.96} & \multicolumn{1}{c|}{\underline{24.42}} & \multicolumn{1}{c|}{\underline{21.35}} & \multicolumn{1}{c|}{\underline{19.95}} & \multicolumn{1}{c|}{\underline{19.15}} & \underline{18.55} \\ \cline{2-17}

  & $F_1$                                           & \multicolumn{1}{c|}{29.01}                        & \multicolumn{1}{c|}{24.72}                        & \multicolumn{1}{c|}{22.46}                        & \multicolumn{1}{c|}{21.32}                        & 20.49                        & \multicolumn{1}{c|}{26.53}                        & \multicolumn{1}{c|}{23.20}                        & \multicolumn{1}{c|}{21.70}                        & \multicolumn{1}{c|}{20.64}                        & 19.81                        & \multicolumn{1}{c|}{24.02}                        & \multicolumn{1}{c|}{21.06}                        & \multicolumn{1}{c|}{19.73}                        & \multicolumn{1}{c|}{18.94}                        & 18.37                        \\ \cline{2-17}

& $F_2$                                           & \multicolumn{1}{c|}{29.20}                        & \multicolumn{1}{c|}{24.81}                        & \multicolumn{1}{c|}{22.56}                        & \multicolumn{1}{c|}{21.28}                        & 20.56                        & \multicolumn{1}{c|}{26.65}                        & \multicolumn{1}{c|}{23.24}                        & \multicolumn{1}{c|}{21.72}                        & \multicolumn{1}{c|}{20.68}                        & 19.82                        & \multicolumn{1}{c|}{24.22}                        & \multicolumn{1}{c|}{21.18}                        & \multicolumn{1}{c|}{19.80}                        & \multicolumn{1}{c|}{19.00}                        & 18.41                        \\ \cline{2-17}

& $F_3$                                           & \multicolumn{1}{c|}{29.23}                        & \multicolumn{1}{c|}{24.76}                        & \multicolumn{1}{c|}{22.64}                        & \multicolumn{1}{c|}{21.30}                        & 20.61                        & \multicolumn{1}{c|}{26.67}                        & \multicolumn{1}{c|}{23.23}                        & \multicolumn{1}{c|}{21.76}                        & \multicolumn{1}{c|}{20.75}                        & 19.83                        & \multicolumn{1}{c|}{24.25}                        & \multicolumn{1}{c|}{21.19}                        & \multicolumn{1}{c|}{19.82}                        & \multicolumn{1}{c|}{19.00}                        & 18.42                        \\ \cline{2-17}

& $F_4$ (SRNO$\ddag$)                             & \multicolumn{1}{c|}{\underline{29.45}} & \multicolumn{1}{c|}{24.87}                        & \multicolumn{1}{c|}{\underline{22.76}} & \multicolumn{1}{c|}{21.45}                        & 20.62                        & \multicolumn{1}{c|}{26.74}                        & \multicolumn{1}{c|}{23.26}                        & \multicolumn{1}{c|}{21.75}                        & \multicolumn{1}{c|}{\underline{20.79}} & 19.88                        & \multicolumn{1}{c|}{24.38}                        & \multicolumn{1}{c|}{21.27}                        & \multicolumn{1}{c|}{19.86}                        & \multicolumn{1}{c|}{19.06}                        & 18.46                        \\ \cline{2-17}

& Metric & \multicolumn{15}{c}{LPIPS$\downarrow$}
\\
\cline{2-17}

& SRNO                                            & \multicolumn{1}{c|}{0.206}                        & \multicolumn{1}{c|}{0.387}                        & \multicolumn{1}{c|}{0.514}                        & \multicolumn{1}{c|}{0.591}                        & 0.628                        & \multicolumn{1}{c|}{0.323}                        & \multicolumn{1}{c|}{0.485}                        & \multicolumn{1}{c|}{0.565}                        & \multicolumn{1}{c|}{0.623}                        & 0.674                        & \multicolumn{1}{c|}{0.267}                        & \multicolumn{1}{c|}{0.474}                        & \multicolumn{1}{c|}{0.595}                        & \multicolumn{1}{c|}{0.665}                        & 0.711                        \\ \cline{2-17}

& $F_1$                                           & \multicolumn{1}{c|}{0.218}                        & \multicolumn{1}{c|}{0.402}                        & \multicolumn{1}{c|}{0.532}                        & \multicolumn{1}{c|}{0.599}                        & 0.636                        & \multicolumn{1}{c|}{0.337}                        & \multicolumn{1}{c|}{0.502}                        & \multicolumn{1}{c|}{0.585}                        & \multicolumn{1}{c|}{0.643}                        & 0.686                        & \multicolumn{1}{c|}{0.294}                        & \multicolumn{1}{c|}{0.511}                        & \multicolumn{1}{c|}{0.626}                        & \multicolumn{1}{c|}{0.688}                        & 0.727                        \\ \cline{2-17}

& $F_2$                                           & \multicolumn{1}{c|}{0.210}                        & \multicolumn{1}{c|}{\underline{0.375}} & \multicolumn{1}{c|}{\underline{0.501}} & \multicolumn{1}{c|}{\underline{0.571}} & \underline{0.614} & \multicolumn{1}{c|}{0.328}                        & \multicolumn{1}{c|}{0.484}                        & \multicolumn{1}{c|}{0.562}                        & \multicolumn{1}{c|}{0.619}                        & 0.666                        & \multicolumn{1}{c|}{0.274}                        & \multicolumn{1}{c|}{0.476}                        & \multicolumn{1}{c|}{0.590}                        & \multicolumn{1}{c|}{0.658}                        & 0.704                        \\ \cline{2-17}

& $F_3$                                           & \multicolumn{1}{c|}{0.210}                        & \multicolumn{1}{c|}{0.384}                        & \multicolumn{1}{c|}{0.513}                        & \multicolumn{1}{c|}{0.578}                        & 0.621                        & \multicolumn{1}{c|}{0.328}                        & \multicolumn{1}{c|}{0.487}                        & \multicolumn{1}{c|}{0.562}                        & \multicolumn{1}{c|}{0.617}                        & 0.668                        & \multicolumn{1}{c|}{0.273}                        & \multicolumn{1}{c|}{0.477}                        & \multicolumn{1}{c|}{0.592}                        & \multicolumn{1}{c|}{0.658}                        & 0.704                        \\ \cline{2-17}

& $F_4$ (SRNO$\ddag$)                             & \multicolumn{1}{c|}{\underline{0.203}} & \multicolumn{1}{c|}{0.378}                        & \multicolumn{1}{c|}{0.503}                        & \multicolumn{1}{c|}{0.582}                        & 0.624                        & \multicolumn{1}{c|}{\underline{0.322}} & \multicolumn{1}{c|}{\underline{0.479}} & \multicolumn{1}{c|}{\underline{0.559}} & \multicolumn{1}{c|}{\underline{0.612}} & \underline{0.663} & \multicolumn{1}{c|}{\underline{0.263}} & \multicolumn{1}{c|}{\underline{0.461}} & \multicolumn{1}{c|}{\underline{0.582}} & \multicolumn{1}{c|}{\underline{0.654}} & \underline{0.703} \\ 

\bottomrule

\end{tabular}}
\end{table*}

\begin{figure*}[!t]
    \centering
    \includegraphics[width=0.98\textwidth]{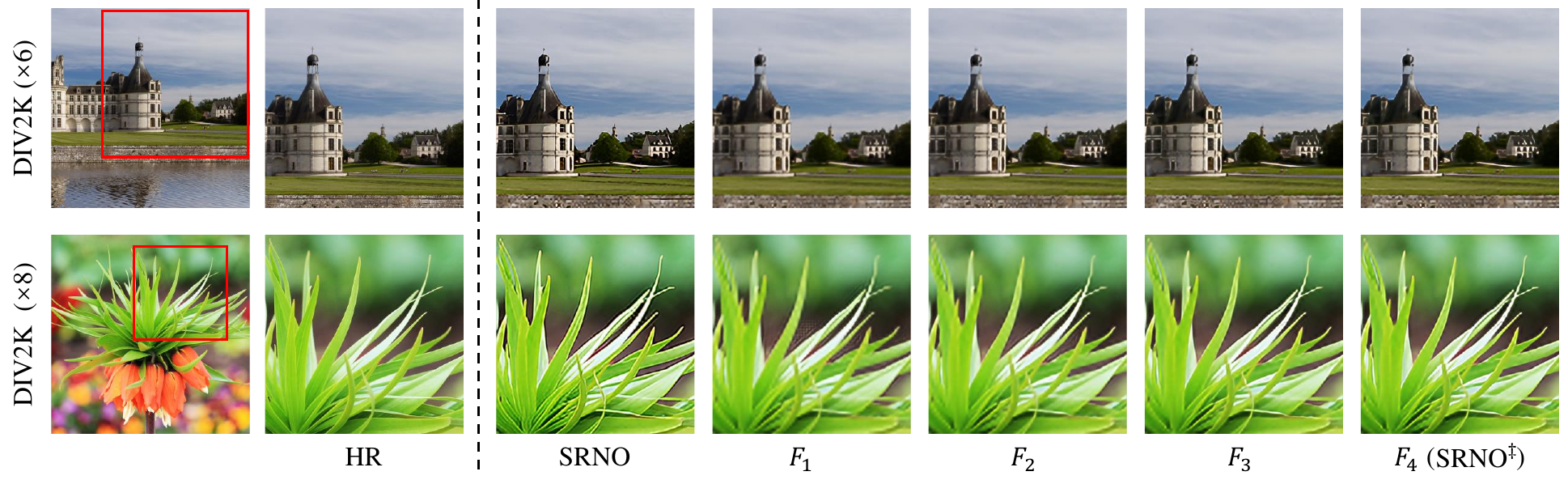}
    \caption{Visualization results of SRNO using smaller subnets (denoted as $F_i$) to perform out-of-distribution scale SR tasks on the EDSR backbone.}
    \label{fig:subnet_out_of_scale}
   \vspace{-0.5em}
\end{figure*}

\subsection{Subnet Exploration}
We evaluate smaller subnets in handling high magnification factors. 
Table\,\ref{tab:subnet_outofscale} and Fig.\,\ref{fig:subnet_out_of_scale} present the quantitative and qualitative results of SRNO and its AnySR variants with subnets of varying sizes, denoted as $F_i$, on out-of-distribution scales. 
Smaller subnets perform lower-scored PSNR, as reflected in the visual results with less detail recovery such as distant lighthouses and the fine edges of plant leaves. However, it improves as subnet size increases. 
The LPIPS scores show slight variation, indicating that AnySR’s subnets generally maintain good perceptual quality in the generated images.

\subsection{Ablation Studies}
\label{sec:ablation}

We conduct ablation studies to validate the effectiveness of individual components of AnySR. 
All ablation experiments are performed on SRNO~\cite{srnowei2023super}, using EDSR~\cite{Lim_2017_CVPR_Workshops} as the backbone. 
We compare our AnySR with two variants: 
1) ``w/o ASE'': AnySR without any-scale enhancement; 
2) ``w/o FI'': AnySR replacing feature-interweaving with a simple concatenation.
Also, we vary the value of reset probability $p$ in Line 4 of Algorithm\,\ref{alg:training}, to show its importance.
%
%

\begin{table*}[!t]
\centering
\caption{Ablation results of PSNR(dB) and LPIPS on alternations of AnySR including the ``w/o ASE'' variant removing any-scale enhancement (ASE) and the ``w/o FI'' variant replacing feature-interweaving (FI) with a simple concatenation. 
\underline{Underlines} indicates the best performance.}
\vspace{-2mm}
\label{tab:ablation1-psnr}
\setlength\tabcolsep{6.5pt}
\resizebox{\textwidth}{!}{

\renewcommand{\arraystretch}{1.5}
\begin{tabular}{c|ccc|ccc|ccc|ccc|ccc}
\toprule
\multirow {2}*{Settings} & \multicolumn{3}{c}{Set5} & \multicolumn{3}{|c}{Set14}
 & \multicolumn{3}{|c}{B100}  & \multicolumn{3}{|c}{Urban100}  & \multicolumn{3}{|c}{Manga109} \\
\cline{2-16}
            & \multicolumn{1}{c|}{$\times2$}   & \multicolumn{1}{c|}{$\times3$}   &  $\times4$
            & \multicolumn{1}{c|}{$\times2$}   & \multicolumn{1}{c|}{$\times3$}   &  $\times4$
            & \multicolumn{1}{c|}{$\times2$}   & \multicolumn{1}{c|}{$\times3$}   &  $\times4$
            & \multicolumn{1}{c|}{$\times2$}   & \multicolumn{1}{c|}{$\times3$}   &  $\times4$
            & \multicolumn{1}{c|}{$\times2$}   & \multicolumn{1}{c|}{$\times3$}   &  $\times4$ \\
\hline
Metric
&\multicolumn{15}{c}{PSNR$\uparrow$}
\\
\hline

AnySR$^\dag$ 
& \multicolumn{1}{c|}{\underline{38.04}}   & \multicolumn{1}{c|}{34.50}   & \underline{32.38}
& \multicolumn{1}{c|}{\underline{33.74}}   & \multicolumn{1}{c|}{\underline{30.47}}   & \underline{28.78}
& \multicolumn{1}{c|}{\underline{32.19}}   & \multicolumn{1}{c|}{\underline{29.15}}   & \underline{27.64}
& \multicolumn{1}{c|}{\underline{32.27}}   & \multicolumn{1}{c|}{\underline{28.45}}   & \underline{26.45}
& \multicolumn{1}{c|}{\underline{38.78}}   & \multicolumn{1}{c|}{\underline{33.77}}   & \underline{30.83}\\
\cline{1-16}

AnySR$^\dag$ (w/o ASE) 
& \multicolumn{1}{c|}{38.01}   & \multicolumn{1}{c|}{\underline{34.51}}   & 32.34
& \multicolumn{1}{c|}{33.72}   & \multicolumn{1}{c|}{30.43}   & 28.73
& \multicolumn{1}{c|}{32.14}   & \multicolumn{1}{c|}{29.11}   & 27.61
& \multicolumn{1}{c|}{\underline{32.27}}   & \multicolumn{1}{c|}{28.44}   & 26.43
& \multicolumn{1}{c|}{38.70}   & \multicolumn{1}{c|}{33.74}   & 30.79\\
\cline{1-16}

AnySR$^\dag$ (w/o FI) 
& \multicolumn{1}{c|}{38.00}   & \multicolumn{1}{c|}{34.47}   & 32.33
& \multicolumn{1}{c|}{33.71}   & \multicolumn{1}{c|}{30.42}   & 28.73
& \multicolumn{1}{c|}{32.14}   & \multicolumn{1}{c|}{29.12}   & 27.61
& \multicolumn{1}{c|}{32.25}   & \multicolumn{1}{c|}{28.43}   & 26.42
& \multicolumn{1}{c|}{38.75}   & \multicolumn{1}{c|}{33.72}   & 30.81\\
\cline{1-16}

AnySR$^\ddag$
& \multicolumn{1}{c|}{\underline{38.12}}   & \multicolumn{1}{c|}{34.51}   &  \underline{32.38} 
& \multicolumn{1}{c|}{\underline{33.81}}   & \multicolumn{1}{c|}{\underline{30.49}}   &  \underline{28.78} 
& \multicolumn{1}{c|}{\underline{32.25}}   & \multicolumn{1}{c|}{\underline{29.17}}   &  \underline{27.64} 
& \multicolumn{1}{c|}{\underline{32.52}}   & \multicolumn{1}{c|}{\underline{28.50}}   &  \underline{26.45}   
& \multicolumn{1}{c|}{\underline{38.96}}   & \multicolumn{1}{c|}{\underline{33.83}}   &  \underline{30.83}  \\
\cline{1-16}

AnySR$^\ddag$ (w/o ASE) 
& \multicolumn{1}{c|}{38.10}   & \multicolumn{1}{c|}{\underline{34.52}}   & 32.34  
& \multicolumn{1}{c|}{33.76}   & \multicolumn{1}{c|}{30.45}   & 28.73  
& \multicolumn{1}{c|}{32.21}   & \multicolumn{1}{c|}{29.12}   & 27.61  
& \multicolumn{1}{c|}{32.51}   & \multicolumn{1}{c|}{28.47}   & 26.43    
& \multicolumn{1}{c|}{38.90}   & \multicolumn{1}{c|}{33.81}   & 30.79   \\
\cline{1-16}

AnySR$^\ddag$ (w/o FI)   
& \multicolumn{1}{c|}{38.10}   & \multicolumn{1}{c|}{34.48}   & 32.33  
& \multicolumn{1}{c|}{33.76}   & \multicolumn{1}{c|}{30.45}   &  28.73 
& \multicolumn{1}{c|}{32.22}   & \multicolumn{1}{c|}{29.14}   & 27.61  
& \multicolumn{1}{c|}{32.50}   & \multicolumn{1}{c|}{28.46}   & 26.42    
& \multicolumn{1}{c|}{38.91}   & \multicolumn{1}{c|}{33.80}   & 30.81   \\

\hline
Metric&\multicolumn{15}{c}{LPIPS$\downarrow$}
\\
\hline

AnySR$^\dag$ 
& \multicolumn{1}{c|}{\underline{0.041}}      & \multicolumn{1}{c|}{\underline{0.092}} & \underline{0.138} & \multicolumn{1}{c|}{\underline{0.072}} 
& \multicolumn{1}{c|}{\underline{0.164}} & \underline{0.227} & \multicolumn{1}{c|}{\underline{0.112}} & \multicolumn{1}{c|}{\underline{0.230}} &  \underline{0.304} & \multicolumn{1}{c|}{\underline{0.048}} & \multicolumn{1}{c|}{\underline{0.120}} &  \underline{0.176} & \multicolumn{1}{c|}{ \underline{0.016}} & \multicolumn{1}{c|}{\underline{0.046}} & \underline{0.078}
\\
\cline{1-16}

AnySR$^\dag$ (w/o ASE) 
& \multicolumn{1}{c|}{0.045}   & \multicolumn{1}{c|}{0.096}   & 	0.142	
& \multicolumn{1}{c|}{\underline{0.072}}   & \multicolumn{1}{c|}{0.167}   & 0.230	
& \multicolumn{1}{c|}{0.116}   & \multicolumn{1}{c|}{0.235}   & 0.305	
& \multicolumn{1}{c|}{0.053}   & \multicolumn{1}{c|}{0.121}   & 0.179	
& \multicolumn{1}{c|}{0.019}   & \multicolumn{1}{c|}{0.053}   & 0.080
\\
\cline{1-16}

AnySR$^\dag$ (w/o FI) 
& \multicolumn{1}{c|}{0.044}   & \multicolumn{1}{c|}{ 0.095}   &  	0.139 	
& \multicolumn{1}{c|}{0.075}   & \multicolumn{1}{c|}{	0.167}   & 0.229 	
& \multicolumn{1}{c|}{0.115}   & \multicolumn{1}{c|}{	0.235}   & 	0.305 	
& \multicolumn{1}{c|}{0.049}   & \multicolumn{1}{c|}{0.122}   & 	0.179 	
& \multicolumn{1}{c|}{0.018}   & \multicolumn{1}{c|}{0.049}   & 	0.079
\\
\cline{1-16}

AnySR$^\ddag$
& \multicolumn{1}{c|}{\underline{0.040}}   & \multicolumn{1}{c|}{\underline{0.093}} 
&  \underline{0.138} 

& \multicolumn{1}{c|}{\underline{0.071}} 
& \multicolumn{1}{c|}{\underline{0.163}} 
&  \underline{0.227} 

& \multicolumn{1}{c|}{\underline{0.112}} 
& \multicolumn{1}{c|}{\underline{0.230}} 
& \underline{0.304} 

& \multicolumn{1}{c|}{\underline{0.047}} 
& \multicolumn{1}{c|}{\underline{0.119}} 
& \underline{0.176} 

& \multicolumn{1}{c|}{\underline{0.016}} 
& \multicolumn{1}{c|}{ \underline{0.046}} & \underline{0.078}
\\
\cline{1-16}

AnySR$^\ddag$ (w/o ASE) 
& \multicolumn{1}{c|}{0.043}   & \multicolumn{1}{c|}{0.096}   & 	0.142 	
& \multicolumn{1}{c|}{0.073}   & \multicolumn{1}{c|}{	0.165}   & 	0.230 	
& \multicolumn{1}{c|}{0.115}   & \multicolumn{1}{c|}{	0.234 }   & 	0.305 	
& \multicolumn{1}{c|}{0.052}   & \multicolumn{1}{c|}{0.121}   & 	0.179 	
& \multicolumn{1}{c|}{0.018}   & \multicolumn{1}{c|}{0.052}   & 0.080
\\
\cline{1-16}

AnySR$^\ddag$ (w/o FI)   
& \multicolumn{1}{c|}{0.043}   & \multicolumn{1}{c|}{0.095}   &  	0.139 
& \multicolumn{1}{c|}{	0.072}   & \multicolumn{1}{c|}{	0.165 }   & 	0.229	
& \multicolumn{1}{c|}{0.114 }   & \multicolumn{1}{c|}{	0.233 	}   & 0.305 	
& \multicolumn{1}{c|}{0.049}   & \multicolumn{1}{c|}{0.121}   & 0.179 	
& \multicolumn{1}{c|}{0.018}   & \multicolumn{1}{c|}{	0.048}   & 	0.079
\\

\bottomrule

\end{tabular}
}
\end{table*}

\begin{table*}[!t]
\centering
\caption{Ablation results of PSNR (dB) on the reset probability $p$. 
\underline{Underlines} indicates the best performance.}
\vspace{-2mm}
\label{tab:ablation2-dag}
\setlength\tabcolsep{5.5pt}
\resizebox{\textwidth}{!}{

\renewcommand{\arraystretch}{1.5}
\begin{tabular}{c|ccc|ccc|ccc|ccc|ccc}
\toprule
\multirow {2}*{Settings} & \multicolumn{3}{c}{Set5} & \multicolumn{3}{|c}{Set14}
 & \multicolumn{3}{|c|}{B100}  & \multicolumn{3}{|c|}{Urban100}  & \multicolumn{3}{|c}{Manga109} \\
\cline{2-16}
            & \multicolumn{1}{c|}{$\times2$}   & \multicolumn{1}{c|}{$\times3$}   &  $\times4$
            & \multicolumn{1}{c|}{$\times2$}   & \multicolumn{1}{c|}{$\times3$}   &  $\times4$
            & \multicolumn{1}{c|}{$\times2$}   & \multicolumn{1}{c|}{$\times3$}   &  $\times4$
            & \multicolumn{1}{c|}{$\times2$}   & \multicolumn{1}{c|}{$\times3$}   &  $\times4$
            & \multicolumn{1}{c|}{$\times2$}   & \multicolumn{1}{c|}{$\times3$}   &  $\times4$ \\

\cline{1-16}

 & \multicolumn{15}{c}{AnySR Variants} \\
\hline

AnySR$^\dag$ (p=0) 
& \multicolumn{1}{c|}{38.01}   & \multicolumn{1}{c|}{34.45}   & 	32.37 	 
& \multicolumn{1}{c|}{33.73 }   & \multicolumn{1}{c|}{	30.45 	}   &  28.77 	
& \multicolumn{1}{c|}{32.18 }   & \multicolumn{1}{c|}{	29.13}   &  	27.64 	
& \multicolumn{1}{c|}{32.25}   & \multicolumn{1}{c|}{	28.44 }   &  	26.44 	  
& \multicolumn{1}{c|}{38.77 	}   & \multicolumn{1}{c|}{33.77 	}   &  30.82 \\
\cline{1-16}

AnySR$^\dag$ (p=0.2) 
& \multicolumn{1}{c|}{38.03}   & \multicolumn{1}{c|}{34.45}   &  32.37 	
& \multicolumn{1}{c|}{33.73}   & \multicolumn{1}{c|}{30.45}   &  28.76 	
& \multicolumn{1}{c|}{32.18}   & \multicolumn{1}{c|}{	29.13}   &  27.64 	
& \multicolumn{1}{c|}{32.25}   & \multicolumn{1}{c|}{28.44}   &  	\underline{26.45} 	 
& \multicolumn{1}{c|}{38.76}   & \multicolumn{1}{c|}{33.77}   &  30.82 \\
\cline{1-16}

AnySR$^\dag$ (p=0.4) 
& \multicolumn{1}{c|}{38.03 }   & \multicolumn{1}{c|}{	34.47}   &  32.37 	
& \multicolumn{1}{c|}{33.72 	}   & \multicolumn{1}{c|}{30.45}   & 	28.77 	
& \multicolumn{1}{c|}{\underline{32.19}  }   & \multicolumn{1}{c|}{	29.13  }   &  	\underline{27.65}	
& \multicolumn{1}{c|}{32.26}   & \multicolumn{1}{c|}{	28.43 }   & 	26.44 	   
& \multicolumn{1}{c|}{38.77}   & \multicolumn{1}{c|}{	33.77}   &  	\underline{30.83} \\
\cline{1-16}

AnySR$^\dag$ (p=0.6) 
& \multicolumn{1}{c|}{\underline{38.04}}  & \multicolumn{1}{c|}{\underline{34.50}}   & \underline{32.38}
& \multicolumn{1}{c|}{\underline{33.74}}   & \multicolumn{1}{c|}{\underline{30.47}}   & \underline{28.78}
& \multicolumn{1}{c|}{\underline{32.19}} & \multicolumn{1}{c|}{\underline{29.15}}   & \underline{27.65}
& \multicolumn{1}{c|}{\underline{32.27}}   & \multicolumn{1}{c|}{\underline{28.45}}   & \underline{26.45}
& \multicolumn{1}{c|}{\underline{38.78}}   & \multicolumn{1}{c|}{33.77}   & \underline{30.83}  \\
\cline{1-16}

AnySR$^\dag$ (p=0.8) 
& \multicolumn{1}{c|}{38.03  }   & \multicolumn{1}{c|}{	34.49}   & 	\underline{32.38} 	 
& \multicolumn{1}{c|}{33.73 }   & \multicolumn{1}{c|}{	30.46 	}   &  28.77 
& \multicolumn{1}{c|}{	32.18 }   & \multicolumn{1}{c|}{	29.14 }   &  	\underline{27.65} 	
& \multicolumn{1}{c|}{32.26 }   & \multicolumn{1}{c|}{	28.44 	}   & \underline{26.45} 	  
& \multicolumn{1}{c|}{\underline{38.78} }   & \multicolumn{1}{c|}{	\underline{33.78} 	 }   &  30.82 \\

\hline
 & \multicolumn{15}{c}{AnySR-Retrained Version}\\
\hline

AnySR$^\ddag$ (p=0) 
& \multicolumn{1}{c|}{38.06}   & \multicolumn{1}{c|}{34.47}   &   32.37
& \multicolumn{1}{c|}{33.76}   & \multicolumn{1}{c|}{30.46}   &   28.77
& \multicolumn{1}{c|}{32.21}   & \multicolumn{1}{c|}{29.14}   &   27.64
& \multicolumn{1}{c|}{32.48}   & \multicolumn{1}{c|}{28.46}   &   26.44  
& \multicolumn{1}{c|}{38.91}   & \multicolumn{1}{c|}{33.82}   &   30.82 \\
\cline{1-16}

AnySR$^\ddag$ (p=0.2)   
& \multicolumn{1}{c|}{38.07}   & \multicolumn{1}{c|}{34.47}   & 32.37  
& \multicolumn{1}{c|}{33.78}   & \multicolumn{1}{c|}{30.46}   &  28.76 
& \multicolumn{1}{c|}{32.22}   & \multicolumn{1}{c|}{29.14}   & 27.64  
& \multicolumn{1}{c|}{32.50}   & \multicolumn{1}{c|}{28.47}   & \underline{26.45}    
& \multicolumn{1}{c|}{38.93}   & \multicolumn{1}{c|}{33.83}   & 30.82   \\
\cline{1-16}

AnySR$^\ddag$ (p=0.4)
& \multicolumn{1}{c|}{38.08}   & \multicolumn{1}{c|}{34.48}   &  32.37 
& \multicolumn{1}{c|}{33.78}   & \multicolumn{1}{c|}{30.46}   &  28.77 
& \multicolumn{1}{c|}{32.22}   & \multicolumn{1}{c|}{29.14}   & \underline{27.65} 
& \multicolumn{1}{c|}{32.50}   & \multicolumn{1}{c|}{28.47}   & 26.44    
& \multicolumn{1}{c|}{38.94}   & \multicolumn{1}{c|}{33.83}   & \underline{30.83}   \\
\cline{1-16}

AnySR$^\ddag$ (p=0.6)
& \multicolumn{1}{c|}{\underline{38.12}}   & \multicolumn{1}{c|}{\underline{34.51}}   &  \underline{32.38} 
& \multicolumn{1}{c|}{\underline{33.81}}   & \multicolumn{1}{c|}{\underline{30.49}}   &  \underline{28.78} 
& \multicolumn{1}{c|}{\underline{32.25}}   & \multicolumn{1}{c|}{\underline{29.17}}   &  \underline{27.65} 
& \multicolumn{1}{c|}{\underline{32.52}}   & \multicolumn{1}{c|}{\underline{28.50}}   &  \underline{26.45}   
& \multicolumn{1}{c|}{\underline{38.96}}   & \multicolumn{1}{c|}{33.83}   &  \underline{30.83}  \\
\cline{1-16}

AnySR$^\ddag$ (p=0.8)
& \multicolumn{1}{c|}{38.11}   & \multicolumn{1}{c|}{\underline{34.51}}   & \underline{32.38}  
& \multicolumn{1}{c|}{33.80}   & \multicolumn{1}{c|}{\underline{30.49}}   & 28.77  
& \multicolumn{1}{c|}{32.24}   & \multicolumn{1}{c|}{29.16}   & \underline{27.65}  
& \multicolumn{1}{c|}{32.51}   & \multicolumn{1}{c|}{28.49}   & \underline{26.45}    
& \multicolumn{1}{c|}{\underline{38.96}}   & \multicolumn{1}{c|}{\underline{33.84}}   & 30.82   \\

\bottomrule

\end{tabular}
}
\end{table*}

\textbf{Any-Scale Enhancement.}
Any-scale enhancement constitutes one of the fundamental branches encompassed within our overarching research framework.
By injecting, excavating, and ameliorating sufficient scale information, we realize customized handling for features at different scales. 
To validate the effectiveness of ASE, we train the network by removing this component, and the results are presented in Table\,\ref{tab:ablation1-psnr}. It is evident that ``w/o ASE'' leads to a certain performance drop.

\textbf{Feature-Interweaving.}
Feature-interweaving considers the non-uniformity of shared weights and overcomes the mutual weight influence across different scales by repeating and inserting scale pairs at regular intervals.
We study this mechanism by substitute the feature-interweaving fashion with a simple concatenation in earlier methods~\cite{chen2021learning,cao2023ciaosr}.
With feature-interweaving, a better performance is achieved in Table\,\ref{tab:ablation1-psnr}.

\textbf{Reset Probability $p$.}
As mentioned in Sec.\,\ref{sec:anyresource}, we reset subnet $F_t$ to the entire network $F$ with a certain probability $p$ to retain the original ability. 
We conduct ablation experiments with different values of $p= \{0, 0.2, 0.4, 0.6, 0.8 \}$.  
Note that $p = 0$ indicates the removal of the reset mechanism.
The results are presented in Table\,\ref{tab:ablation2-dag}. A low $p$, in particular for $p = 0$, leads to minimal impact on enhancing the performance of the entire network during inference. 
Hence, opting for $p=0.6$ or $0.8$ is deemed more appropriate.
Nevertheless, it is observed that with $p=0.8$, the performance improvement for the entire network inference is slightly higher than $p=0.6$, and a higher probability incurs higher training costs.
Considering the trade-off between the performance of the entire network inference and network training costs, we ultimately select $p=0.6$ as the experimental setting for the study presented in this paper.

\section{Conclusion and Future Work}
\label{sec:conclusion}
In this paper, we presented AnySR, a simple yet versatile approach to transform existing arbitrary-scale super-resolution methods into implementations that adapt to any scale and resource availability.
By rebuilding arbitrary-scale tasks into an any-resource implementation, we enable the completion of smaller-scale SISR tasks with reduced computational resources and no additional parameters. 
To maintain performance, we enhance any-scale features through a feature-interweaving fashion, ensuring sufficient scale information and correct feature/scale processing. 
Extensive experiments on benchmark datasets demonstrate the efficiency and scalability of our AnySR method in arbitrary-scale SISR applications.

An alternative approach is resorting to a more complex NAS (Neural Architecture Search), which may achieve better performance and will be our major future exploration.

\bibliographystyle{IEEEtran}
\bibliography{main}

\begin{IEEEbiography}[{\includegraphics[width=1in,height=1.25in,clip,keepaspectratio]{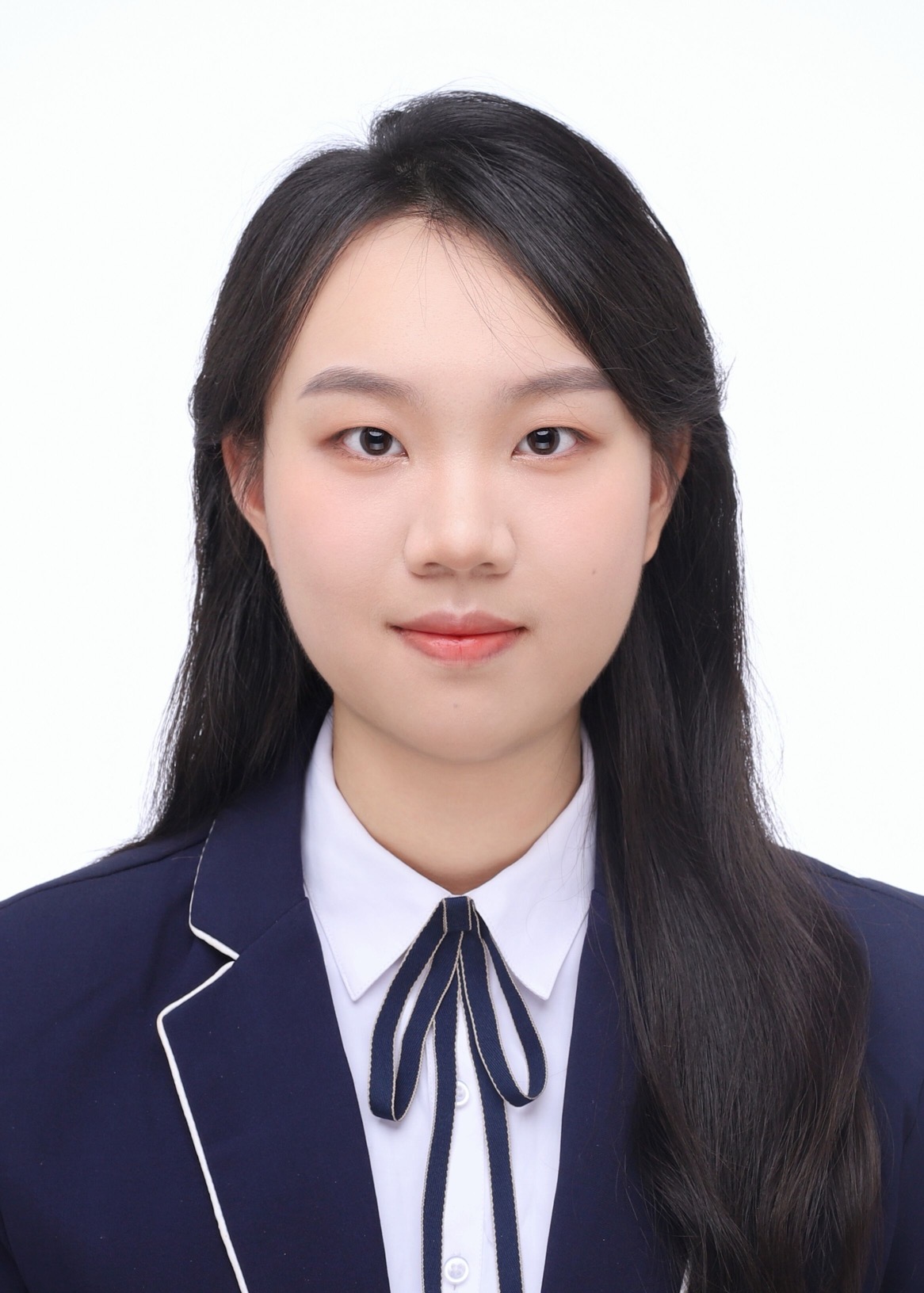}}]{Wengyi Zhan}  completed her undergraduate studies and obtained the Bachelor's degree in Intelligence Science and Technology from Xiamen University, Xiamen, China, in 2023. 

She is currently a master's student at the School of Informatics, Xiamen University, specializing in Intelligence Science and Technology. Her current research interest focuses on diffusion models for consistent video/image generation.
\end{IEEEbiography}

\begin{IEEEbiography}[{\includegraphics[width=1in,height=1.25in,clip,keepaspectratio]{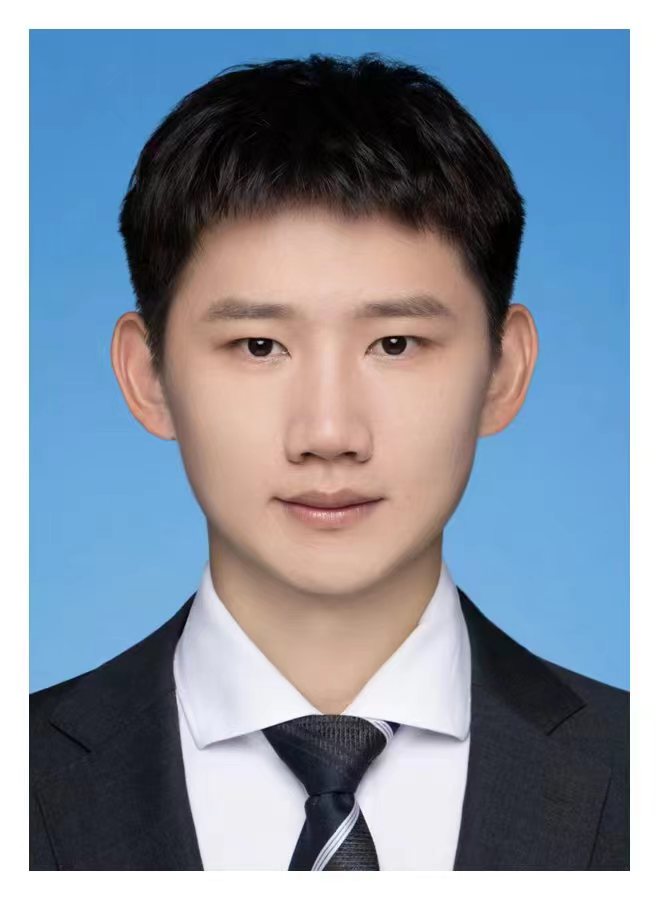}}]{Mingbao Lin} finished his M.S.-Ph.D. study and obtained the Ph.D. degree in intelligence science and technology from Xiamen University, Xiamen, China, in 2022. Earlier, he received the B.S. degree from Fuzhou University, Fuzhou, China, in 2016.

He is currently a research scientist with the Skywork AI, Singapore, and also an adjunct industry supervisor with Xiamen University. His current research interest is to develop low-latency multimodal interaction system, such as text, audio, image, video.
\end{IEEEbiography}
\vspace{-5mm}

\begin{IEEEbiography}[{\includegraphics[width=1in,height=1.25in,clip,keepaspectratio]{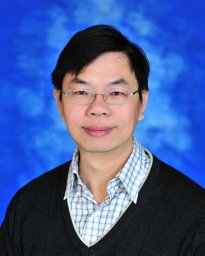}}]{Chia-Wen Lin}
(Fellow, IEEE) received his Ph.D. degree in Electrical Engineering from National Tsing Hua University (NTHU), Hsinchu, Taiwan, in 2000.  
	Dr. Lin is currently a Distinguished Professor with the Department of Electrical Engineering and the Institute of Communications Engineering, NTHU.  His research interests include image/video processing and computer vision.  He has served as a Fellow Evaluation Committee member (2021--2023), BoG Member-at-Large (2022--2024), and Distinguished Lecturer (2018--2019) of IEEE Circuits and Systems Society.   He was Chair of IEEE ICME Steering Committee (2020--2021). He served as TPC Co-Chair of IEEE ICIP 2019, IEEE ICME 2010, and PCS 2022, and General Co-Chair of IEEE VCIP 2018 and PCS 2024.  He received two best paper awards from VCIP 2010 and 2015. He has served as an Associate Editor of \textsc{IEEE Transactions on Image Processing}, \textsc{IEEE Transactions on Circuits and Systems for Video Technology}, \textsc{IEEE Transactions on Multimedia},  \textsc{IEEE Multimedia}, and Neural Networks.
\end{IEEEbiography}
\vspace{-5mm}

\begin{IEEEbiography}[{\includegraphics[width=1in,height=1.25in,clip,keepaspectratio]{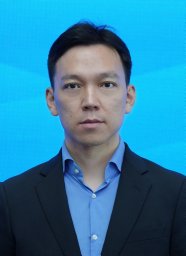}}]{Rongrong Ji}
(Senior Member, IEEE) is a Nanqiang Distinguished Professor at Xiamen University, the  Director of the Office of Science and Technology at Xiamen University, and the Director of Media Analytics and Computing Lab. He was awarded as the National Science Foundation for Excellent Young Scholars (2014), the National Ten Thousand Plan for Young Top Talents (2017), and the National Science Foundation for Distinguished Young Scholars (2020). His research falls in the field of computer vision, multimedia analysis, and machine learning. He has published 50+ papers in ACM/IEEE Transactions, including TPAMI and IJCV, and 100+ full papers on top-tier conferences, such as CVPR and NeurIPS. His publications have got over 20K citations in Google Scholar. He was the recipient of the Best Paper Award of ACM Multimedia 2011. He has served as Area Chairs in top-tier conferences such as CVPR and ACM Multimedia. He is also an Advisory Member for Artificial Intelligence Construction in the Electronic Information Education Committee of the National Ministry of Education.
\end{IEEEbiography}

\newpage

\vfill

\end{document}